\RequirePackage{fix-cm}
\documentclass[11pt]{article}
\usepackage[a4paper, total={6.5in, 10in}]{geometry}
\usepackage{authblk}

\usepackage{graphicx}
\usepackage{latexsym}
\usepackage[caption=false]{subfig}
\usepackage{graphicx}
\usepackage{amssymb}
\usepackage{amsmath,amscd,amsfonts}
\usepackage[dvipsnames]{xcolor}
\usepackage{times}
\usepackage{tikz}
\usepackage[version=4]{mhchem}

\RequirePackage[english,linesnumbered,algoruled]{algorithm2e}
\usepackage[subnum]{cases}
\usepackage{hhline}
\usepackage{multirow}
\usepackage{siunitx}
\usepackage{diagbox}
\usepackage{bbold}

\RequirePackage[english,linesnumbered,algoruled]{algorithm2e}

\graphicspath{{picts/}}

\usepackage[hyperindex=true]{hyperref}

\usepackage{color}
\definecolor{linkcol}{rgb}{0,0,0.4}
\definecolor{citecol}{RGB}{90,95,0}
\hypersetup
{
bookmarksopen=true,
pdftitle="Auto-weighted Bayesian Physics-Informed Neural Networks and robust estimations for multitask inverse problems in pore-scale imaging of dissolution",
pdfauthor="Sarah Perez and Philippe Poncet",
pdfsubject="We present a novel data assimilation strategy in pore-scale imaging and demonstrate that this makes it possible to robustly address reactive inverse problems incorporating Uncertainty Quantification (UQ).", 
pdfmenubar=true, 
pdfhighlight=/O, 
colorlinks=true, 
pdfpagemode=UseNone, 
pdfpagelayout=SinglePage, 
pdffitwindow=true, 
linkcolor=linkcol, 
citecolor=citecol, 
urlcolor=linkcol 
}

\def\ds{\displaystyle}
\def\Im{\mathrm{Im}}
\def\Da#1{\mathrm{Da}_\textsc{\romannumeral #1}} 
\def\Re{{\mathrm{Re}}} 
\def\Pe{{\mathrm{Pe}}} 
\def\L{\mathbb{L}}
\def\eps{\varepsilon}
\def\RAI{\text{RAI}}

\renewcommand{\bar}[1]{\overline{#1}}

\renewcommand{\leq}{\leqslant}
\renewcommand{\geq}{\geqslant}

\begin{document}

\title{\bf Auto-weighted Bayesian Physics-Informed Neural Networks\\ and robust estimations for multitask inverse problems\\ in pore-scale imaging of dissolution\footnote{This version of the article has been accepted for publication in Computational Geosciences, after peer review and is subject to Springer Nature’s AM terms of use, but is not the Version of Record and does not reflect post-acceptance improvements, or any corrections. The Version of Record is available online at: \url{https://dx.doi.org/10.1007/s10596-024-10313-x}}} 


\author[$\dag,*$]{Sarah Perez}\author[$\dag$]{Philippe Poncet}

\affil[$\dag$]{\small Universite de Pau et des Pays de l’Adour, E2S UPPA, CNRS, LMAP UMR CNRS-UPPA 5142, Pau, France.}
\affil[$*$]{\small The Lyell Centre, Heriot-Watt University, Edinburgh, United Kingdom.}
\renewcommand\Authands{ and }

\date{Received: August 28, 2023 / Accepted: July 25, 2024}

\maketitle

\begin{abstract}

In this article, we present a novel data assimilation strategy in pore-scale imaging and demonstrate that this makes it possible to robustly address reactive inverse problems incorporating Uncertainty Quantification (UQ). Pore-scale modeling of reactive flow offers a valuable opportunity to investigate the evolution of macro-scale properties subject to dynamic processes in the context of Carbon Capture and Storage (CCS). Yet, they suffer from imaging limitations arising from the associated X-ray microtomography (X-ray $\mu$CT) process, which induces discrepancies in the properties estimates. Assessment of the kinetic parameters also raises challenges, as reactive coefficients are critical parameters that can cover a wide range of values. We account for these two issues and ensure reliable calibration of pore-scale modeling, based on dynamical $\mu$CT images, by integrating uncertainty quantification in the workflow. 

The present method is based on a multitasking formulation of reactive inverse problems combining data-driven and physics-informed techniques in calcite dissolution. This allows quantifying morphological uncertainties on the porosity field and estimating reactive parameter ranges through prescribed PDE models, with a latent concentration field, and dynamical $\mu$CT observations. The data assimilation strategy relies on sequential reinforcement incorporating successively additional PDE constraints and suitable formulation of the heterogeneous diffusion differential operator leading to enhanced computational efficiency. We provide a robust and unbiased uncertainty quantification by straightforward adaptive weighting of Bayesian Physics-Informed Neural Networks (BPINNs), ensuring reliable micro-porosity changes during geochemical transformations.

We demonstrate successful Bayesian Inference in 1D+Time calcite dissolution based on synthetic $\mu$CT images with meaningful posterior distribution on the reactive parameters and dimensionless numbers. We eventually apply this framework to a more realistic 2D+Time data assimilation problem involving heterogeneous porosity levels and synthetic $\mu$CT dynamical observations. 
\end{abstract}

{\small

\noindent{\bf Funding:} This work was partially supported by E2S UPPA and by the Carnot Institute ISIFoR project MicroMineral under Grant Agreement P450902ISI, and by ANR project 20-CE45-0022 MucoReaDy.\\

\noindent{\bf Article Highlights:} 
\begin{itemize}\itemsep 0pt 
\item The method presented can carry out the porosity and concentration fields (observable or latent) in pore-scale reactive flows with evolving fluid-solid interface, together with characteristic numbers (Damköhler, molecular diffusion, chemical coefficients and tortuosity index).
\item The multitask formulation of such inverse problems is based on an automatic weighting of Bayesian Physics-Informed Neural Networks (BPINNs) and involves naturally uncertainty quantification. 
\item The inverse problems of dissolution 1D+Time and 2D+Time at the pore scale are provided.
\end{itemize}
}

\newpage

\normalsize
\section{Introduction}

Studying reactive flows in porous media is essential to manage the geochemical effects arising from \ce{CO2} capture and storage in natural underground reservoirs. While long-term predictions are commonly modeled at the field scale \cite{class_benchmark_2009}, pore-scale approaches meanwhile provide insights into local geochemical interactions between the injected \ce{CO2} and the aquifer structure \cite{payton_pore_2022}. Through mathematical homogenization of the sub-micrometer porous medium and appropriate modeling, one can simulate the reactive processes that occur at the pore scale and predict their impact on the macro-scale properties \cite{allaire_homogenization_2012, ALLAIRE20102292}. Geochemical processes are critical components for understanding the mineral trapping mechanisms and local evolving interfaces, either due to precipitation, crystallization, or dissolution within the porous environment. In this sense, investigating the impact of such reactive processes provides insight into reservoir safety submitted to chemical interactions that may compromise the aquifer structure. Pore-scale modeling of reactive flow hence appears as a complementary mean to field scale studies wherein homogenization theory bridges the gap between these scales. 

Pore-scale modeling in porous media is intrinsically related to X-ray microtomography (X-ray $\mu$CT) experiments. Advances in this imaging technique coupled with efficient numerical simulation offer a valuable opportunity to investigate dynamical processes and study the evolving macro-scale properties, such as the upscaled porosity and permeability \cite{andra_digital_2013-1, mostaghimi_computations_2013}. This is of great importance in the risk management perspective of \ce{CO2} storage, and therefore ensuring the reliability of pore-scale modeling and simulation appears as crucial. Uncertainties, however, arise from the microtomography imaging process where artifacts, noise, and unresolved morphological features are intrinsic limitations inducing important deviations in the estimation of petrophysical properties \cite{perez_deviation_2022, CARRILLO2022}. In particular, quantifying the impact of sub-resolved porosity in $\mu$CT images is identified as critical for geosciences applications \cite{LIN2016306, smal_automatic_2018}. This limiting factor arises from the compromise between the field of view being investigated and the image resolution. For multi-scale porous media such as carbonate rocks, this trade-off can readily result in scan resolutions that do not fully resolve morphological features of the pore space. Intrinsic limiting factors remain in the X-ray $\mu$CT imaging process, and investigating their effects and related uncertainties is fundamental to developing more accurate predictive models at the pore scale. 

In addition to these imaging uncertainties, proper assessment of the kinetic parameters raises challenges in the pore-scale modeling of reactive flows. Mineral reactivities, including reactive surface area, are critical parameters to account though they commonly suffer from discrepancies of several orders of magnitude \cite{noiriel_direct_2019}. Providing uncertainty estimates on these kinetic parameters is essential to ensure reliable calibration of pore-scale models for \ce{CO2} mineral storage assessment. Unsuitable characterization of the reactive surface area, for instance, will considerably affect the numerical model generating highly distinct behaviors that can become inconsistent with experimental investigations. Such concern is widely known, and several experimental works have developed potential solutions that address dynamical $\mu$CT imaging processes of carbonate dissolution \cite{Noiriel4D, menke_dynamic_2015}. This relies on 4D $\mu$CT and differential imaging techniques to derive averaged reaction rates and provide local maps of mineral reactivity at the porous medium surface. However, dynamical $\mu$CT scans also suffer from trade-off issues that may disrupt the identification of these parameters \cite{Zhang_challenges}. In addition to potential sub-resolved porosity, one needs to consider the compromise between the acquisition time capturing the dynamical process and the image quality. This may result in noisy observation data or non-physical variations leading to misleading estimations of the kinetic parameters. Querying the reliability of reactive parameters involved in pore-scale modeling is crucial, and time-resolved experiments of dynamical processes offer such an opportunity while suffering from imaging limitations.

Overall, we identify two current challenges to address reliable pore-scale modeling of reactive flows based on $\mu$CT images and ensure trustable evolutions of the macro-scale properties. The first challenge aims at quantifying morphological uncertainties on the porous medium sample due to unresolved features resulting from X-ray $\mu$CT. Investigating the uncertainties in the micro-porosity field is a major concern, and neglecting these uncertainty effects can bias the determination of the evolving petrophysical properties in geological applications. The second challenge concerns the uncertainty quantification of the kinetic parameters for reactive processes. In this sense, providing reliable mineral reactivity from dynamical $\mu$CT remains critical in order to perform relevant direct numerical simulation at the pore scale. The present article addresses these two challenges and incorporates Uncertainty Quantification (UQ) concerns in the workflow of pore-scale modeling.

Accounting for these concerns, however, requires developing efficient data assimilation techniques to perform extensive parameter estimation studies, uncertainty quantification assessments, and improve model reliability. In fact, uncertainty quantification is commonly achieved through stochastic PDE models \cite{Stochastic_PDE, coheur_bayesian_2023} or probabilistic Markov Chain Monte Carlo (MCMC) methods embedding Bayesian inference \cite{MONDAL2010241, siena_impact_2020}. The main drawback being this requires numerous evaluations of the PDE model and can thus quickly become computationally expensive. To overcome such computational constraints, machine learning methods have appeared as a popular framework in geosciences and have shown effectiveness in building efficient surrogate models in PDE-based data assimilation problems \cite{WANG2021103555, gmd-2022-309}. This offers alternatives and complementary means to traditional numerical methods to improve predictive modeling based on observation data and investigate uncertainty quantification within a Bayesian context. The development of machine-learning surrogate modeling incorporating uncertainty has, therefore, garnered increasing interest for a wide range of scientific applications \cite{delia_machine_2022, coheur_bayesian_2023}.

A popular framework combining physics-based techniques, data-driven methodology, and intrinsic uncertainty quantification are Bayesian Physics-Informed Neural Networks (BPINNs) \cite{yang_b-pinns_2021}. This benefits from the advantages of neural network structures in building parameterized surrogate models and Bayesian inference standards in estimating probabilistic posterior distribution. BPINNs can, however, be prone to a range of pathological behaviors, especially in multi-objective and multiscale inverse problems. This is because their training amounts to sample from a weighted multitask posterior distribution for which the setting of the weights parameters is challenging. Ensuring robust Bayesian inference, indeed, hinges on properly estimating these distinct task weights. We thus rely on the efficient BPINNs framework developed in \cite{PEREZ2023112342}, which robustly addresses multi-objective and multiscale Bayesian inverse problems including latent field reconstruction. The strategy relies on an adaptive and automatic weighting of the target distribution parameters and objectives, which is an unlocking advance in bringing robustness and addressing more challenging problems within the BPINNs framework. Indeed, it benefits from enhanced convergence and stability compared to conventional formulations and reduces sampling bias by avoiding manual tuning of critical weighting parameters \cite{maddu_inverse_2022}. The adjusted weights bring information on the task uncertainties, improve the reliability of the noise-related and model adequacy estimates and ensure unbiased uncertainty quantification. All these characteristics are crucial to address reliable reactive inverse problems of calcite dissolution, and we thus built the present methodology upon this efficient data-assimilation framework.

In this article, we focus on a multitask inverse problem for reactive flows at the pore scale through data assimilation that incorporates uncertainty quantification by means of the Bayesian Physics-Informed Neural Networks framework presented in \cite{PEREZ2023112342}. We intend to develop a novel approach for pore-scale imaging problems that combines dynamical microtomography and physics-based regularization induced by the PDE model of dissolution processes, for which the images are substantially noisy. To the best of our knowledge, investigating morphological and mineral reactivity uncertainties from the perspective of coupling physics-based models with data-driven techniques is the main novelty of this work. This formulation presents the joint ability to infer altogether kinetic parameters and quantify the residual micro-porosity generated by unresolved features in the $\mu$CT images. Overall, we aim to ensure reliable calibration of the PDE model and account for the morphological imaging uncertainty to provide meaningful evolution of the petrophysical properties due to the reactive process. 

The present methodology relies on sequential reinforcement of the target posterior distribution, which successively incorporates additional constraints from the PDE model into the data assimilation process. This sequential splitting formulation arises from the strong coupling, in the reactive model, between the micro-porosity field related to the $\mu$CT observations and the solute concentration, which is a latent field. In particular, achieving the reconstruction of an adequate latent concentration for the reactive fluid field is crucial for the identification of the inverse parameters along with their uncertainty ranges, and requires the development of a dedicated sequential reinforcement approach. Therefore, we consider successive sampling steps dedicated to 1) preconditioning the micro-porosity surrogate model with pure regression on the dynamical $\mu$CT images, 2) preconditioning the latent reactive fluid and inferring a first reactive parameter through PDE-constrained tasks given a first modeling approximation of the coupled PDE problem, and 3) considering the overall data assimilation problem given the exact PDE model with two inverse parameters, a predictive posterior distribution on the micro-porosity and insight on the latent concentration field.In this sense, while the current strategy draws significant inspiration from the PINNs framework, in which the regression of physical fields typically precedes the incorporation of physics-based constraints, the sequential reinforcement approach presented in this article also introduces a modeling reinforcement which is essential for inferring latent fields in a strongly coupled PDE system. On top of that, we propose a differentiation strategy wherein we consider a reformulation of the heterogeneous diffusion differential operator involved in the PDE model. This enhances the computational efficiency of the BPINN surrogate model and shows that suitable differential operator expressions considerably improve the computational cost, especially when dealing with complex non-linear operators.

The main contributions of this article are summarized below: \begin{enumerate}
    \item We infer reactive inverse parameter uncertainty ranges in prescribed PDE models through suitable dimensionless formulations in inverse problems, for which we identify and define the corresponding dimensionless numbers.
    \item We quantify morphological uncertainties from a pore-scale perspective, coupling image-based and physics-informed techniques in dynamical dissolution processes.
    \item We improve the relevance and reliability of predictions in dynamical systems through data-driven approaches and robust Bayesian Inference methodology. 
    \item We provide reliable quantification of the micro-porosity changes during geochemical transformations, with a focus on calcite dissolution processes. 
    \item We built an intrinsic data assimilation strategy for pore-scale imaging inverse problems relying on a sequential reinforcement approach and suitable formulation of the heterogeneous diffusion differential operator.  
\end{enumerate}

The remainder of this manuscript is organized as follows: In Sect. \ref{sec:General_context}, we review the current challenges arising from uncertainty quantification concerns in pore-scale modeling of reactive flows, with a focus on $\mu$CT limitations and model reliability issues. We focus in Sect. \ref{subsec:DNS} on the formulation of pore-scale modeling of reactive flows that we consider to study the dynamical dissolution of calcite. Sect. \ref{sec:Pb_setup} describes the dimensionless expressions of the dissolution PDE model for direct and inverse problems. We identify the main differences in their formulations and we establish in Sect. \ref{subsec:Adim_inverse} the dimensionless inverse problem on calcite dissolution that we address in the data assimilation approach, which ends up with equation \eqref{eq:inverse_chem_adim_Vf}. Sect. \ref{sec:BPINNs_Method} is dedicated to presenting the efficient adaptive framework for Bayesian Physics-Informed Neural Networks, which has been developed in our previous work. In Sect. \ref{sec:Method}, we describe the proposed data assimilation strategy for pore-scale imaging inverse problems, with sequential reinforcement of the target posterior distribution and computational strategy for the differential operator expressions. We validate this strategy in Sect. \ref{sec:Results1D} on several 1D+Time test cases of calcite dissolution based on synthetic $\mu$CT images. This particularly demonstrates successful Bayesian inference of the reactive parameters with posterior distributions on the dimensionless numbers. This also highlights consistent UQ on the micro-porosity field with uncertainty ranges on the residual micro-porosity, potentially unresolved, arising from the $\mu$CT dynamical images. Finally, we apply in Sect. \ref{sec:Results} our methodology to a more realistic 2D+Time data assimilation problem of calcite dissolution with heterogeneous porosity levels and synthetic $\mu$CT dynamical observations.

\section{Uncertainty Quantification in pore-scale modeling of reactive flows: context and motivation}
\label{sec:General_context}

Pore-scale modeling of reactive flows plays a crucial role in the long-term management of \ce{CO2} capture and storage in natural underground reservoirs. Understanding the local geochemical interactions between the injected \ce{CO2} and the aquifer structure and how it impacts the reservoir macro-scale properties is an active field in porous media research \cite{li_upscaling_2006, mehmani_multiblock_2012, steefel_pore_2013, andrew_pore-scale_2014}. These geochemical effects include mineral trapping through precipitation and crystallization but also dissolution reactions associated with flow, and transport mechanisms \cite{payton_pore_2022, Hameli_2022}. Mathematical models of such processes, at the pore scale, are usually combined with Direct Numerical Simulations (DNS) of highly coupled and non-linear Partial Differential Equations (PDE). Such PDE systems characterize the local evolving interfaces and provide insight into reservoir safety submitted to chemical interactions \cite{menke_dynamic_2015, molins_mineralogical, noiriel_pore-scale_2021}. In this risk management perspective, ensuring the reliability of pore-scale modeling and simulation of reactive flows is therefore essential, and this requires embedding Uncertainty Quantification (UQ) concerns. 

\subsection{Modeling of pore-scale dissolution}
\label{subsec:DNS}

The study of geochemical processes related to \ce{CO2} capture and storage is crucial in the context of risk management and investigation of the coupled mechanisms occurring within aquifers. In particular, the dissolution of the carbonate rock architecture by the injected \ce{CO2} may compromise the integrity of the geological reservoir. Pore-scale modeling of dissolution phenomena in porous media, therefore, remains an extensive research area \cite{soulaine_roman_kovscek_tchelepi_2018, molins_simulation_2021}. These mathematical models require a thin description of the highly heterogeneous pore structure in order to account for local interactions. The present article focus on the pore-scale dissolution of calcite subject to acidic transport in the subsoil. We, therefore, target the following irreversible chemical reaction with uniform stoichiometric coefficients:
\begin{equation}
    \label{eq:chem_eq}
    \centering
     \ce{CaCO3 (s) + H+ -> Ca^{2+} + HCO3-}  
\end{equation} 

In this section, we present the mathematical model used to simulate the calcite dissolution process \eqref{eq:chem_eq} at the pore scale. We introduce a spatial domain $\Omega \subset \mathbb{R}^n$, $n=1,2,3$ which corresponds to the porous medium described at its pore scale. This sample description involves a pure fluid region $\Omega_F$, also called void-space and assumed to be a smooth connected open set, and a surrounding solid matrix $\Omega_S$ itself considered as a porous region. This region $\Omega_S=\Omega\smallsetminus\Omega_F$ is seen as complementing the full domain $\Omega$, which in practice represents the computational box of the numerical simulations, and the internal fluid/solid interface is denoted $\Sigma$. We denote by $\varepsilon=\varepsilon_f=1-\varepsilon_s$ the micro-porosity field defined on $\Omega$, given $\varepsilon_f $ and $\varepsilon_s$ respectively the volume fractions of void and solid according to usual notations \cite{soulaine_mineral_2017, ETANCELIN2020103780}. This defines a micro continuum description of the porous medium such that $\varepsilon=1$ in the pure fluid region $\Omega_F$ and takes a small value in the surrounding matrix $\Omega_S$. In fact, the local micro-porosity $\varepsilon$ is assumed to have a strictly positive lower bound $\varepsilon(x,t)\geq \varepsilon_0>0$ for all $(x,t)$ in the spatiotemporal domain $\Omega\times (0, T_f)$. This lower bound $\varepsilon_0$ characterizes the residual, potentially unresolved, porosity of the porous matrix. In practice, we set throughout this article $\varepsilon_0 = 5\%$. This micro-continuum formulation relies on a two-scale representation of the sample characterized by its micro-porosity field $\varepsilon$.  

Such a two-scale description of the local heterogeneities in the carbonate rocks is appropriate to simulate the pore-scale physics and establish the governing flow and transport equations in each distinct region. Indeed, we consider the model on superficial velocity $u$ introduced and derived rigorously by Quintard and Whitaker in the late 80s \cite{QW88} and commonly used until nowadays \cite{LasseuxTiPM, golfier1, soulaine_mineral_2017, molins_simulation_2021}:
\begin{equation}
    \label{eq:superficial}
    \varepsilon^{-1}\frac{\partial\rho u}{\partial t} +\varepsilon^{-1}\nabla\cdot(\varepsilon^{-1}\rho u\otimes u) -\varepsilon^{-1}\nabla\cdot(2\mu D(u)) +\mu^*K_\varepsilon^{-1}u = f - \nabla p 
\end{equation}
along with the divergence-free condition $\nabla\cdot u = 0$. In this equation, $D(u)=(\nabla u+\nabla u^T)/2$ is the shear-rate tensor, $\mu$ is the dynamic viscosity, $p$ is the volumic pressure, $f$ the volumic driving force and $\rho$ the fluid density. The related viscosity $\mu^*$ coincides usually with the fluid viscosity $\mu$ but may be different in order to account for viscous deviations. The quantities $\rho$, $\mu$, $\mu^*$ and $f$ are assumed to be constant. In contrast, the permeability $K_\varepsilon$ refers to the micro-scale permeability and depends on the local micro-porosity field $\varepsilon$. In fact, the permeability of the micro-porous domain is modeled by the empirical Kozeny-Carman relationship \cite{Kozeny_1927, Carman_1937, Carrier2003}:
\begin{equation}
\label{eq:KC}
    K_\varepsilon^{-1} = \kappa_0^{-1} \frac{(1-\varepsilon)^2}{\varepsilon^3}
\end{equation}
where $\kappa_0$ is a coarse estimation of the reference macro-scale permeability. In this article, we consider both $K_\varepsilon$ and $\kappa_0$ as scalars, meaning we restrict ourselves to the isotropic case although this formalism can be extended to anisotropic porous media. The superficial velocity formulation \eqref{eq:superficial} defines a two-scale model that can be solved on the overall domain $\Omega$ --- using for instance penalization principles --- and retrieves the usual Navier-Stokes equation in the pure fluid region $\Omega_F$ (since $K_\varepsilon^{-1} = 0$ for $\varepsilon = 1$). At low Reynolds numbers and for highly viscous Darcian flows, equation \eqref{eq:superficial} reduces to the following Darcy-Brinkman Stokes (DBS) model:
\begin{equation}
    \label{eq:DBS}
    -\nabla\cdot(2\mu D(u)) +\mu\kappa_0^{-1}\frac{(1-\varepsilon)^2}{\varepsilon^2}u =\varepsilon(f-\nabla p), \quad \text{in} \quad \Omega
\end{equation}
where $\mu^* = \mu$ for sake of readability. In the present work, we consider this DBS equation \eqref{eq:DBS}, which is adequate in the flow regime hypothesis of low Reynolds number representative in pore-scale modeling. The DBS equation based on the superficial velocity is an efficient formalism to model the hydrodynamic in multi-scale porous media. 

The flow model \eqref{eq:DBS} needs to be complemented by transport-reaction-diffusion equations of the different species involved in the geochemical processes. These equations are derived from the mass balance of the chemical species \cite{soulaine_mineral_2017}, and can be written under the form:
\begin{equation}
\label{eq:react2}
    \frac{\partial \varepsilon \widetilde C_k}{\partial t}+\nabla\cdot(u\widetilde C_k)-\nabla\cdot\left(\alpha_k(\varepsilon)\varepsilon\nabla \widetilde C_k\right)= \widetilde R(\widetilde C_k),
\end{equation}
where $\widetilde C_k = \rho_f \overline{\omega}_{f,k} / M_k$ is a concentration per unit of fluid (following the notations introduced by Quintard and Whitaker in \cite{quintard_two-phase_1988}, and afterward by Soulaine and al. in \cite{soulaine_mineral_2017}) with $M_k$ the molar mass of the $k^\text{th}$ specie. The term $\alpha_k(\varepsilon)$ is a space-variable effective diffusion coefficient and accounts for a reduced diffusion in the surrounding porous matrix due to the tortuosity effect, which is usually quantified using Archie's law \cite{Archie_1942}:
\begin{equation}
    \alpha_k(\varepsilon)=D_{m,k}\varepsilon^\beta.
\end{equation}
In this empirical relationship, $\beta$ refers to the tortuosity index and $D_{m,k}$ to the molecular diffusion of the considered species \cite{Wakao_1962}. We finally introduce the concentration per unit of volume defined by $C_k = \varepsilon \widetilde C_k$, so that the equation \eqref{eq:react2} is written
\begin{equation}
\label{eq:react3}
    \frac{\partial C_k}{\partial t}+\nabla\cdot(\varepsilon^{-1}uC_k)-\nabla\cdot\left(D_{m,k}\varepsilon^{1+\beta}\nabla(\varepsilon^{-1}C_k)\right)= R(C_k),
\end{equation}
which is no more than superficial modeling of the chemistry. In the context of the current article, we are not interested in monitoring the dissolution products of \eqref{eq:chem_eq} (\emph{i.e} the \ce{Ca^{2+}} and \ce{ HCO3-} ions), hence we mainly focus on the concentrations of the acid and calcium carbonate species respectively denoted $C(x,t)=[\ce{H+}]$ and $C_s(x,t) = [\ce{CaCO3}]$. The solid concentration is linked to the porosity through the molar volume of calcite $\upsilon$ by the relation $ C_s  = (1-\varepsilon)/\upsilon$ with $\upsilon=36.93\, \mathrm{cm^{3}.mol^{-1}}$. In this configuration, the evolution of the acid phase (\emph{i.e} the concentration field $C$) follows the equation \eqref{eq:react3}, and the evolution of the solid phase with superficial concentration $C_s$ is given by the same equation without transport nor diffusion:
\begin{equation} 
\label{eq:react_solid}
    \frac{\partial C_s}{\partial t}= R(C).
\end{equation}
This reaction rate --- related to the chemical reaction \eqref{eq:chem_eq} --- is written \cite{Molins_Steefel_2012}: 
\begin{equation} \label{eq:react_defr}
   R(C)=-K_sA_s\gamma_{\ce{H+}} C \mathbb{1}_{\{(1-\varepsilon)>0\}}
\end{equation}
where $K_s$ is the dissolution rate constant, $A_s$ the specific reactive area, and $\gamma_{\ce{H+}}$ the activity coefficient of the acid, whose physical units are respectively $\mathrm{mol.m^{-2}.s^{-1}}$, $\mathrm{m^{-1}}$ and $\mathrm{m^3.mol^{-1}}$ (such that the chemical activity $a_{H^+} = \gamma_{\ce{H+}} C$ is dimensionless). The notation $\mathbb{1}$ refers to a characteristic or activation function and ensures the rate of the chemical reaction is non-zero only in the presence of solid minerals. Along with its boundary and initial conditions, this defines a set of partial differential equations modeling reactive flows at the pore scale \cite{Steefel_1994, Molins_Steefel_2012}: 
\begin{equation}
    \label{eq:forward_chem_pb}
    \left\{
    \begin{array}{ll}
         \ds -\nabla\cdot(2\mu D(u)) +\mu\kappa_0^{-1}\frac{(1-\varepsilon)^2}{\varepsilon^2}u =\varepsilon(f-\nabla p), &\quad \text{in} \quad \Omega\times(0,T_f) \\[4mm]
          
          \ds \frac{\partial C}{\partial t}+\nabla\cdot(\varepsilon^{-1}uC)-\nabla\cdot\left(D_m\varepsilon^{1+\beta}\nabla(\varepsilon^{-1}C)\right)= -K_sA_s\gamma_{\ce{H+}} C \mathbb{1}_{\{(1-\varepsilon)>0\}}, &\quad \text{in} \quad \Omega\times(0,T_f)\\[4mm]

         \ds\frac{\partial \varepsilon}{\partial t}= \upsilon\, K_sA_s\gamma_{\ce{H+}} C \mathbb{1}_{\{(1-\varepsilon)>0\}}, &\quad \text{in} \quad \Omega\times(0,T_f)\\[4mm]
         
        \text{+ adequate boundary and initial conditions, along with $\nabla\cdot u = 0$}
    \end{array}
\right.
\end{equation}
which is strongly coupled, since $u$ and $C$ (by means of $\varepsilon$) depend on each other. Finally, one can notice that the reactive system \eqref{eq:forward_chem_pb} is valid on the whole domain $\Omega$, whether the local state is fluid or not. In the pure fluid region, this system indeed converges toward a Stokes hydrodynamic model coupled with a standard transport-diffusion equation for the acid, with its molecular diffusion $D_m$. The overall system \eqref{eq:forward_chem_pb} defines the direct formulation of the calcite dissolution problem, following the chemical equation \eqref{eq:chem_eq}, at the pore-scale. 

Nonetheless, appropriate model calibration of the kinetic input parameters, such as the specific surface area $A_s$ or the dissolution rate constant $K_s$, that compare with experimental results remains challenging. This comes from the observation these reactive constants can span over a wide range of orders of magnitude, inducing highly different behaviors in the system. Quantifying the uncertainties on these kinetic parameters thereby appears as a necessity to provide reliable reactive flow models at the pore scale.
\subsection{X-ray microtomography limitations: toward the uncertainty quantification assessment}
\label{subsec:Xray_limits}

\begin{figure}
    \centering
    \includegraphics[width = 0.8\linewidth]{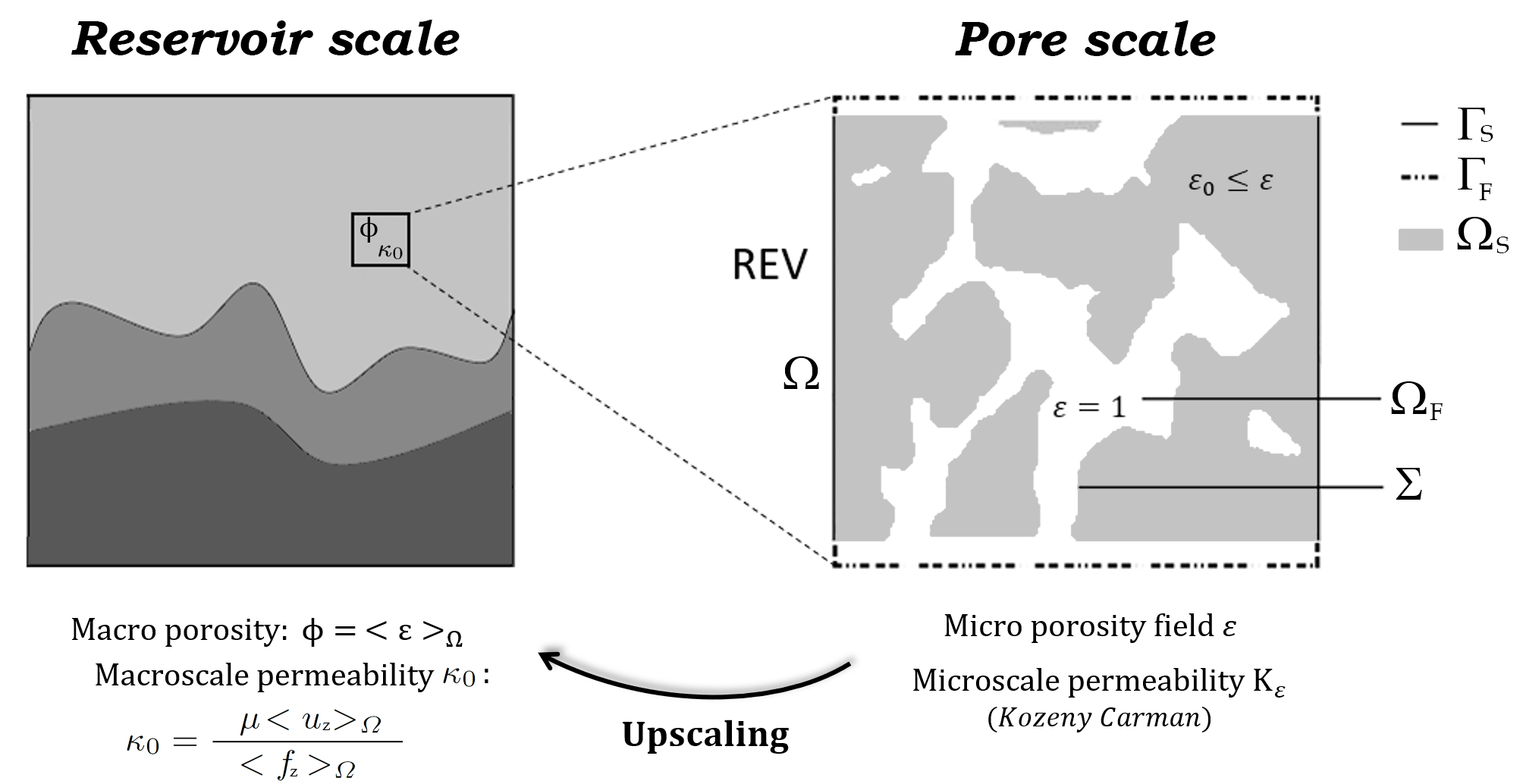}
    \caption{\textbf{From the pore-scale to the reservoir scale: an upscaling principle.} Schematic representation of a reservoir scale structure, on the left, with its inherent averaged isotropic macro-properties $\phi$ and $\kappa_0$ computed on a representative elementary volume (REV). Local description of the pore-scale heterogeneities in this REV, on the right, along with its intrinsic micro-scale properties. These are the local micro porosity field $\varepsilon$ (bounded by the least physically possible porosity $\varepsilon_0>0$) and the microscale permeability $K_\varepsilon$, based on the Kozeny-Carman relationship from equation \eqref{eq:KC}.}
    \label{fig:upscaling}
\end{figure}

Independently of the modeling aspects developed in Sect. \ref{subsec:DNS} and the choice of the numerical method used as a direct solver, pore-scale simulations are intrinsically related to X-ray microtomography (X-ray $\mu$CT). In fact, the latter provides, beforehand, scans of the complex shape geometry on a representative elementary volume (REV), defined as the characteristic minimal volume on which the microscopic variables can be averaged \cite{bachmat_concept_1987}. Pore-scale numerical simulations of reactive dynamical processes are then performed on this REV initial geometry --- which defines the domain $\Omega$ --- tracking the dynamical interface evolutions and micro-properties changes. This REV concept also allows the passage from the pore scale to the Darcy scale by referring to representative criteria of the domain in terms of averaged properties, such as the macro-porosity $\phi$, bulk permeability $\kappa_0$, and the reactive surface area $A_s$. These bulk parameters are derived from the evolving micro-structures through homogenization principles, and upscaling of the governing equations \cite{quintard_transport_1993, whitaker1999theory}, as illustrated in Fig \ref{fig:upscaling}. Indeed, at the Darcy scale of an isotropic material, the upscaled porosity and the scalar absolute permeability $\kappa_0$ are respectively defined by:
\begin{equation}
\phi=<\varepsilon>_{\Omega} \text{ and }\ 
\kappa_0=\frac{\mu <u_z>_{\Omega}}{<f_z>_{\Omega}}\simeq\frac{\mu\phi <u_z>_{\Omega_F}}{<f_z>_{\Omega_F}},
\end{equation}
using the notations introduced in Sect. \ref{subsec:DNS}, where $u_z$ and $f_z$ represent the vertical components of the velocity and driving force and where $<.>$ denotes the average on the corresponding domain.
Therefore, trustable measurement of the impact of the reactive processes on the porous medium macro-properties requires ensuring reliable quantification of the changes in micro-properties. This can be achieved under the constraint of having a fine description of the pore space, with correct knowledge of the surrounding solid matrix defined by the local micro-porosity field $\varepsilon$. An efficient representation of the porous sample at the pore scale is necessary to guarantee reliable estimation of the macro-properties evolutions along the reactive processes. 

Advances in X-ray microtomography offer such an opportunity. X-ray $\mu$CT is regarded as a powerful high-resolution imaging technique able to non-destructively determine the inner structure of a porous sample up to a characteristic scale, which defines the voxel size. The voxels are small elementary volumes (of a few $\mu m$) that compose the overall 3D reconstructed sample geometry and are identified by different grey levels characterizing the local attenuation of the material. The resulting dataset can either be segmented to separate the pore space (fluid phase) from the surrounding solid matrix or benefit from the information related to the greyscale values of the different voxels. The segmented images lend themselves to numerical simulations that require an explicit representation of the fluid-solid interfaces (\emph{e.g.} Lattice-Boltzmann \cite{ahkami_lattice-boltzmann_2020}), unlike the Darcy-Brinkman Stokes formulation presented in Sect. \ref{subsec:DNS}, which incorporates the voxel greyscale values. Indeed, these grey-level shades, depicting the material local attenuation, are correlated to the porosity field description $\varepsilon$ and can be taken into account in the DBS model through the equation \eqref{eq:DBS}. This introduces Digital Rock Physics applications as the joint use of high-resolution X-ray computed microtomography and advanced simulation techniques to characterize, \emph{inter alia}, the rock petrophysical properties and their evolutions \cite{andra_digital_2013, andra_digital_2013-1}. Pure imaging alternatives readily regard the resulting dataset to derive the sample's effective physical properties (porosity, permeability, dispersivity...) but also geochemical rates and mineral reactivity in dynamical processes \cite{smal_automatic_2018, noiriel_direct_2019}. Therefore, X-ray microtomography is both a complementary means to numerical modeling at the pore scale and a fundamental imaging process on its own to study the \ce{CO2} storage implications on porous material.

However, limitations in the $\mu$CT imaging process may affect the determination of medium effective properties, and query the reliability of the predictive models based on these inputs. In fact, several imaging artifacts exist and disrupt the efficient description of the pore space morphology. Firstly, the finite resolution of the $\mu$CT pipeline is challenging, as the interfaces appear blurry and do not manifest themselves as sharp intensity steps in the images, but rather as gradual intensity changes spanning over several voxels \cite{Schlueter2014}. Actually, the local attenuation signal within a voxel is influenced by the material heterogeneity in its neighborhood, then the resulting grey scale value represents averaged properties: this is known as the partial volume effect \cite{Tomo}. This phenomenon is also involved when morphological features of interest are smaller than the characteristic voxel size, resulting in unresolved micro-porosity or roughness of the pore space walls. Quantifying sub-resolution porosity, which is a prevalent imaging artifact in $\mu$CT, and measuring its impact on numerical modeling and simulation is identified as critical for geosciences applications \cite{LIN2016306, smal_automatic_2018, CARRILLO2022}. Such an issue is well-known and arises from a compromise between the sample volume being investigated and the scan resolution. For porous media covering a wide range of pore scales, this trade-off can readily result in voxel sizes that are not able to capture fully resolved morphological features of the pore space. Finally, in the presence of sharp density transitions, the different refraction index at either side of the interface furthermore leads to so-called edge enhancement which manifests itself as an over- and undershoot of the grey level immediately next to the interface \cite{Banhart2008}. Consequently, the position of the material interface is prone to uncertainty, in addition to the roughness of the pore space walls, and therefore results in an approximation of the true morphology. 

While the mentioned effects can be minimized, they cannot be eliminated and add uncertainties to the estimation of the effective properties, the characterization of the void/solid interfaces, and the reliability of the numerical models. In addition, the accuracy of X-ray $\mu$CT images is challenged by additional artifacts coming from both inherent physical and technical limitations \cite{KETCHAM2001381}. It includes, among them, instrumental noise, beam hardening \cite{Wildenschild2013} which results in cupping (an underestimation of the attenuation at the center of the object compared to its edges) or drag/streak appearances (due to an underestimation between two areas of high attenuation), beam fluctuations along the scanning process and scatter radiations coming from the object and/or the detector. These variations can manifest as noise, ring or streak artifacts, and halos that are often hard to distinguish from real features and therefore hinder the identification of sample heterogeneities at multiple scales. Ubiquitous limiting factors remain in the X-ray $\mu$CT imaging process, and the assessment of their related uncertainties is fundamental to developing more accurate predictive models.

\subsection{Dynamical microtomography: mineral reactivity and imaging morphological uncertainties}
\label{subsec:Xray_UQ}

Accounting for the $\mu$CT morphological uncertainties and sub-resolution porosity, introduced in Sect. \ref{subsec:Xray_limits}, is essential in providing reliable pore-scale simulations of reactive flows. This is of primary importance when considering risk assessment and predicting meaningful evolutions of the rock macro-properties under geochemical effects. The study of these overall X-ray imaging limitations, therefore, raises concerns in the research community, and investigations are conducted on quantifying their implications on the effective properties. In fact, sub-resolution porosity may lead to a misleading estimation of the pore-space connectivity that disrupts the flow description within the REV and induces significant deviations in the computed permeability. Several modeling approaches, mainly based on upscaling principles, aim at quantifying these deviations. They cover DBS formulation altogether with the Kozeny-Carman equation \eqref{eq:KC}, which estimates the permeability of the micro-porous domain through a heuristic relation with the residual micro-porosity \cite{Soulaine_16}. However, in the absence of prior knowledge of this unresolved residual porosity, the setting of the micro-porous permeability becomes controversial. Alternatives rely on appropriate boundary conditions to model the unresolved features and wall roughness through slip-length formalism, and range from theoretical implications \cite{LasseuxTiPM, PhysRevFluids_Lasseux} to the practical computation of the permeability deviations on real 3D $\mu$CT scans \cite{perez_deviation_2022}. Apart from the modeling quantification of the effective properties uncertainties, experimental and imaging approaches are developed to resolve the sub-resolution porosity. This involves differential imaging techniques based on comparisons between several enhanced contrast scans \cite{LIN2016306}, statistical studies based on $\mu$CT histograms \cite{zhuang_novel_2022}, or deep learning methodologies such as Convolutional Neural Networks (CNN) and Generative Adversarial Networks (GAN) that provide super-resolved segmented images \cite{alqahtani_super-resolved_2022, YANG2022104411}. Overall, uncertainty quantification of the X-ray $\mu$CT limitations either relies on appropriate mathematical modeling with the estimation of computed deviations or experimental approaches based on image treatment analysis of the $\mu$CT scans. 

The reliability of pore-scale modeling related to X-ray $\mu$CT scans is questioned due to its inherent imaging limitations and morphological uncertainties. At the same time, proper assessment of the kinetic parameters in dynamic phenomena, including mineral reactivity and reactive surface area, also raise challenges. Actually, mineral reactivity is a critical parameter to account for in many geosciences applications though discrepancies of several orders of magnitude can be found in the literature \cite{black_rates_2015, DUTKA2020119459}. However, these parameters are usually regarded as input in the numerical models and eventually tuned to aggregate experimental results. Providing reliable uncertainty estimates on these kinetic parameters is, therefore, of great interest to provide trustable pore-scale reactive simulations. Such concern has received attention over the past decades, and considering dynamic imaging processes subsequently appears as a necessity. Several experimental works have already focused on 4D imaging techniques of carbonate dissolution to provide fundamental information on mineral reaction rates \cite{Noiriel4D}. These kinetic characterization studies mostly rely on voxel-to-voxel subtraction of consecutive images in order to quantify the change of greyscale values, hence the evolution of the dissolution process and calcite retreat. This is referred to, in the literature, as differential imaging techniques and has been investigated for different imaging techniques such as X-ray $\mu$CT and atomic force microscopy (AFM) \cite{LIN2016306, siena_statistical_2021}. These approaches enable to capture heterogeneous spatial distributions of calcite dissolution rates through successive real-time measurements. They provide local maps of mineral reactivity at the crystal surfaces and quantification of their morphological evolutions \cite{siena_statistical_2021, noiriel_direct_2019}. Menke et al. \cite{menke_dynamic_2015} also performed \emph{in situ} time-resolved experiments of carbonate dissolution under reservoir conditions (in terms of pressure and temperature) to derive averaged reaction rates and evaluate dynamical changes in the effective properties. Investigation of mineral surface reactivity is another challenging concern to ensure reliable calibration of pore-scale models for \ce{CO2} dissolution, and is usually achieved through dynamical $\mu$CT experiments.

Nonetheless, dealing with dynamical $\mu$CT images brings its own challenges \cite{Zhang_challenges}. In addition to the unresolved features, dynamical imaging of chemical processes requires making a compromise between the acquisition time and the image quality. Indeed, capturing fast-dissolution processes, for instance, imposes short acquisition times and could result in highly noisy data since statistically, the number of photons reaching the detector would be reduced. In such a case, differential imaging makes it difficult to distinguish between true morphological changes and the derivation of highly noisy data. On top of that, any additional movement in the sample, not related to the dissolution process but rather resulting from instrumentation artifacts, makes it challenging to work with dynamical samples only to characterize $\mu$CT errors and uncertainty. Indeed, Zhang and al. \cite{Zhang_challenges} identified on a Bentheimer sampler that about 32\% of the voxels have at least a 2\% difference in greyscale values between two consecutive fast scans. These differences are not physical-based variations but rather intrinsic uncertainties measurements. They also show that such artifacts' uncertainties can be reduced by using slower acquisition time, though this is not always feasible to capture fast-dynamical processes. Time-resolved experiments of dynamical processes can provide insights into kinetic dissolution rates, though this also suffers from imaging limitations that can lead to misleading estimations.

Inferring reliable mineral reactivity from dynamical microtomography and quantifying imaging morphological uncertainties are identified as the major issues that can bias the determination of evolving petrophysical properties in geological applications. Current methodologies addressing these problems mainly fall into two categories: on one side, purely model-related approaches based on static $\mu$CT scans and upscaling principles, and on the other side, image treatment analysis relying on experimental static or dynamical images. Nonetheless, neither morphological uncertainty nor reaction rate quantification has been investigated from the perspective of coupling physics-based models with data-driven techniques. To the best of our knowledge, the development of data assimilation approaches on pore-scale imaging problems that combine dynamical microtomography and physical regularization induced by the PDE model of reactive processes is the main novelty of the present manuscript. The motivation for this formulation lies in its joint ability to infer mineral reactivity parameters and quantify the residual micro-porosity generated by unresolved features in the microtomography imaging process. Therefore, we assert that a proper balance between dynamical microtomography imaging and their PDE-based physical formulation could provide insights into the uncertainty quantification issues related to reactive pore-scale modeling. In this direction, we propose a novel methodology that uses a physics-based dissolution model as regularization constraints to dynamical data-driven microtomography inference. This aim at quantifying both the uncertainties on kinetic parameters to perform reliable model calibration and the morphological imaging uncertainty on the unresolved micro-porosity field. 



\section{Direct and inverse problem setup}
\label{sec:Pb_setup}
This section is dedicated to setup the dimensionless versions of the dissolution PDE model for direct and inverse problems. We establish the main differences in their dimensionless formulations and define the modeling assumptions used in the present article.

\subsection{Usual dimensionless formulation of the direct problem}
\label{subsec:Adim_direct}
The overall calcite dissolution PDE system, defined in equation \eqref{eq:forward_chem_pb}, can model a wide range of dissolution regimes and patterns characterized by well-established dimensionless numbers. By setting $x^*=x/L$ and $t^*=t D_m/L^2$ and following the notations of Sect. \ref{subsec:DNS}, one can introduce the so-called Peclet and Reynolds numbers
\begin{equation}
\Pe=\bar u L/D_m \ \text{ and }\ \Re=\rho \bar u L/\mu
\end{equation}
where $\bar u$ and $L$ are respectively the characteristic velocity and length of the sample. In the context of pore-scale simulations, the inertial effects become negligible compared to viscous forces due to low Reynolds numbers --- typically we have the assumption $\Re \ll 1$ throughout this article. Regarding the chemical reactions, two dimensionless numbers are defined: the catalytic Damköhler number denoted $\Da2$ and its inherited convective number $\Da1$, expressed as
\begin{equation}
\label{eq:Da2_Da1}
\Da2=\frac{K_s A_s \gamma_{\ce{H+}} L^2}{D_m} \quad\text{ and }\quad \Da1=\frac{\Da2}{\Pe}=\frac{K_sA_s\gamma_{\ce{H+}} L}{\bar u}.
\end{equation}
The characteristic length $L$ is usually related to average pore throat diameters or $L^2$ can be set as the surface of a section divided by the average number of grains (\emph{e.g.} see \cite{Hume2021} for practical cases). Otherwise, it is possible to set the characteristic length of the problem as $L=\sqrt{\kappa_0}$, provided an experimental or numerical estimation of $\kappa_0$ \cite{soulaine_mineral_2017}. All these dimensionless numbers are meaningful in direct dissolution problems to qualify the different dominant regimes in terms of diffusion, reaction, and advection.  

Using the dimensionless variables $(x^*, t^*)$, the normalized concentration $C^* = C/C_0$ and velocity $u^* = u/\bar u$, one finally gets the dimensionless formulation of the overall reactive flow system \eqref{eq:forward_chem_pb} on the dimensionless spatiotemporal domain $\Omega^*\times(0,T_f^*)$. This leads to the usual PDE model:
\begin{equation}
    \label{eq:dimless_model}
    \left\{
    \begin{array}{ll}
         \ds -\Delta u^* + L^2\kappa_0^{-1}\frac{(1-\varepsilon)^2}{\varepsilon^2}u^* =\varepsilon(f^*-\nabla p^*), &\quad \text{in} \quad \Omega^*\times(0,T_f^*) \\[4mm]
          
         \ds \frac{\partial C^*}{\partial t^*} +\Pe\, \nabla\cdot(\varepsilon^{-1}u^*C^*) -\nabla\cdot\left(\varepsilon^{1+\beta}\nabla(\varepsilon^{-1}C^*)\right)= -\Da2 C^* \mathbb{1}_{\{(1-\varepsilon)>0\}}, &\quad \text{in} \quad \Omega^*\times(0,T_f^*) \\[4mm]

         \ds \frac{1}{\upsilon C_0}\frac{\partial \varepsilon}{\partial t^*}= \Da2 C^* \mathbb{1}_{\{(1-\varepsilon)>0\}}, &\quad \text{in} \quad \Omega^*\times(0,T_f^*) \\[4mm]
         
        \text{+ adequate boundary and initial conditions, along with $\nabla^*\cdot u^* = 0$}
    \end{array}
\right.
\end{equation}
obtained by means of multiplying the hydrodynamic DBS equation by $L/\rho {\bar u}^2$ and the chemical equations in \eqref{eq:forward_chem_pb} by $L^2/C_0D_m$. The notations $f^*$ and $p^*$ in the dimensionless DBS equation are defined by $f^*=fL^2/(\mu{\bar u})$ and $p^*=p L/(\mu{\bar u})$, and finally $C_0$ is a characteristic constant for the acid concentration field. This PDE system defines the overall dimensionless formulation of the direct problem of calcite dissolution. 


\subsection{Modeling assumptions on the direct and inverse problems}
\label{subsec:model_assumptions}
In the applications Sect. \ref{sec:Results1D} and \ref{sec:Results}, we consider inverse problems in the dissolution process of calcite cores with heterogeneous porosity levels for 1D and 2D spatial configurations. Although the 1D+Time test case is purely synthetic and aims to validate the method developed in Sect. \ref{sec:Method}, the 2D+Time application addresses a more realistic problem that can be applied to isotropic porous samples. Several modeling assumptions are, though, made to address both the reactive direct and inverse problems. These modeling assumptions are detailed hereafter and determine the dissolution regime considered in the applications. 

The dimensionless numbers $\Re$ and $\Pe$ are common to both the direct and inverse formulations and respectively establish the viscosity-dominated regime and the convective or diffusive transport regime. At the initial state in the dynamic imaging process, we assume that the porous medium is completely saturated with the acid by capillary effect. The amount of reactant at the pore interface is initially homogeneously distributed, and therefore we expect, at first, a cylindrical dissolution regime of the calcite core with spherical symmetry. Subsequently, the dissolution process may deviate from this cylindrical pattern due to local heterogeneities in the micro-porosity field $\varepsilon$. We also suppose a low Peclet hypothesis $\Pe \ll 1$, so that the reactant diffusion is dominant over the advection phenomena resulting in more homogeneous dissolution rates at the interface (\emph{e.g.} see \cite{soulaine_mineral_2017} for the dissolution regimes characterization). In this sense, continuous acid injection is maintained at a given fluid flow rate to ensure a diffusive-dominated regime for the dissolution. Consequently, we neglect the advection effects in the present article and focus on the following reaction-diffusion system:

\begin{equation}
    \label{eq:forward_chem_pb_simple_adim}
    \left\{
    \begin{array}{ll}       
          \ds \frac{\partial C^*}{\partial t^*}-\nabla\cdot\left(\varepsilon^{1+\beta}\nabla(\varepsilon^{-1}C^*)\right)= -\Da2 C^* \mathbb{1}_{\{(1-\varepsilon)>0\}}, &\quad \text{in} \quad \Omega^*\times(0,T_f^*) \\[4mm]

         \ds \frac{1}{\upsilon C_0}\frac{\partial \varepsilon}{\partial t^*}= \Da2 C^* \mathbb{1}_{\{(1-\varepsilon)>0\}}, &\quad \text{in} \quad \Omega^*\times(0,T_f^*) \\[4mm]
         C^* = 1, &\quad \text{on} \quad \partial\Omega^*\times(0,T_f^*)  \\[2mm]
         C^*(x,0) = C_{\mathrm{init}}(x)/C_0:=C^*_{\mathrm{init}}, &\quad \text{in} \quad \Omega^*\times\{0\}
    \end{array}
\right.
\end{equation}
written in its dimensionless form with the normalized concentration field $C^* = C/C_0$. The continuous acid injection is modeled through non-homogeneous Dirichlet boundary conditions on $C^*$, with the characteristic constant $C_0$ chosen as the value of the Dirichlet boundary conditions on $C$. The initial condition on the micro-porosity field $\varepsilon$ arises from the dry microtomography scan or the initial synthetic porous medium. 
Ultimately, we obtain a PDE model driven by one dimensionless number, namely the catalytic Damköhler number, characterizing the ratio of the reaction rate over the diffusion effects. The system \eqref{eq:forward_chem_pb_simple_adim} is, therefore, consistent with the standard dimensionless formulation of the direct problem, subject to a diffusive-dominated transport regime. 

However, we merely cannot consider the dimensionless temporal variable $t^*=t D_m/L^2$ in a reactive inverse problem as it strongly depends on molecular diffusion $D_m$, which is among the unknown kinetic parameters to be estimated. In the next section, we focus on the challenge arising from the dimensionless formulation of a reactive inverse problem in the context of calcite dissolution.
\subsection{Dimensionless inverse problem on calcite dissolution}
\label{subsec:Adim_inverse}

In this article, we address pore-scale imaging inverse problems in dissolution processes. We aim to recover and quantify uncertainties both on the micro-porosity field description $\varepsilon$ and the reactive parameters involved in the diffusion-reaction system. Among these inverse kinetic parameters, one can find the molecular diffusion $D_m$, the tortuosity index $\beta$, the dissolution rate constant $K_s$, and even the specific surface area $A_s$ --- usually estimated on the dry $\mu$CT scan. Consequently, these parameters -- in particular $D_m$ -- cannot be used for the non-dimensionalization of the model since they are to be determined. Apart from special considerations of the tortuosity index of the sample, the other inverse parameters characterize the dissolution regime of the dynamical $\mu$CT experiment. In this sense, they provide insight into the physical catalytic Damköhler number $\Da2$, though the direct dimensionless formulation \eqref{eq:forward_chem_pb_simple_adim} is inappropriate for an inverse problem. The dimensionless temporal variable $t^*$ in the PDE system \eqref{eq:forward_chem_pb_simple_adim} is, indeed, closely related to the unknown molecular diffusion, compromising its application to inverse modeling. Establishing the dimensionless formulation of the inverse dissolution problem is not straightforward and therefore requires a different dimensionless time.

In the inverse problem, we consequently introduce the new temporal variable 
\begin{equation}
\label{t_star}
    \ds t^* = \frac{tD_{\mathrm{ref}}}{L^2},
\end{equation}
where $D_{\mathrm{ref}}$ is a scaling factor of the dimensionless formulation for the chemical kinetics. This scaling factor can also be defined as $D_{\mathrm{ref}} = L^2/T$ introducing $T$ the characteristic time for the dimensionless problem, which is not the physical characteristic time for the diffusion since the latter is unknown. In practice, we can rely on a rough estimation of physical dissolution time $T_f$ --- determining the dynamical process end --- and a given dimensionless final time --- usually $T_f^* = 1$ --- to set the factor $D_{\mathrm{ref}}$. The estimations of this scaling parameter will be detailed on a case-by-case basis throughout the applications developed in Sect. \ref{sec:Results1D} and \ref{sec:Results}. Using the new dimensionless variables $(x^*, t^*)$, the normalized concentration $C^* = C/C_0$ along with the definition of $D_{\mathrm{ref}}$, one can obtain the dimensionless formulation of the reaction-diffusion system in the context of inverse modeling, which leads to:
\begin{equation}
    \label{eq:inverse_chem_adim}
    \left\{
    \begin{array}{ll}       
          \ds \frac{\partial C^*}{\partial t^*}-D_m^*\nabla\cdot\left(\varepsilon^{1+\beta}\nabla(\varepsilon^{-1}C^*)\right)= -\Da2^* C^* \mathbb{1}_{\{(1-\varepsilon)>0\}}, &\quad \text{in} \quad \Omega^*\times(0,T_f^*) \\[4mm]

         \ds \frac{1}{\upsilon C_0}\frac{\partial \varepsilon}{\partial t^*}= \Da2^* C^* \mathbb{1}_{\{(1-\varepsilon)>0\}}, &\quad \text{in} \quad \Omega^*\times(0,T_f^*) \\[4mm]
         C^* = 1, &\quad \text{on} \quad \partial\Omega^*\times(0,T_f^*)  \\[2mm]
         C^*(x,0) = C^*_{\mathrm{init}}, &\quad \text{in} \quad \Omega^*\times\{0\}.
    \end{array}
\right.
\end{equation}
In reactive inverse problems, we thus obtain a PDE model driven by two dimensionless numbers denoted $\Da2^*$ and $D_m^*$ which are defined as:
\begin{equation}
    \label{eq:dim_numb_inv}
\ds \Da2^* := K_sA_s\gamma_{\ce{H+}} T = \frac{K_sA_s\gamma_{\ce{H+}} L^2}{D_{\mathrm{ref}}} \ \text{ and }\ \ds D_m^* := \frac{D_m T}{L^2} = \frac{D_m}{D_{\mathrm{ref}}}.
\end{equation}
Finally, the physical Damköhler number corresponding to the dynamical $\mu$CT experiment is recovered as the a-posteriori ratio $\Da2=\Da2^*/D_m^*$. From now on, we consider this dimensionless formalism and forget the star notation on the differential operator, domains, and field descriptions for the sake of readability. This results in the following inverse dimensionless PDE system: 
\begin{equation}
    \label{eq:inverse_chem_adim_Vf}
    \left\{
    \begin{array}{ll}       
          \ds \frac{\partial C}{\partial t}-D_m^*\nabla\cdot\left(\varepsilon^{1+\beta}\nabla(\varepsilon^{-1}C)\right)= -\Da2^* C \mathbb{1}_{\{(1-\varepsilon)>0\}}, &\quad \text{in} \quad \Omega\times(0,T_f)\\[4mm]

         \ds \frac{1}{\upsilon C_0}\frac{\partial \varepsilon}{\partial t}= \Da2^* C \mathbb{1}_{\{(1-\varepsilon)>0\}}, &\quad \text{in} \quad \Omega\times(0,T_f)\\[4mm]
         C = 1, &\quad \text{on} \quad \partial\Omega\times(0,T_f) \\[2mm]
         C(x,0) = C_{\mathrm{init}}, &\quad \text{in} \quad \Omega\times\{0\}
    \end{array}
\right.
\end{equation}
with $\Da2^*$ and $D_m^*$ the inverse parameters to estimate, and $C_0$ and $\upsilon$ constant parameters. The tortuosity index $\beta$ is either set through a-priori estimation on the porous sample, modeling through the empirical Archie law, or regarded as an additional inverse parameter. Especially, the index $\beta=1$ is often considered for porous media with strong pore connections \cite{Wakao_1962, soulaine_mineral_2017} although practical upscaling of the diffusion can result in intermediate index values \cite{Hume2021}.

In addition to inferring the reactivity parameters $\Da2^*$ and $D_m^*$, we aim to estimate the spatial variability on the porosity field $\varepsilon$. In this sense, we develop a data assimilation approach on pore-scale imaging that combine dynamical $\mu$CT experiments of calcite dissolution and physical regularization induced by the dimensionless PDE model \eqref{eq:inverse_chem_adim_Vf}. It benefits from the joint ability to quantify the ranges of mineral reactivity and the residual micro-porosity generated by unresolved features in the microtomography imaging process. This formulation also relevantly combines the advantages of experimental and modeling approaches and overcomes their own limitations. On the one hand, the dissolution process observation will bring insights into the unresolved morphological features and lead to a better characterization of the sample's initial state. On the other, the PDE model regularization can efficiently substitute the differential imaging approach, which is controversial for fast-dynamical processes subject to poor imaging quality. Therefore, one can address mineral reactivity inference for highly noisy dynamical $\mu$CT resulting from the compromise between scan quality and time resolution, as introduced in Sect. \ref{subsec:Xray_UQ}. The major challenge of this data assimilation formulation, though, relies on the PDE constraint for the concentration field as the $\mu$CT experiments do not provide information on the flow, transport, or diffusion of the chemical reactant. In the reactive inverse problem, the acid concentration is thus a latent field whose only the dimensionless boundary conditions are known in equation \eqref{eq:inverse_chem_adim_Vf} through the normalizing constant $C_0$. In the next section, we will develop the methodology adopted to solve such a dissolution inverse problem, accounting for all the established modeling assumptions.


\section{Bayesian Physics-Informed Neural Networks in pore-scale imaging: concepts and methods}
\label{sec:BPINNs_Method}

Developing efficient data assimilation techniques is crucial to perform extensive parameter estimations, uncertainty quantification, and improving the reliability of direct pore-scale predictions. In particular, inverse problems are often subject to various sources of uncertainty that need to be quantified to ensure trustable estimations. This includes approximate model accuracy whose reliability can be questioned, with sparse or noisy data exhibiting measurement variability. Integrating physical principles, such as conservation laws or PDE models, in these inverse problems can though compensate for the lack of massive or accurate measurements through additional regularization constraints \cite{meng_multi-fidelity_2021}. At the same time, embedding these physical regularizations allows addressing model accuracy in the total uncertainty quantification, especially when misleading a-priori uncertainty is assumed on the physical constraints \cite{psaros_uncertainty_2023}. Therefore, the combination of physics-based and data-driven methods offers an efficient alternative to overcome the limitations of both purely data-driven or purely modeling approaches. This has established data-driven inference as a complementary partner to theory-driven models in data assimilation and inverse modeling incorporating uncertainty quantification.

\subsection{Uncertainty Quantification in coupled physics-based and data-driven inverse problems}
\label{subsec:BPINNs}

Several approaches were developed to address uncertainty concerns in the context of data assimilation. These uncertainty quantification problems either require stochastic PDE models \cite{BALTAS20191, Stochastic_PDE} --- also used in sensitivity analysis --- or probabilistic approaches such as Markov Chain Monte Carlo (MCMC) methods. The latter can be used in the Bayesian Inference framework to sample from a target posterior distribution, though this usually requires numerous evaluations of the forward PDE model. In this sense, developing efficient MCMC methodologies remains challenging since repeatedly solving a complex coupled PDE system is computationally expensive and therefore can quickly become prohibitive for uncertainty assessments. These computational concerns have motivated the emergence of surrogate models in Bayesian Inference to speed up the forward model evaluation. This covers methods ranging from Polynomial Chaos Expansions \cite{MARZOUK20091862, yan_adaptive_2019} which rely on a representation of the physical model by a series of low-order polynomials of random variables, to neural network proxies \cite{Yan_Adaptive_2020, ALBERTS2023112100}. Both approaches present the advantage of creating a surrogate model that can be evaluated inexpensively compared to solving the forward problem through usual direct numerical simulations. Nonetheless, Polynomial Chaos expansions suffer from truncation errors due to the low order of the polynomials yielding inaccurate estimates of the posterior distributions \cite{LU2015138}. On the contrary, deep learning methods have shown effectiveness in building surrogate models for a wide range of complex and non-linear PDEs encoding the underlying physical principles. Developing fast surrogate models based on machine learning has garnered increasing interest in accelerating Bayesian inference for a wide range of scientific applications \cite{delia_machine_2022, coheur_bayesian_2023}.


A popular framework in deep learning integrating both physics regularization, measurement data, and uncertainty estimates are Bayesian Physics-Informed Neural Networks (BPINNs)~\cite{yang_b-pinns_2021, linka_bayesian_2022}. BPINNs benefit from the combined advantages of neural network structures in building parameterized surrogate models based on physical principles and Bayesian inference standards in integrating uncertainty quantification. Introducing the Bayesian neural network parameters $\theta \in \mathbb{R}^d$ building the surrogate model and the inverse parameters of the PDE model $\mathcal{P}_{\mathrm{inv}}\in\mathbb{R}^p$, we define the joint set of unknown parameters as $\Theta = \{\theta, \mathcal{P}_{\mathrm{inv}}\}$. The BPINN formulation aims to explore the posterior distribution of $\Theta$
\begin{equation}
    \label{post_dist_Bayes}
    P(\Theta |\mathcal{D}, \mathcal{M}) \propto  P(\mathcal{D}| \Theta) P(\mathcal{M}| \Theta ) P(\Theta) 
\end{equation} 
given some measurement data $\mathcal{D}$ and a presumed model $\mathcal{M}$ with unknown parameters. The posterior distribution expression \eqref{post_dist_Bayes} basically involves a likelihood term $P(\mathcal{D}| \Theta)$ evaluating the distance to the experimental data, a PDE-likelihood term $P(\mathcal{M}| \Theta )$ characterizing the potential modeling discrepancies, and a joint prior distribution $P(\Theta)$. Through a marginalization process, the posterior distribution \eqref{post_dist_Bayes} on the parameters $\Theta$ then transfers into a posterior distribution of the predictions, also called a predictive Bayesian Model Average (BMA) distribution (\emph{e.g.} see \cite{wilson_bayesian_2020}): 
\begin{equation}
    \label{BMA}
P(y|x,\mathcal{D}, \mathcal{M}) = \int P(y|x,\Theta)P(\Theta|\mathcal{D}, \mathcal{M}) \mathrm{d}\Theta 
\end{equation}
where $x$ and $y$ respectively refer to the input (\emph{e.g.} spatial and temporal points) and output (\emph{e.g.} field prediction of the micro-porosity) of the neural network. The BPINN formulation hence provides a predictive distribution \eqref{BMA} of the quantities of interest (QoI), such as the output micro-porosity field, as well as posterior distributions over the model inverse parameters $\mathcal{P}_{\mathrm{inv}}$. Sampling from the posterior distribution \eqref{post_dist_Bayes} is achieved through MCMC methods, which efficiently combine with fast surrogate models based on deep learning. In particular, one of the most popular MCMC schemes for BPINNs is Hamiltonian Monte Carlo (HMC), which provides a particularly efficient sampler for high-dimensional inference problems~\cite{betancourt_conceptual_2018}. In addition to theoretical analyses, the HMC-BPINNs formulation also demonstrates numerical performances on both forward and inverse problems \cite{yang_b-pinns_2021}. BPINNs with the HMC sampler appear as an efficient data-assimilation alternative coupling physics-based with data-driven approaches, and incorporating intrinsic uncertainty quantification.

The HMC sampler, in particular, introduces the dynamics of a fictive system composed of the unknown parameters $\Theta$ --- regarded as particle positions in the physical analogy --- and auxiliary momentum variables $r$ --- regarded as particle velocities. It describes a conservative Hamiltonian system whose energy denoted $H(\Theta, r)$ is the sum of a potential energy $U(\Theta)$ which characterize the inverse problem formulation and a kinetic energy $K(r)$ accounting for momentum perturbations. The latter enables the sampler to diffuse across several energy levels and hence results in an efficient exploration of the joint posterior distribution $\pi(\Theta, r)$ in the phase space, defined as follows: 
\begin{equation}
    \label{joint_dist}
    \pi(\Theta, r) \sim \mathrm{e}^{-H(\Theta, r)}.
\end{equation}
The potential energy definition relies on a Bayesian probabilistic formulation of the inverse problem such that it depends on the posterior distribution \eqref{post_dist_Bayes} by the relation $U(\Theta) = -\mathrm{ln} P(\Theta |\mathcal{D}, \mathcal{M})$. Along with a Euclidean-Gaussian assumption for the kinetic energy (\emph{e.g.} see \cite{betancourt_conceptual_2018} or \cite{PEREZ2023112342}), this ensures that the marginal distribution of $\Theta$ provides immediate samples of the target posterior distribution:
\begin{equation}
    \label{post_dist_potential}
    P(\Theta |\mathcal{D}, \mathcal{M})\sim \mathrm{e}^{-U(\theta)}.
\end{equation} 
Efficient exploration of the joint distribution $\pi(\Theta, r)$ in the phase space hence projects to samples of the target distribution \eqref{post_dist_Bayes} and then provides predictive BMA distributions on the QoI given by equation \eqref{BMA}. The successive samples $(\Theta,r)$ are generated by solving for the Hamiltonian dynamical system for the frictionless fictive particle of positions $\Theta$
\begin{equation}
    \label{Ham_dyn_sys}
    \left\{\begin{array}{l}
    \mathrm{d}\Theta = \mathbf{M}^{-1}r\,\mathrm{d}t \\
    \mathrm{d}r = -\nabla U(\Theta) \mathrm\,{d}t, 
    \end{array}\right.
\end{equation}
through a symplectic integrator, such as the Störmer-Verlet also known as the leapfrog method. This account for a deterministic exploration of specific energy level sets --- since the Hamiltonian energy is theoretically preserved by symplectic integrators --- while the kinetic energy, through momentum sampling, enables a stochastic exploration between the energy levels. The HMC-BPINN formulation ensures efficient sampling of the target posterior distribution thanks to the description of a conservative Hamiltonian system related to the inverse problem description.  

Overall, the potential energy term can be expressed under the general form (see \cite{PEREZ2023112342} for detailed development of this weighted multi-potential energy):
\begin{equation}
\label{multi_pot}
    \ds U(\Theta) = \sum_{k=0}^{K} \lambda_k \mathcal{L}_k(\Theta) + \lambda_{K+1}\|\Theta\|^2 
\end{equation}
where $ L_k = \lambda_k\mathcal{L}_k$ refers to the weighted $k^{th}$ objective term, either corresponding to data-fitting log-likelihood or PDE regularization tasks. We assume here that the prior distribution on the set of parameter $\Theta$ follows a Gaussian distribution such that $P(\Theta) \sim\mathcal{N}(0, I_{p+d})$. The weights $\lambda_k$ are positive parameters integrating the various sources of uncertainties. Indeed, the deterministic PDE model is completed by stochastic representations of the model discrepancy, and the data-fitting likelihood is itself supplemented by stochastic modeling of the experimental noise, both affecting the weights $\lambda_k$. In this sense, a HMC-BPINN intends to capture and estimate the various sources of uncertainties whether aleatoric --- arising from variability or randomness in the observations like sensor noise --- or epistemic --- caused by imperfect modeling hypothesis or ignorance in the model adequacy. Automatic management of these uncertainties, though, remains challenging as this relies on the appropriate setting of the critical weighting parameters $\lambda_k$ arising from the expression of the multi-potential energy \eqref{multi_pot}. Although some of these parameters --- mainly the noise estimation --- can be adjusted \emph{offline} with pre-trained Generative Adversarial Networks (GAN) as proposed by Psaros et al. in \cite{psaros_uncertainty_2023}, proper estimation of these weights is crucial to ensure robust uncertainty quantification. Unsuitable choices of these weights can lead to biased predictions and pathological behaviors of the HMC-BPINN sampler, especially in the context of complex real-world Bayesian inference involving multi-objective, multiscale, and stiffness issues. This needs the development of data assimilation strategies that robustly address these issues to achieve reliable uncertainty quantification in inverse problems. 

\subsection{A robust adaptive weighting sampling strategy for complex real-world Bayesian inference}
\label{subsec:AW-HMC}


The Bayesian Physics-Informed Neural Network paradigm offers the opportunity to query altogether the confidence in the predictions, the estimations of inverse parameters, and the model adequacy in inverse problems incorporating uncertainty quantification. Despite their effectiveness, BPINNs can be difficult to use correctly in complex real-world Bayesian inference as they are prone to a range of pathological behaviors. These instabilities arise from the multi-potential energy term \eqref{multi_pot} in multi-objective inverse problems which likely involve conflicting tasks or multiscale issues. In particular, such a multi-potential energy directly translates to a weighted multitask posterior distribution for which achieving successful and unbiased sampling is challenging. Ensuring robust Bayesian inference in this context hinges on properly estimating the distinct task weights. Indeed, unsuitable choices of the weights $\lambda_k$ in \eqref{multi_pot} can result in biased predictions, vanishing task behavior, or substantial instabilities in the Hamiltonian conservation. This can even prevent the sampler from identifying the highest posterior probability region, namely the Pareto front neighborhood, corresponding to predictions that correctly balance all the different tasks. While manual calibration of the critical $\lambda_k$ weights is still commonplace \cite{meng_multi-fidelity_2021, linka_bayesian_2022, molnar_flow_2022}, robust Bayesian inference strategies should not rely on a-priori hand-tuning or biased calibration of the posterior distribution. Indeed, appropriately setting these parameters is neither easy nor computationally efficient, especially for multi-objective inverse problems arising from real-world data. Developing an alternative that accounts for this multitask consideration becomes crucial to ensure robust sampling when dealing with coupled physics-based and data-driven inference. 

We benefit from an efficient BPINN framework developed in our previous work \cite{PEREZ2023112342}, which robustly addressed multitask Bayesian inference problems with potential multiscale effects, stiffness issues, or competing tasks. This new strategy relied on an adaptive and automatic weighting of the target posterior distribution based on an Inverse Dirichlet control of the weights $\lambda_k$ \cite{maddu_inverse_2022}, which leverages gradient variances information of the different tasks: 
\begin{equation}
\label{Lambda_weights_Inv_Dir}
    \lambda_k =\left( \frac{\gamma^2}{\mathrm{Var}\{\nabla_\Theta \mathcal{L}_k \} }\right)^{1/2}, \quad \text{with} \quad \gamma^2 := \min_{t=0..K} (\mathrm{Var}\{\nabla_\Theta \mathcal{L}_t \}),\quad \forall k = 0,...,K.
\end{equation}
This results in an alternative sampler called Adaptively Weighted Hamiltonian Monte Carlo (AW-HMC) (see Appendix \ref{sec:Algo_Stop_Criterion} for the AW-HMC algorithm and \cite{PEREZ2023112342} for the detailed methodological development) that ensures the gradients distributions of the weighted potential energy terms $L_k= \lambda_k\mathcal{L}_k$ in \eqref{multi_pot} have balanced variances. In this sense, the AW-HMC sampler avoids imbalanced conditions between the different tasks. Moreover, this enables us to concentrate the sampling on the Pareto front exploration after the adaptive procedure, as balancing the target distribution based on the minimum variance of the gradients $\gamma^2$ in \eqref{Lambda_weights_Inv_Dir} contributes to adjusting the weights $\lambda_k$ with respect to the most likely task in the multi-objective inverse problem. In particular, the most uncertain tasks will experience an automatic increase of their uncertainties. The weights $\lambda_k$ are adjusted during the sampling for a number of adaptive steps $\tau \leq N$, characterizing the convergence of the weighted posterior distribution toward the high probability-density region and, therefore, allowing the efficient exploration of the Pareto front neighborhood after their adaptation. The stopping criterion of the adaptive weighting is based on the evolution of the weighted Hamiltonian energy $H_{\lambda_\tau}(\Theta, r)$, defined by 
\begin{equation}
    \label{Weighted_Ham}
    H_{\lambda_\tau}(\Theta, r) = \sum_{k=0}^{K+1} \lambda_k(\tau) \mathcal{L}_k(\Theta) + K(r)
\end{equation}
for the adaptive sampling steps $\tau$. In particular, we rely on both a maximum number of adaptive steps $N_{\max}$ and a threshold $S_{\min}$ on the local variation of the Hamiltonian to set up the effective number of adaptive steps $N$ (see Algorithm \ref{Adaptively Weighted Hamiltonian Monte Carlo} in Appendix \ref{sec:Algo_Stop_Criterion}).

The AW-HMC sampler benefits from enhanced convergence and stability compared to conventional samplers, such as HMC or NUTS \cite{homan_no-u-turn_2014}, and reduces sampling bias by avoiding manual tuning of critical weighting parameters. In terms of computational efficiency, it retains comparable training costs to the previous alternatives while circumventing the issue of excessively reducing the leapfrog time step $\delta t$ to ensure the Hamiltonian conservation, as can occur with NUTS. Indeed, when integrating multiple sources of uncertainties in a multi-objective inverse problem, the NUTS sampler will adjust $\delta t$ to satisfy the stiffest constraint, which can require very small time steps and is thereby more likely to generate random walk behaviors. On the contrary, the AW-HMC sampler guarantees the Hamiltonian conservation with optimal leapfrog time step in terms of convergence rates for a similar training cost, therefore improving the sampler convergence and robustness. This new alternative also demonstrated efficiency in managing the scaling sensitivity of the different terms either to noise distributions (homo- or hetero-scedastic) or multi-scale issues. In fact, the adjusted weights bring information on the distinct task uncertainties. This improves the reliability of the noise-related and model adequacy estimates as the uncertainties are quantified with minimal a-priori assumptions on their scaling. Our novel sampling strategy has demonstrated outstanding performances on several levels of complexity. This covers applications ranging from data-fitting predictions based on sparse measurements, physics-based data-assimilation problems, data-assimilation in inverse problems with unknown PDE model parameters, and data-assimilation in inverse problems with unknown parameters and latent fields \cite{PEREZ2023112342}. Taken together, the AW-HMC sampler enhances BPINN robustness and offers a promising alternative as an overall data-assimilation strategy, extending its applications to more complex Bayesian inference problems. Indeed, this adaptive weighting sampling presents the ability to effectively address multiscale and multitask inverse problems, to couple UQ with physical priors, and to handle sparse noisy data. It also showed effectiveness in addressing stiff dynamics problems including latent field reconstruction and deriving unbiased uncertainty information from the measurement data. The AW-HMC strategy appears as an unlocking advance bringing robustness to the BPINNs and, therefore, provides a promising data assimilation framework to address robust and reliable Bayesian inference in multitask inverse problems.


\section{Data assimilation strategy: sequential reinforcement and operator differentiation}
\label{sec:Method}

In the present work, we focus on a multitask inverse problem for reactive flows at the pore scale involving the identification of two inverse parameters ($\Da2^*$ and $D_m^*$) and a latent concentration field $C$. This novel approach combines dynamical imaging data of calcite dissolution and physics-based regularization induced by the dimensionless PDE model \eqref{eq:inverse_chem_adim}. We build the current data assimilation approach upon the efficient AW-HMC framework for Bayesian Physics-Informed Neural Networks, presented in Sect. \ref{subsec:AW-HMC}, to quantify both morphological and chemical parameter uncertainties. The challenges arising from reactive inverse problems in pore-scale imaging currently lie in the limited variety of available data, as the calibration of the coupled PDE system involves merely dynamical information obtained from the $\mu$CT scans. In particular, the lack of information on the concentration field evolution throughout dissolution necessitates identifying the latent field $C$, whose reconstruction is crucial for inferring the inverse parameters. Moreover, this work is positioned within a diffusive-dominated regime (see Sect. \ref{subsec:model_assumptions}) and therefore neglects the advective effects, which would require estimating additional latent variables for the velocity field in the coupled PDE system \eqref{eq:dimless_model}. The shape complexity of 3D porous samples, along with the non-linear operators involved in the PDE model, are additional concerns that make such an application challenging, especially when targeting full 4D data assimilation on real-life materials. In this section, we present the data assimilation method developed to handle such pore-scale imaging inverse problems. Our methodology emphasizes the sequential reinforcement of the multi-potential energy and the efficient computation of the heterogeneous diffusion operator arising from Archie's law. This first requires setting up a few dedicated notations.

\subsection{Domain decomposition and sampling notation setup}

Dynamical synthetic or experimental $\mu$CT images are available on the overall spatiotemporal domain $\Omega\times(0, T_f)$, and provide dissolution observations subject to noise and imaging limitations (see Sect. \ref{subsec:Xray_limits} and \ref{subsec:Xray_UQ}): we introduce the two-index set $\mathfrak{Im}_{i,j}$ of image intensities (dissolution measurements) defined for the whole image voxels. Then, we define a subset of this image for sampling purposes, involving the positions $\mathcal{D}$ as a subset of $\bar\Omega\times(0, T_f)$ together with their image intensities $\Im$, corresponding to $N_{\text{obs}}$ partial and corrupted training observations:
\begin{equation}
\label{D}
     \mathcal{D} = \left\{  (x_k,t_k), \quad k=1...N_{\text{obs}} \right\}
\end{equation}
On this set of discrete points, there exists mappings $i$ and $j$ such as the image intensity satisfies
\[
\Im_k := \mathfrak{Im}_{i(k),j(k)} = 1-\varepsilon(x_k,t_k) + \xi(x_k,t_k), \, k=1...N_{\text{obs}}
\]
where the noise $\xi \sim \mathcal{N}(0,\sigma^2I)$ for which the standard deviation $\sigma$ is automatically estimated in the AW-HMC sampler by means of the $\lambda_k$ adjustment in equation \eqref{multi_pot}. This relationship between the microtomography images $\Im$ and $\varepsilon$ comes from the correlation between the $\mu$CT values and the material local attenuation. Indeed, in a greyscale tomographic scan, the minimum signal corresponds to the least attenuating or the least dense areas (in $\Omega_F$ where $\varepsilon = 1$), while the maximum signal refers to the most attenuating areas (in $\Omega_S$ where $0<\varepsilon_0\leq\varepsilon<1$). Due to the micro-continuum description of the medium based on the two-scale porosity assumption, each distinct region of the domain --- namely $\Omega_S$ and $\Omega_F$ --- is though regarded as a different term in the multi-potential energy definition. Such a distinction is prescribed since there is no guarantee that the data corruption is uniform: the measurement variability can differ locally when facing heteroscedastic noise. In particular, the artifact limitations tend to enhance the blurring effects at the material fluid/solid interface $\Sigma$. This motivates the special consideration of this interface neighborhood to account for the unresolved features and provide reliable morphological uncertainties. 

In this sense, we introduce the Reactive Area of Interest (RAI) as the evolving fluid/mineral interface along the dissolution process that is defined by: 
\begin{equation}
    \label{RAI}
    \text{RAI} = \left\{  (x_k,t_k) \in \Omega\times(0, T_f) \quad \text{such that}\quad \mathfrak{Im}_{i(k)+1,j(k)} < \mathfrak{Im}_{i(k),j(k)}  \quad \text{and} \quad  0.1<\Im_k<0.9  \right\}
    \end{equation}
where the imaging extreme values are ignored, as a correction criterion, to avoid integrating noise derivation artifacts into the definition of the RAI. These thresholds rely on the analysis of the $\mu$CT histogram of the initial porous medium dataset. We also define:
\begin{itemize}
    \item The extended Reactive Area of Interest, denoted $\text{RAI}^+$, 
    as the RAI augmented by a fluid tubular neighborhood of the RAI both in space and time, that is to say, the fluid region close to the evolving interface in $\Omega\times(0,T_f)$,
    \item The reduced Reactive Area of Interest, denoted $\text{RAI}^-$, based on the acceptability criterion --- positivity of the $D_m^*$ estimations --- defined thereafter and related to the relations \eqref{eq:dC_ZIR}-\eqref{eq:RAI-}.    
\end{itemize}

Moreover, we introduce several discrete domains 
$\mathcal{D}^\bullet$ defined as the intersection between the non-reactive part of $\mathcal{D}$ and their respective time-dependent regions: fluid, solid, reactive, or boundary. 
For instance, one gets $\mathcal{D}^F = (\mathcal{D}\cap Q_F)\smallsetminus\RAI$ with $Q_F$ the evolving fluid region defined as 
\begin{equation}
    Q_F = \left\{(x,t)\in\Omega\times(0,T_f)  \quad \text{such that}\quad x\in\Omega_F(t) \right\}.
\end{equation}
In the same way, $\mathcal{D}^S = (\mathcal{D}\cap Q_S)\smallsetminus\RAI$ where $Q_S$ is the evolving solid region and $\mathcal{D}^\partial = \mathcal{D}\cap\{\partial\Omega\times(0,T_f)\}$. The overall domain $\Omega\times(0,T_f)$ is decomposed into several regions and respective training datasets that are involved in the sequential reinforcement of the multi-potential energy. 

Finally, these different domains satisfy the following properties:
\begin{itemize}
    \item $\text{RAI}^- \subset \text{RAI} \subset \text{RAI}^+$,
    \item $\Omega_F\cup \Omega_S = \Omega$, where $\Omega_F=\mathbb{1}_{(\eps<1)}$ in $\Omega$ is an open set,
    \item $\mathcal{D} = \mathcal{D}^F \cup \mathcal{D}^S\cup \mathcal{D}^\partial\cup \RAI$ .
\end{itemize}


\subsection{Sequential reinforcement of the multi-potential energy}
\label{subsec:seq_reinforce}
The data assimilation strategy developed in the present article relies on a sequential design of the multi-potential energy $U(\Theta)$, which will be reinforced to incorporate additional constraints and successive modeling approximations through dedicated sampling steps. While this sequential reinforcement approach draws significant inspiration from the PINNs framework, where the regression of physical fields typically precedes the incorporation of physics-based constraints, it also introduces a modeling reinforcement, crucial for inferring latent fields in strongly coupled PDE systems. In particular, this sequential splitting is necessary due to the coupling between the porosity field $\varepsilon$, related to the $\mu$CT imaging process, the latent concentration field $C$, and the two unknown inverse parameters $\Da2^*$ and $D_m^*$. The first sampling step is, therefore, dedicated to providing a-priori estimations of the micro-porosity field through data-fitting terms only, while the following steps address the modeling reinforcement.

\begin{figure}
    \centering
    \includegraphics[width=0.85\linewidth]{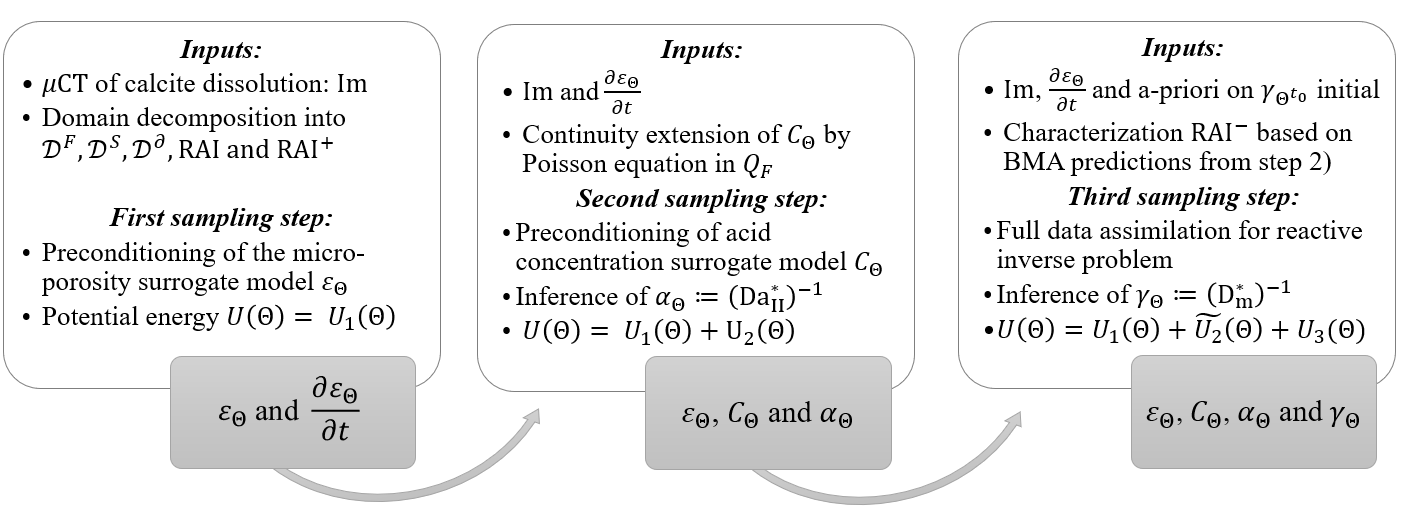}
    \caption{\textbf{Sequential graph of the potential energy reinforcement:} our data assimilation strategy incorporates additional physics-based constraints arising from the PDE model \eqref{eq:inverse_chem_adim_Vf} through successive sampling steps. The notations are defined in Sect. \ref{subsec:seq_reinforce}.}
    \label{fig:Algorithm}
\end{figure}

This sequence of tasks is then decomposed into three steps, from the one for which we have the most information to the set of tasks involving all the aspects and constraints we need to consider, as displayed in Fig~\ref{fig:Algorithm} and developed thereafter:
\begin{itemize}
    \item Step 1: Preconditioning on the micro-porosity by pure regression on image data,
    \item Step 2: Preconditioning of the latent reactive fluid with additional PDE constraint, given a first modeling approximation of the coupled PDE system
    \item Step 3: Overall data assimilation potential with the full and exact reactive model.
\end{itemize}

Extensions of the present methodology can be investigated for applications in other disciplines. However, the elaboration of the sequential reinforcement approach would rather be problem-dependent, and in this sense, a automatic formalisation of this process currently seems out of reach. It would essentially depend on the number of inverse parameters to identify, the presence of latent fields, the amount of data available, and would overall be affected by the under-determined or ill-conditioned nature of the problem. For instance, when facing numerous unknowns in a system with limited data, splitting the constraints and utilizing intermediate approximate modeling becomes increasingly necessary to infer the full dynamics. In this sense, the idea of sequential reinforcement, especially when dealing with complex and strongly coupled dynamic behaviors, can be applied to other fields.

For instance, calibrating a large number of inverse parameters in a system of chemical kinetic equations for an uncertainty-quantified pyrolysis model, with applications to the atmospheric entry of spacecraft, also requires sampling a Bayesian posterior in high dimensions with limited knowledge of various sources of uncertainty \cite{coheur_bayesian_2023}. In biology, while may have access to a slightly larger variety of data, including measurements of concentrations and velocity fields in the context of kinetic reactions between proteins, the sparsity of the information and the absence of predictive models can, however, raise additional challenges \cite{maddu_stability_2019}.

Thereafter, we develop each of the sequential reinforcement steps within the framework of reactive inverse problems in pore-scale imaging.

\subsubsection{Step 1: Preconditioning by pure regression on image data}

The first sampling step 1 of the sequential splitting strategy aims to provide a preconditioning description of the surrogate micro-porosity field $\varepsilon_\Theta$. We consider a task differentiation between $\mathcal{D}^S$ and $\text{RAI}^+$ and hence, we discard from the training the fluid measurements which are far from the mineral interface --- as they are of no interest in this first step to characterize morphological uncertainty on $\varepsilon$. The resulting potential energy term writes: 
\begin{equation}
    \label{U1}
    \ds U_1(\Theta) = \frac{\lambda_0}{2\sigma_0^2}\left\|1-\varepsilon_\Theta - \Im\right\|_{\mathcal{D}^S}^2
    +  \frac{\lambda_1}{2\sigma_1^2}\left\|1-\varepsilon_\Theta - \Im\right\|_{\text{RAI}^+}^2 + \frac{1}{2\sigma_\Theta^2}\|\Theta\|^2 
\end{equation}
where $\sigma_k$ are unknown standard deviations characterizing the noise distributions on their respective areas, and we assume a prior distribution on $\Theta$ given by $P(\Theta) \sim \mathcal{N}(0, \sigma_\Theta^2 I_{p+d})$. The notation $\|\cdot\|$ refers to either the RMS (root mean square) norm --- inherited from the functional $\L^2$-norm --- for the two first log-likelihood terms or to the usual Euclidean norm for the last log-prior term. In practice, we do not rely on a-priori manual calibration of the noise magnitudes --- all $\sigma_k$ are set to be equal --- but rather use the AW-HMC sampler to automatically and adaptively estimate these uncertainties through adjustments of the $\lambda_k$. This is especially meaningful in the neighborhood of the evolving fluid/solid interface (corresponding to the $\text{RAI}^+$) to study edge-enhancement implications and in the pure solid region to quantify the unresolved features. At the end of step 1, one gets a first a-priori estimation of the field $\varepsilon$, which presents the advantage of being denoised compared to the $\mu$CT images and hence is more suitable to differentiate. In this sense, we now have access to the time derivative of the surrogate porosity $\varepsilon_\Theta$ that is subsequently used to provide some preconditioning of the latent concentration field $C$. 

\subsubsection{Step 2: Preconditioning of the latent reactive fluid}

The second sampling step 2 relies on this first insight of the Bayesian neural network parameters obtained through step 1. We hence restart an adaptive weighting procedure with the AW-HMC sampler by adding additional constraints arising from the PDE model \eqref{eq:inverse_chem_adim_Vf}. As the acid concentration is a latent unknown field in our reactive inverse problem, we benefit from this second sampling step to provide a surrogate estimation $C_\Theta$ of this field and identify a first reactive parameter, namely $\Da2^*$. In this sense, we impose a physics-based regularization linking the porosity derivative to the surrogate concentration field through the PDE equation: 
\begin{equation}
    \label{eq:der_eps_zir}
     \ds \frac{1}{\upsilon C_0}\frac{\partial \varepsilon_\Theta}{\partial t} - \Da2^* C_\Theta  \mathbb{1}_{\{(1-\varepsilon)>0\}}= 0
\end{equation}
where the calcite molar volume $\upsilon$ and $C_0$ are constant parameters --- we assume the concentration $C_0$ of continuous acid injection, defining the Dirichlet boundary conditions on $C$, to be known. In a direct formulation, equation \eqref{eq:der_eps_zir} is imposed over the whole domain $\Omega\times(0,T_f)$, though the PDE constraint in inverse modeling mainly brings meaningful information on the RAI. Indeed, we have $\ds  \frac{\partial \varepsilon_\Theta}{\partial t} \simeq -\frac{\partial \Im}{\partial t}>0$ on the reactive area of interest, which is useful to characterize the reaction regime through the $\Da2^*$ dimensionless number. In the pure solid region where $\ds\frac{\partial \varepsilon_\Theta}{\partial t} = 0$, the PDE constraint \eqref{eq:der_eps_zir} translates into low acid penetration in $Q_S$ that we impose through the condition $C_\Theta = c_0 = \num{e-7}$. In the fluid region $Q_F$, the latent acid concentration field is a solution of the following heat equation: 
\begin{equation}
    \label{eq:heat_C}     
        \ds \frac{\partial C_\Theta}{\partial t}-D_m^* \Delta C_\Theta= 0
\end{equation}
with initial and boundary conditions, respectively unknown in the inverse formulation and non-homogeneous Dirichlet boundary conditions (see the dimensionless PDE model \eqref{eq:inverse_chem_adim_Vf} from Sect. \ref{subsec:Adim_inverse}). Following the modeling assumptions of Sect. \ref{subsec:model_assumptions}, especially on the diffusive dominated regime with $\Pe\ll 1$, we though assume as a first approximation that the surrogate acid concentration field $C_\Theta$ is driven by the quasy-stationary Poisson equation $\Delta C_\Theta = 0$ in $Q_F$. This behaves as a continuity extension of the surrogate concentration field from the domain boundary to the mineral evolving interface defined by the $\text{RAI}$ and furthermore serves as a first modeling approximation of the PDE system, which is necessary to infer the latent concentration field. Along with the PDE equation \eqref{eq:der_eps_zir} in the RAI, this defines augmented multi-potential energy for the second sampling step: 
\begin{equation}
    \label{U2}
    \begin{aligned}
    \ds U(\Theta) & = U_1(\Theta) +\frac{\lambda_2}{2\sigma_2^2}\left\| (\upsilon C_0)^{-1} \alpha_\Theta \frac{\partial \varepsilon_\Theta}{\partial t} - C_\Theta \right\|_{\text{RAI}}^2
    +  \frac{\lambda_3}{2\sigma_3^2}\left\| \Delta C_\Theta \right\|_{\mathcal{D}^F}^2 \\[2mm]
    & + \frac{\lambda_4}{2\sigma_4^2} \left( \left\| 1 -C_\Theta \right\|_{\mathcal{D}^\partial}^2+  \left\| c_0 -C_\Theta \right\|_{\mathcal{D}^S}^2 \right) + \frac{1}{2\sigma_\Theta^2}\|\Theta\|^2 \\[2mm]
    & :=   U_1(\Theta) + U_2(\Theta)
    \end{aligned}
\end{equation}
where the constant constraints on the boundary $\mathcal{D}^\partial$ and solid $\mathcal{D}^S$ datasets are gathered as a single term. The notation $\alpha_\Theta$ here refers to the first inverse parameter effectively sampled, with $\alpha_\Theta := (\Da2^*)^{-1}$. This second step reinforces the sampling of the surrogate micro-porosity $\varepsilon_\Theta$ by providing insight into the latent concentration field $C_\Theta$ on the RAI and posterior distribution on the inverse parameter $\Da2^*$. 

\subsubsection{Step 3: Overall data assimilation potential with full reactive model}

Finally, the third sampling step 3 will address the overall reactive inverse problem, to refine the micro-porosity and acid concentration predictions accounted for the exact and fully coupled PDE model \eqref{eq:inverse_chem_adim_Vf} and provide uncertainty quantification on the inverse parameters $\Da2^*$ and $D_m^*$. The extension by continuity of the acid concentration --- from the Poisson equation in $Q_F$ --- is replaced by its corresponding heat equation term \eqref{eq:heat_C} (see equation \ref{U3} bellow). We also use the diffusion-reaction PDE coupling $\varepsilon_\Theta$ and $C_\Theta$ to infer the dimensionless number $D_m^*$:
\begin{equation}
    \label{eq:dC_ZIR}
    \frac{\partial C_\Theta}{\partial t}-D_m^*\nabla\cdot\left(\varepsilon_\Theta^{1+\beta}\nabla(\varepsilon_\Theta^{-1}C_\Theta)\right)+\Da2^* C_\Theta = \frac{\partial C_\Theta}{\partial t}-D_m^*\nabla\cdot\left(\varepsilon_\Theta^{1+\beta}\nabla(\varepsilon_\Theta^{-1}C_\Theta)\right)+ \frac{1}{\upsilon C_0}\frac{\partial\varepsilon_\Theta}{\partial t} = 0
\end{equation}
which is theoretically valid on the whole RAI for the inverse modeling. Nonetheless, the heterogeneous diffusion term $\mathcal{D}_i(\varepsilon, C):= \nabla\cdot\left(\varepsilon^{1+\beta}\nabla(\varepsilon^{-1}C)\right)$ arising from Archie's law becomes highly sensitive at the mineral boundary due to jumps in the porosity derivatives at the interface. This may disrupt the identification of the inverse parameter $D_m^*$. The PDE constraint \eqref{eq:dC_ZIR} therefore needs to be imposed on a reduced neighborhood of the reactive area of interest, namely the $\text{RAI}^-$ domain. 

This restricted RAI is then defined by the eligible points of the RAI domain where $D_m^*$ is predicted positive. From the overall samples of step 2, we compute the predictive BMA distributions of the two operators 
\begin{equation} 
\label{eq:RAI-}
    \ds\frac{\partial C_\Theta}{\partial t} + \frac{1}{\upsilon C_0}\frac{\partial\varepsilon_\Theta}{\partial t} \quad \text{and}\quad \mathcal{D}_i(\varepsilon_\Theta, C_\Theta) 
\end{equation}
on the domain $\Omega\times(0, T_f)$ and then estimate $D_m^*$ through equation \eqref{eq:dC_ZIR} to define the $\text{RAI}^-$ domain.

From this procedure, one also gets an estimate of the posterior distribution of $D_m^*$ after sampling step 2 which is regarded as an initial a-priori on this inverse parameter in step 3. This will be further detailed in the applications (see Sect. \ref{sec:Results1D} and \ref{sec:Results}). Taken together, the fully reinforced multi-potential energy for the third sampling step writes: 
\begin{equation}
    \label{U3}
    \begin{aligned}
    \ds U(\Theta) & =  U_1(\Theta) +\frac{\lambda_2}{2\sigma_2^2}\left\| (\upsilon C_0)^{-1} \alpha_\Theta \frac{\partial \varepsilon_\Theta}{\partial t} - C_\Theta \right\|_{\text{RAI}}^2
    +  \frac{\lambda_3}{2\sigma_3^2}\left\| \gamma_\Theta \frac{\partial C_\Theta}{\partial t} - \Delta C_\Theta \right\|_{\mathcal{D}^F}^2 \\[2mm]
    & + \frac{\lambda_4}{2\sigma_4^2} \left( \left\| 1 -C_\Theta \right\|_{\mathcal{D}^\partial}^2+  \left\| c_0 -C_\Theta \right\|_{\mathcal{D}^S}^2 \right) \\[2mm]
    & + \frac{\lambda_5}{2\sigma_5^2}\left\| 
    \gamma_\Theta\left( \frac{\partial C_\Theta}{\partial t} + 
    (\upsilon C_0)^{-1}\frac{\partial \varepsilon_\Theta}{\partial t} \right) -\nabla\cdot\left(\varepsilon_\Theta^{1+\beta}\nabla(\varepsilon_\Theta^{-1}C_\Theta)\right) 
    \right\|_{\text{RAI}^-}^2 + \frac{1}{2\sigma_\Theta^2}\|\Theta\|^2 \\[2mm]
    &:= U_1(\Theta) + \Tilde{U_2}(\Theta) + U_3(\Theta)
    \end{aligned}
\end{equation}
where $ \gamma_\Theta:= (D_m^*)^{-1}$ such that the set of inverse parameters we infer in practice is $(\mathcal{P}_\mathrm{inv})_\Theta=\left\{\alpha_\Theta, \gamma_\Theta \right\}$. The data assimilation strategy developed in the present article incorporates successive physics-based constraints using a sequential reinforcement of the multi-potential energy $U(\Theta)$. This is achieved by splitting the sampling steps, which is required due to the strong coupling of the overall PDE system \eqref{eq:inverse_chem_adim_Vf} involving latent field and unknown parameters. This overall algorithm is summarized in Fig \ref{fig:Algorithm}.

\subsection{Computational strategy for differential operator expression}

This section is dedicated to the development of a differentiation strategy for efficient computation of the heterogeneous diffusion $\mathcal{D}_i(\varepsilon, C)$ arising from Archie's law. Indeed, the third sampling step in the sequential reinforcement of the multi-potential energy (see Sect. \ref{subsec:seq_reinforce}) involves the computation of this diffusion operator through a neural network surrogate model. This implies the use of automatic differentiation (AD) which is a prevalent technique in deep-learning frameworks such as Physics-Informed Neural Networks (PINNs) and Bayesian Physics-Informed Neural Networks (BPINNs). Such an automatic differentiation relies on gradient backpropagation to compute the derivatives of the neural network functional outputs with respect to its inputs, by using the chain rule principle. AD is thus a fast computational technique when it comes to the evaluation of first and second-order derivatives of the output fields, namely the spatial gradient and Laplacian operators, and the temporal partial derivatives. More complex non-linear operators resulting from two successive differentiation of non-trivial functional compositions --- as this is the case for the $\mathcal{D}_i(\varepsilon, C)$ operator --- can though readily lead to high-computational cost. This observation leads to reconsidering the heterogeneous diffusion term as a succession of sum and product of first and second-order operators. Consequently, we consider the diffusion operator $\mathcal{D}_i(\varepsilon, C)$ under two formulations: its compact form \eqref{eq:Diff_op_original} and its developed form \eqref{eq:Diff_op_dev} reading

\begin{subequations}
\label{eq:Diff_op}
     \begin{align}
      \mathcal{D}_i(\varepsilon, C) &= \nabla\cdot\left(\varepsilon^{\beta+1}\nabla(\varepsilon^{-1}C)\right) 
      \label{eq:Diff_op_original}\\
      &= \nabla\cdot\left(\varepsilon^\beta\nabla C - \varepsilon^{\beta-1}C\nabla\varepsilon\right)
      = \nabla\cdot\left(\varepsilon^{\beta-1}( \varepsilon\nabla C-C\nabla\varepsilon )\right) \\
       &= \varepsilon^{\beta-1}\left(\varepsilon\Delta C-C\Delta \varepsilon \right) + (\beta-1)\varepsilon^{\beta-1}\, \nabla\varepsilon \cdot\nabla C+(\beta-1)\varepsilon^{\beta-2}\,C\,\nabla\varepsilon\cdot\nabla\varepsilon \label{eq:Diff_op_dev}
     \end{align}
\end{subequations}
Then we replace the expression of the diffusion in the multi-potential energy \eqref{U3} with the novel operator formulation \eqref{eq:Diff_op_dev}. This makes possible to reduce the computational cost of evaluating this diffusion operator through merely the auto differentiation of the following terms: $\nabla \varepsilon_\Theta$, $\nabla C_\Theta$, $\Delta \varepsilon_\Theta$, and $\Delta C_\Theta$. Finally, we observe on the developed expression \eqref{eq:Diff_op_dev} that the case $\beta = 1$ even results in a more straightforward expression of Archie's law which then writes $\mathcal{D}_i(\varepsilon, C) = \varepsilon\Delta C-C\Delta \varepsilon$. This is particularly convenient as the tortuosity index $\beta = 1$ can be regarded as an approximation of the effective diffusivity in pore-scale models (\emph{e.g.} see \cite{soulaine_mineral_2017}). Furthermore, this reduced expression confirms the high sensitivity of the heterogeneous diffusion term at the mineral boundary $\sigma$ due to the micro-porosity Laplacian involved in Archie's law. Considering suitable differential operator expressions can enhance the surrogate model efficiency by reducing the automatic differentiation cost. 

\begin{table}[t]
    \centering
    \renewcommand{\arraystretch}{1.25}
    \begin{tabular}{|c|c|c|c|c|}
    \multicolumn{5}{l}{a) Comparison of diffusion operators on the 1D+Time}\\[4pt]
    \hline
        & $T_{\mathrm{CPU}}$ (ms) & Speedup $S$ & $T_{\mathrm{GPU}}$ (ms) & Speedup $S$ \\ \hline
        Original operator $\mathcal{D}_i(\varepsilon, C)$ & 37.13 & 1 & 11.32 & 3.28   \\ \hline
        Developed operator \eqref{eq:Diff_op_dev} with $\beta \neq 1$ & 24.78 & 1.49 & 6.985 & 5.32 \\ \hline
        Developed operator \eqref{eq:Diff_op_dev} with $\beta = 1$ & 10.66 & 3.48 & 6.739 & 5.51 \\ \hline
        \multicolumn{5}{l}{}\\
        \multicolumn{5}{l}{b) Comparison of diffusion operators on the 2D+Time}\\[4pt]
    \hline
        & $T_{\mathrm{CPU}}$ (ms) & Speedup $S$ & $T_{\mathrm{GPU}}$ (ms) & Speedup $S$\\ \hline
         Original operator $\mathcal{D}_i(\varepsilon, C)$ & 83.88 & 1 & 13.69 & 6.12   \\ \hline
        Developed operator \eqref{eq:Diff_op_dev} with $\beta \neq 1$ & 72.63 & 1.16 & 10.63 & 7.89 \\ \hline
        Developed operator \eqref{eq:Diff_op_dev} with $\beta = 1$ & 68.99 & 1.22 & 10.59 & 7.92  \\ \hline
    \end{tabular}
\caption{\textbf{Computational times of the diffusion operators on the 1D+Time and 2D+Time inverse problem:} comparison between several expressions of the differential operator for CPU and GPU implementations. The Speedup is computed as $S = T_0/T_\bullet$ with $T_0:= T_{\mathrm{CPU}}$ for the original expression $\mathcal{D}_i(\varepsilon, C) = \nabla\cdot\left(\varepsilon^{\beta+1}\nabla(\varepsilon^{-1}C)\right)$. All the computational times are expressed in milliseconds (ms) and are averaged on several evaluations of the diffusion operator during the sampling procedure.}
    \label{tab:Operators_time_1D}
\end{table}

We perform some validations of this first insight through computational time measurements for the different expressions of the heterogeneous diffusion term. We hence compare the original formulation of $\mathcal{D}_i(\varepsilon, C)$ with the developed operator \eqref{eq:Diff_op_dev} for tortuosity coefficients $\beta = 0.5$ --- to consider the most general form --- and $\beta = 1$ that leads to the reduced evaluation of the heterogeneous diffusion. We account for computational times on both CPU and GPU devices and perform several evaluations of the diffusion operator expressions to provide averaged computational times along the sampling procedure of step 3. These results are summarized in Table \ref{tab:Operators_time_1D} for the 1D+Time test case (Table \ref{tab:Operators_time_1D}.a) and 2D+Time application (Table \ref{tab:Operators_time_1D}.b) explicitly developed in the validation Sect. \ref{sec:Results1D} and application Sect. \ref{sec:Results}. We define in each case the reference computational cost $T_0$ as the CPU time necessary to evaluate the original operator expression $\mathcal{D}_i(\varepsilon, C)$. This respectively leads to $T_0 = 37.13$ ms and $T_0 = 83.88$ ms for 1D+Time and 2D+Time applications. We then evaluate the speedup, denoted $S$, of the distinct operator formulations as $S = T_0/T_\bullet$ where $T_\bullet$ are their respective computational times. It results from this comparison an effective improvement of the computational costs, either on CPU or GPU, when considering the developed operator \eqref{eq:Diff_op_dev} from equation \eqref{eq:Diff_op} --- even in its general form for $\beta \neq 1$ (see second rows of Table \ref{tab:Operators_time_1D}.a and 
 \ref{tab:Operators_time_1D}.b). This highlights that the configuration that best optimizes the speedup is to use the operator \eqref{eq:Diff_op_dev}  on GPU devices. The improvement between the general and reduced form of the operator \eqref{eq:Diff_op_dev} is, however, less significant especially in 2D+Time. This can be readily explained by the fact that most of the computational time is spent evaluating the gradient and Laplacian operators rather than their combination. In this sense, the general form \eqref{eq:Diff_op_dev} can be used effectively regardless of the tortuosity index value $\beta$. Overall, the developed heterogeneous diffusion operator \eqref{eq:Diff_op_dev} contributes to reducing the computational cost of its single evaluation. 

Suitable choices in the differential operator expressions considerably improve the AD cost when considering complex non-linear operators with non-trivial functional compositions, as for Archie's law. This, therefore, reduces the computational time spent in evaluating one instance of these operators. Such improvements are meaningful, accounting for the numerous surrogate model evaluations required for an appropriate sampling of the target posterior distribution \eqref{post_dist_Bayes} through MCMC samplers. Considering the developed formulation \eqref{eq:Diff_op_dev} of the heterogeneous diffusion operator on GPU devices hence implies a significant enhancement of the overall computational time of the present data assimilation strategy.

\section{Validation on synthetic 1D+Time calcite dissolution}
\label{sec:Results1D}

In this article, we develop a novel data assimilation strategy to address reactive inverse problems in pore-scale imaging with uncertainty quantification. This aims to quantify morphological uncertainty on the micro-porosity field $\varepsilon$ and estimate reliable ranges of chemical parameters through dynamical $\mu$CT noisy observations augmented with PDE models of dissolution. The present strategy is based on the robust Bayesian framework presented in \cite{PEREZ2023112342} along with the AW-HMC sampler (see Sect. \ref{subsec:AW-HMC}) and relies on sequential reinforcement of the multi-potential energy. 

In this section, we validate the present methodology on inverse problems of calcite dissolution with heterogeneous porosity in artificial 1D spatial configurations. All the $\mu$CT measurements that we consider are synthetic observations resulting from direct numerical simulations of reactive flows with noise perturbation, in order to validate our methodology in well-established test cases. The validation test case is a purely synthetic 1D+Time problem, for which we check two configurations with distinct tortuosity indices with $\beta = 1$ and $\beta = 0.5$.

\subsection{Direct reactive model: problem set up}
\label{subsec:1D_DNS}
\begin{figure}
    \centering
    \includegraphics[width=0.45\linewidth]{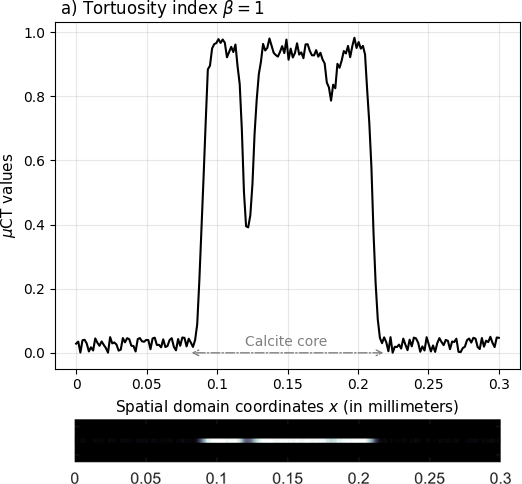}
    \includegraphics[width=0.45\linewidth]{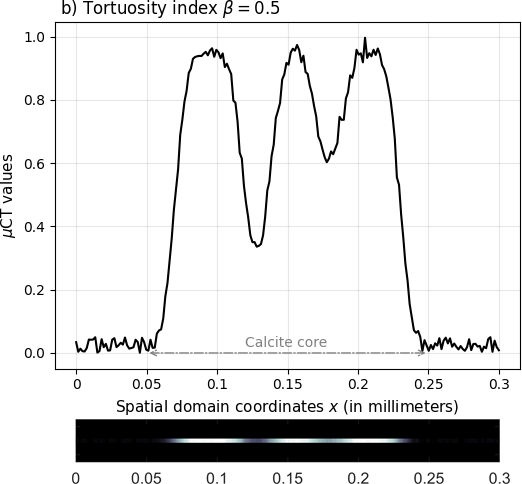}
    \caption{\textbf{Initial $\mu$CT images defining the porous sample geometries:} synthetic cases with tortuosity indices a) $\beta = 1$ and b) $\beta = 0.5$. The $\mu$CT measurements are normalized, corrupted with noise, and provide the dataset $\Im$ before the dissolution process. The calcite core regions are identified by the double-headed arrows, and correspond to the maximum intensity in the greyscale tomographic scans displayed below. }
    \label{fig:Initial_geo_1D}
\end{figure}

We consider two heterogeneous samples of 1D synthetic calcite cores whose initial geometries are characterized by the numerical $\mu$CT images presented in Fig \ref{fig:Initial_geo_1D} on a 'physical' spatial domain $\Omega$ of width $0.3$ mm. These initial images correspond to normalized greyscale tomographic scans, corrupted with noise that either accounts for sensor noise or unresolved morphological features. Direct numerical simulations (DNS) of reactive processes are then performed on these initial geometries to provide synthetic $\mu$CT dynamical images of dissolution. These observation data are generated by solving the reaction-diffusion system \eqref{eq:forward_chem_pb_simple_adim} by means of mesh-based and particle methods --- namely 
a Backward Euler or Mid Point method for the time integration coupled with Particle Strength Exchanges scheme for the heterogeneous diffusion --- on a Cartesian spatiotemporal grid of resolution $N_x = 200$ and $N_t = 240$. Continuous acid injection is maintained through non-homogeneous Dirichlet boundary conditions on the domain $\Omega$ to ensure a diffusive-dominated regime. Numerically, we consider a strong acid solution with $\mathrm{pH} = 0$ such that the normalizing constant $C_0$ equals $1$. The characteristic length $L$ of these porous samples is set to $L=0.1$ mm and the reactive parameters are respectively defined by $K_s = 0.8913\, \mathrm{mol.m^{-2}.s^{-1}}$, $D_m = \num{e-9}\, \mathrm{m^{2}.s^{-1}}$,  and $\gamma_{\ce{H+}} = \num{e-3}\, \mathrm{m^{3}.mol^{-1}}$ --- taken from the benchmark \cite{molins_simulation_2021}. The reactive specific area $A_s$ is set to $A_s = \num{e3}\,\mathrm{m}^{-1}$, and we do not account for the calcite molar volume $\upsilon$ in these test cases --- as such 1D+Time examples do not mean to be physically consistent but rather serve validation purposes. We also consider distinct tortuosity indexes, namely $\beta = 1$ and $\beta = 0.5$, on the different geometries to address both the compact form \eqref{eq:Diff_op_original} and develop form \eqref{eq:Diff_op_dev} of the diffusion operator $\mathcal{D}_i(\varepsilon, C)$ in the data assimilation problem. The DNS is performed until the overall calcite core is dissolved which corresponds to a characteristic final time $T_f = 24$ s. Taken together, one gets a sequence of synthetic $\mu$CT images $\mathfrak{Im}_{i,j}$, similar to Fig \ref{fig:Initial_geo_1D}, characterizing the dissolution process of the two calcite cores on the 'physical' spatiotemporal domain $\Omega\times(0, T_f)$.

\subsection{Dimensionless inverse problem and dimensionless numbers}

From the setting of these reactive parameters, one identifies the dissolution regime of these test cases given by the dimensionless catalytic Damköhler $\Da2 = 8.913$ from equation \eqref{eq:Da2_Da1}. The inference of this Damköhler number is though not straightforward in inverse problems, as developed in Sect. \ref{subsec:Adim_inverse}, and we define the dimensionless time $t^*$ such that the its related final time is $T_f^* = 1$. The dimensionless spatial variable $x^*$ is computed as in the dimensionless formulation of the direct problem using $x^* = x/L$. For the data assimilation, we hence consider the dimensionless domain $\Omega^*\times(0,T_f^*) = [0,3]\times(0,1)$ to extract the observation dataset
\begin{equation}
    \mathcal{D} = \left\{  (x_k,t_k) \in [0,3]\times(0,1) , \quad k=1...N_{\text{obs}} \right\}
\end{equation}
where the number of training points $N_\text{obs} = 7725$ represents about 16\% of the data required for the full field reconstructions on $\Omega^*\times(0,T_f^*)$. The dataset $\mathcal{D}$ is divided into the corresponding datasets $\mathcal{D}^S$, $\mathcal{D}^F$, $\mathcal{D}^\text{RAI}$, $\text{RAI}^+$, $\text{RAI}^-$ and $\mathcal{D}^\partial$ that respectively cover around 50\%, 4\%, 15\%, 13\%, 10\% and 8\% of the $N_\text{obs}$ training measurements. One also determines the scaling dimensionless factor $D_\mathrm{ref}$ which appears in the dimensionless inverse formulation \eqref{eq:inverse_chem_adim_Vf}:
\begin{equation}
    D_\mathrm{ref} = \frac{T_f^* L^2}{T_f} = \num{4.16e-10}\, \mathrm{m^2.s^{-1}}
\end{equation}
according to the relation \eqref{t_star} and the estimations of $T_f, \, T_f^*$ and $L$. We characterize the dissolution regime in these reactive inverse problems by means of the two dimensionless numbers defined in \eqref{eq:dim_numb_inv}, and one gets: 
\begin{equation}
    \ds \Da2^* = 21.3912 \quad  \text{ and } \quad D_m^* = 2.4
\end{equation}
which are related to the inverse parameters to infer through $(\mathcal{P}_\mathrm{inv})_\Theta=\left\{\alpha_\Theta, \gamma_\Theta \right\} = \left\{ (\Da2^*)^{-1}, (D_m^*)^{-1} \right\}$. For such 1D+Time reactive inverse problems, the overall multi-potential energy finally writes in the case $\beta = 1$:
\begin{equation}
\label{multi_pot_1DTime}
    \begin{aligned}
    \ds U(\Theta) & =  \frac{\lambda_0}{2\sigma_0^2}\left\|1-\varepsilon_\Theta - \Im\right\|_{\mathcal{D}^S}^2
    +  \frac{\lambda_1}{2\sigma_1^2}\left\|1-\varepsilon_\Theta - \Im\right\|_{\text{RAI}^+}^2
    +\frac{\lambda_2}{2\sigma_2^2}\left\| \alpha_\Theta \frac{\partial \varepsilon_\Theta}{\partial t} - C_\Theta \right\|_{\text{RAI}}^2\\[2mm]  
    & +  \frac{\lambda_3}{2\sigma_3^2}\left\| \gamma_\Theta \frac{\partial C_\Theta}{\partial t} - \frac{\partial^2 C_\Theta}{\partial x^2} \right\|_{\mathcal{D}^F}^2 
    + \frac{\lambda_4}{2\sigma_4^2} \left( \left\| 1 -C_\Theta \right\|_{\mathcal{D}^\partial}^2+  \left\| \num{e-7} -C_\Theta \right\|_{\mathcal{D}^S}^2 \right) \\[2mm]
    & + \frac{\lambda_5}{2\sigma_5^2}\left\| 
    \gamma_\Theta\left( \frac{\partial C_\Theta}{\partial t} + 
    \frac{\partial \varepsilon_\Theta}{\partial t} \right) - \left(\varepsilon_\Theta \frac{\partial^2 C_\Theta}{\partial x^2} - C_\Theta \frac{\partial^2 \varepsilon_\Theta}{\partial x^2} \right) 
    \right\|_{\text{RAI}^-}^2 + \frac{1}{2\sigma_\Theta^2}\|\Theta\|^2 
    \end{aligned}
\end{equation}
and is sequentially reinforced, as presented in Sect. \ref{subsec:seq_reinforce}, through three successive sampling steps using the AW-HMC sampler. The hyperparameters setting of the sequential AW-HMC samplers together with the neural network architecture are detailed hereafter. 
\subsection{Deep learning configuration}
\label{subsec:deep_learning_1D}
Regarding the deep learning strategy for the overall data assimilation problem, we use two distinct neural network architectures to define the micro-porosity and acid concentration surrogate models. Each surrogate model has, therefore, one single output corresponding to $\varepsilon_\Theta$ and $C_\Theta$, respectively. This is preferred to merging the two outputs into a single neural network architecture to avoid a strong correlation between the output fields. Indeed, providing surrogate models not strongly correlated with a multiple-output neural network may require numerous hidden layers that straightforwardly impact the overall computational cost. On the contrary, using distinct neural networks makes it possible to build independent surrogate models while retaining few hidden layers and, therefore, a reasonable number of neural network parameters. Correlations between the two neural network architectures, and then the outputs fields $\varepsilon_\Theta$ and $C_\Theta$, are merely achieved through the PDE model defining the multi-potential energy \eqref{multi_pot_1DTime}. This deep learning configuration is more meaningful for such a reactive data assimilation problem since a high correlation --- due to the neural network architecture --- with the latent field $C_\Theta$ can highly disrupt the micro-porosity recovery. Relying on the PDE model to ensure relevant correlation hence appears as the most appropriate strategy. 

\begin{figure}
    \centering
    \includegraphics[width=0.9\linewidth]{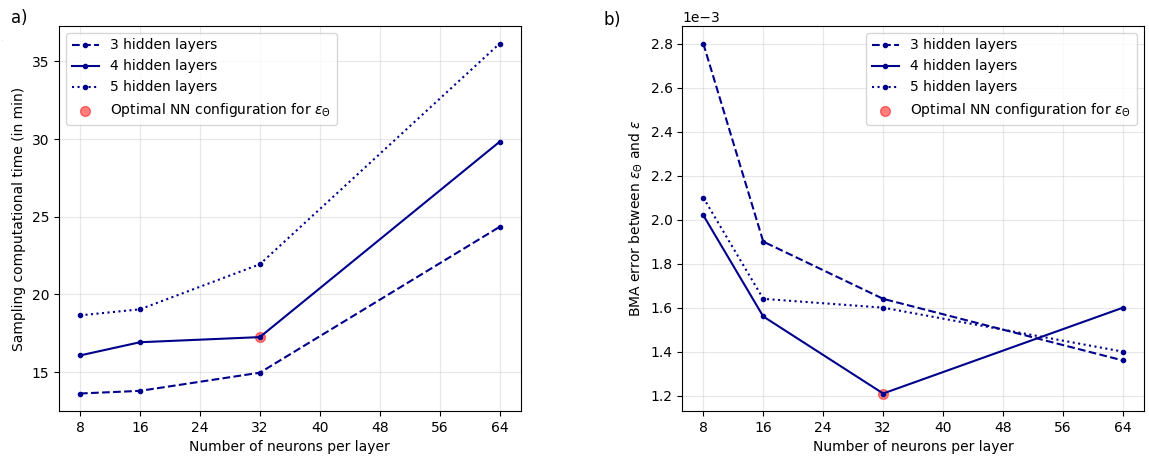}
    \caption{\textbf{Neural Network configuration choice for the surrogate model on the micro-porosity field:} a) Computational cost of sampling step 1 with respect to the neural network architectures both in terms of the number of hidden layers and neurons per layer. b) Bayesian Model Average (BMA) error between the surrogate model $\varepsilon_\Theta$ and groundtruth $\varepsilon$, computed as in equation \eqref{BMA_eps}, for different neural network architectures.}
    \label{fig:NN_architecture_1D}
\end{figure}

The first neural network establishing the micro-porosity surrogate model $\varepsilon_\Theta$ is composed of 4 hidden layers with 32 neurons per layer and a hyperbolic tangent activation function. The output layer is complemented by a rectified hyperbolic tangent $\mathrm{Tanh}^r(z) = 0.5(\mathrm{Tanh}(z)+1)$ to ensure output values between 0 and 1. This neural network complexity provides the best approximation of the micro-porosity during the first sampling step 1 while maintaining moderate computational costs. Indeed, we analyze in Fig \ref{fig:NN_architecture_1D} the impact of the neural network architecture both on the computational time spent on the sampling procedure and the Bayesian Model Average (BMA) accuracy, computed as:
\begin{equation}
    \label{BMA_eps}
     \ds \text{BMA-E}^\varepsilon = \left\| P\left(\varepsilon_\Theta\,|\,(x, t), \mathcal{D}, \mathcal{M}\right) - \varepsilon) \right\|^2_{\Omega^*\times(0,T_f^*)} =  \left\| P\left(\varepsilon_\Theta\,|\,(x, t), \mathcal{D}, \mathcal{M}\right) - (1-\Im)) \right\|^2_{\Omega^*\times(0,T_f^*)}
\end{equation}
where the notation $\|\cdot\|$ used here refers to the functional $\L^2$-norm and $P\left(\varepsilon_\Theta\,|\,(x, t), \mathcal{D}, \mathcal{M}\right)$ is the BMA approximation from equation \eqref{BMA} (\emph{e.g.} see \cite{wilson_bayesian_2020} or \cite{PEREZ2023112342} for more details). For the optimal NN configuration, one estimates the sampling computational cost, providing the overall samples of the posterior distribution \eqref{post_dist_Bayes} based on the $N_\mathrm{obs}$ training measurements, to about 17 min. The BMA prediction obtained through equation \eqref{BMA}, over the whole computational domain $\Omega^*\times(0, T_f^*)$ is meanwhile immediate --- less than 1 s on GPU and a few seconds on CPU. Recovering the latent concentration $C_\Theta$ will require less neural network expressivity compared to the micro-porosity field which needs to integrate noisy data, and unresolved morphological features in its reconstruction. In this sense, we assume that the second neural network defining the acid surrogate model is composed of 3 hidden layers with 32 neurons per layer and a hyperbolic tangent activation function. This is only one layer less compared to the first NN for $\varepsilon_\Theta$ but enables saving $1056$ parameters. The number of network parameters is, therefore, $3297$ for the first sampling and increases to $5538$ for the second and third sampling steps with the two additional inverse parameters. 

The other hyperparameters concerning the AW-HMC sampler are summarized in Table \ref{tab:AW_param_1D} for the sequential sampling steps. These sampler parameters involve, \emph{inter alia}, several adaptive steps $N$ during which the critical weights $\lambda_k$ are automatically adjusted through an Inverse Dirichlet basis using the equation \eqref{Lambda_weights_Inv_Dir}, and a number of overall sampling steps $N_s$. The two other parameters, namely $L$ and $\delta t$, are intrinsically related to the Hamiltonian Monte Carlo structure of the AW-HMC sampler. Indeed, they are involved in the deterministic step that relies on the leapfrog symplectic integrator to solve for the Hamiltonian dynamical system \eqref{Ham_dyn_sys} (see Sect. \ref{subsec:BPINNs}). We also refer to our methodological article \cite{PEREZ2023112342} for more details on these hyperparameters and especially to Algorithm 1 (Adaptively Weighted Hamiltonian Monte Carlo) for their respective role in the sampling phases. 

\begin{table}[t]
    \centering
    \renewcommand{\arraystretch}{1.25}
    \begin{tabular}{|c|c|c|c|c|}
    \hline
        \multirow{2}{*}{Sampling step} & Number of & Number of  & Number of & Leapfrog time  \\ 
                      & adaptive steps $N$ & samples $N_s$ & leapfrog steps $L$ & step $\delta t$\\ \hline
        1) Preconditioning $\varepsilon_\Theta$ & 50 & 200 & 200 & \num{1e-3}   \\ \hline
        2) Preconditioning $C_\Theta$ + Inference $\alpha_\Theta$ & 20& 200 & 200 & \num{5e-4} \\ \hline
        3) Full data assimilation & 4 & 200 & 200 & \num{2e-4} \\ \hline
    \end{tabular}
\caption{\textbf{AW-HMC hyperparameters on the 1D+Time reactive inverse problem:} Setting of the sampler hyperparameters for the three sequential sampling steps defined in Fig \ref{fig:Algorithm}. The number of adaptive steps $N$ along with the leapfrog parameters $L$ and $\delta t$ are AW-HMC sampler-specific parameters. }
    \label{tab:AW_param_1D}
\end{table}
\subsection{Numerical results}
\label{subsec:num_results_1D}

\begin{figure}
    \centering
    \includegraphics[width=\linewidth]{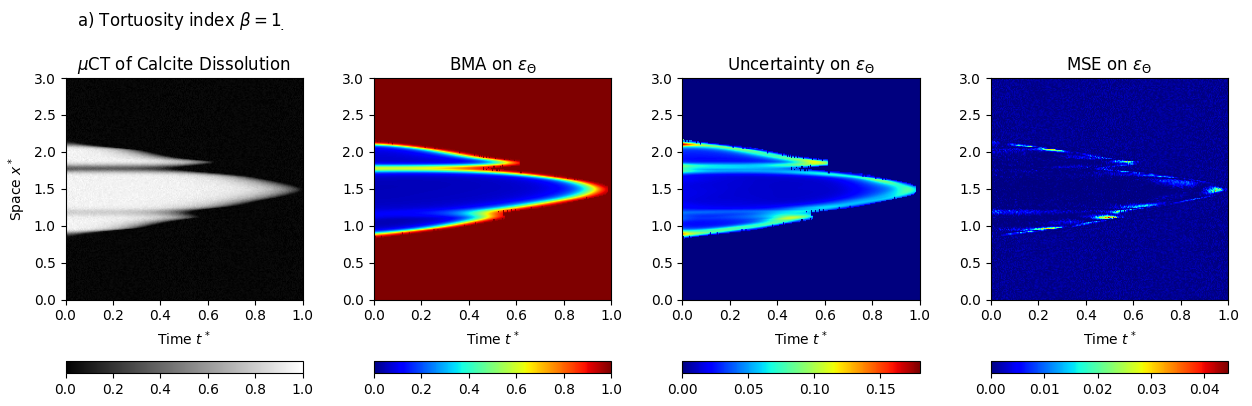}\\[2mm]
    \includegraphics[width=\linewidth]{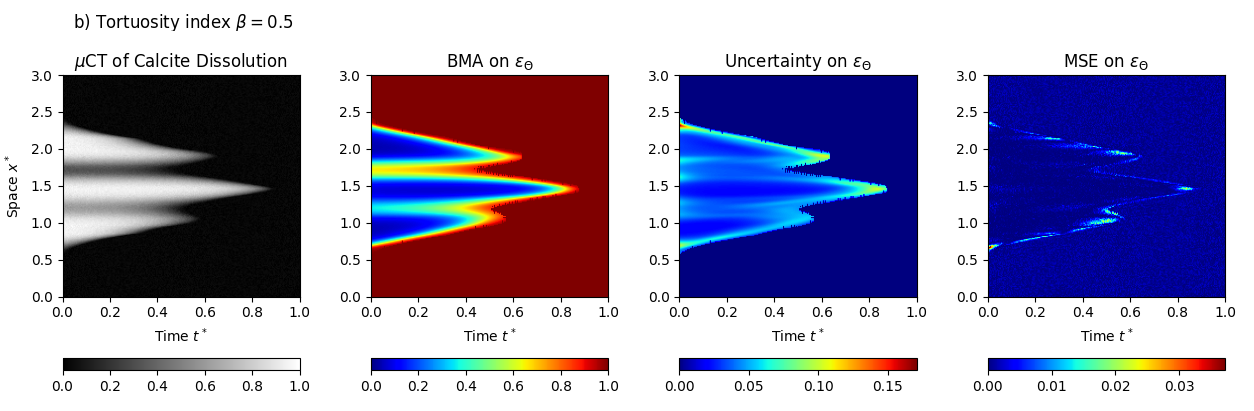}
    \caption{\textbf{Uncertainty Quantification on 1D+Time reactive inverse problem with data assimilation:} Bayesian Model Average (BMA) predictions on the micro-porosity field $\varepsilon_\Theta$ with their local uncertainties --- given by the standard deviation on the posterior distribution of the predictions --- and mean squared errors (MSE). The top row corresponds to the initial geometry from Fig \ref{fig:Initial_geo_1D}a with tortuosity index $\beta = 1$. The bottom row is related to the initial porous sample from Fig \ref{fig:Initial_geo_1D}b with $\beta = 0.5$.}
    \label{fig:1D_esp}
\end{figure}

\begin{figure}
    \centering
    \includegraphics[width=0.49\linewidth]{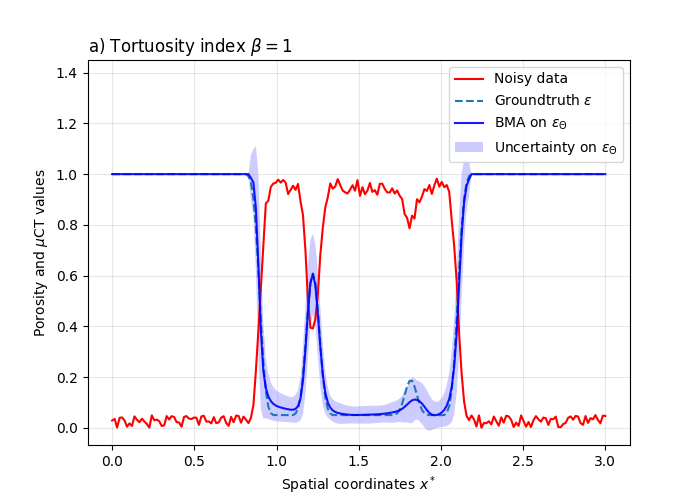}
    \includegraphics[width=0.49\linewidth]{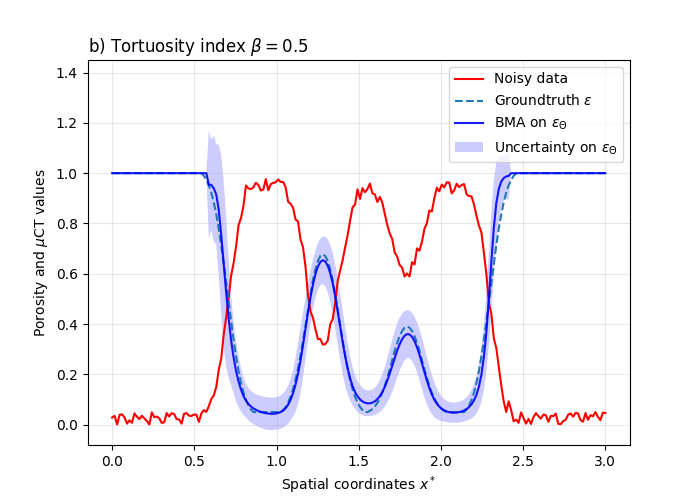}
    \caption{\textbf{Uncertainty Quantification on the micro-porosity field at the initial state ($t^*=0$) for 1D+Time reactive inverse problem:} Corrupted $\mu$CT image before dissolution, groundtruth on $\varepsilon$,  BMA prediction, and uncertainty on $\varepsilon_\Theta$ plotted along the spatial dimensionless coordinates $x^*$.  Validation test cases with tortuosity indexes a) $\beta = 1$ and b) $\beta=0.5$.}
    \label{fig:1D_esp_sliceT0}
\end{figure}

We demonstrate the validity of our data assimilation approach on synthetic inverse problems of calcite dissolution whose initial core geometries are characterized in Fig \ref{fig:Initial_geo_1D}, and for which the dynamical $\mu$CT images are provided through DNS (see Sect. \ref{subsec:1D_DNS}). In the data assimilation Bayesian framework, we first select log-normal prior distributions on $(\mathcal{P}_{\mathrm{inv}})_\Theta=\left\{\alpha_\Theta, \gamma_\Theta \right\}$, which ensures the positivity of the inverse parameters, and independent normal distribution for the neural network parameters $\theta$. Nonetheless, appropriate change of variables on the inverse parameters, namely $\bullet_\Theta = e^{\Tilde{\bullet_\Theta}}$, makes it possible to consider Gaussian prior distributions on the newly defined set of parameter $\Theta= \{\theta, \Tilde{\mathcal{P}}_{\mathrm{inv}}\}$ (\emph{e.g.} see \cite{yang_b-pinns_2021} or \cite{PEREZ2023112342}). This is the underlying hypothesis considered when defining the log-prior term in the potential energy \eqref{multi_pot_1DTime}, where we assume $P(\Theta)\sim\mathcal{N}(0, \sigma_\Theta^2 I_{p+d})$. In practice, we use the standard deviation $\sigma_\Theta = 10$ in the applications such that slightly diffuse distribution induces weakly informed priors on the $\Theta$ parameters. We also impose weakly informed priors on the inverse parameters such that we do not rely on biased a-priori on their respective scaling. In this sense, we benefit from the AW-HMC sampler advantages to handle multiscale inverse problems with unknown informative priors. We also avoid hand-tuning of the distinct task uncertainties by setting all the $\sigma_k$, $k=0...5$, to be equal in equation \eqref{multi_pot_1DTime}. On the contrary, automatic adjustment of the weighting parameters $\lambda_k$ will provide intrinsic task uncertainties during the sampling procedure.

From the overall sampling procedure, we first obtain through equation \eqref{BMA} a Bayesian Model Average prediction on the porosity field $\varepsilon_\Theta$ which is approximated by (\emph{e.g.} see \cite{wilson_bayesian_2020}): 
\begin{equation}
\label{approx_BMA}
    P\left(\varepsilon_\Theta|(x,t), \mathcal{D}, \mathcal{M}\right) \simeq \frac{1}{N_s-N}\sum_{\tau=N}^{N_s} P\left(\varepsilon_\Theta|(x,t),\Theta^{t_\tau}\right) 
\end{equation}
where $P\left(\varepsilon_\Theta\,|\,(x,t), \Theta^{t_\tau}\right)$ is the surrogate model prediction of the micro-porosity resulting from the sampling iteration $\tau$ for the set of parameters $\Theta$ --- including both the neural network and inverse parameters. Similarly, one can also compute local uncertainties on the output porosity field given the standard deviation metric on the posterior distribution of the predictions. These uncertainty quantification results are presented in Fig \ref{fig:1D_esp} along the whole dissolution time $t^*$ for both the initial core geometries with distinct tortuosity indexes. We also compare the local uncertainties on the micro-porosity field with the traditional Mean Squared Errors (MSE) between the BMA surrogate prediction obtained by equation \eqref{approx_BMA} and the groundtruth $\varepsilon$. This shows enhanced mean squared errors on the core edges during the calcite core dissolution which are, however, embedded in the local uncertainties. The latter uncertainties also tend to increase in these regions, characterizing the challenge of capturing reliable core interfaces from the dynamical $\mu$CT images. In this sense, one can query the confidence of mineral reactivity assessment using merely differential imaging techniques on the dynamical $\mu$CT scans. From the dynamical observation of the calcite core dissolution, we obtain uncertainties on the initial state geometry represented in Fig \ref{fig:1D_esp_sliceT0}. This shows that the posterior prediction on $\varepsilon_\Theta$ covers the groundtruth micro-porosity field $\varepsilon$ and provides upper and lower bounds for the residual, potentially unresolved, micro-porosity $\varepsilon_0$ estimation in the porous matrix --- 
\emph{e.g.} $1.8\%\leq\varepsilon_0\leq 9\%$ in the case $\beta = 1$ for the 95\% confidence interval, corresponding to approximately two standard deviations.

\begin{figure}
    \centering
    \includegraphics[width=0.49\linewidth]{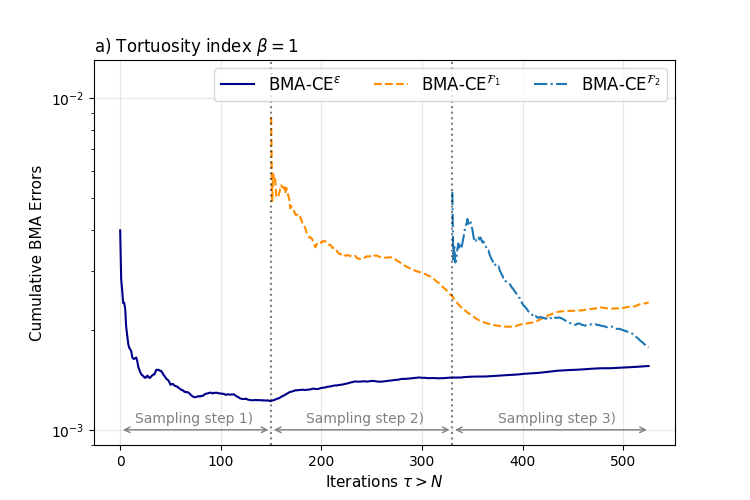}
    \includegraphics[width=0.49\linewidth]{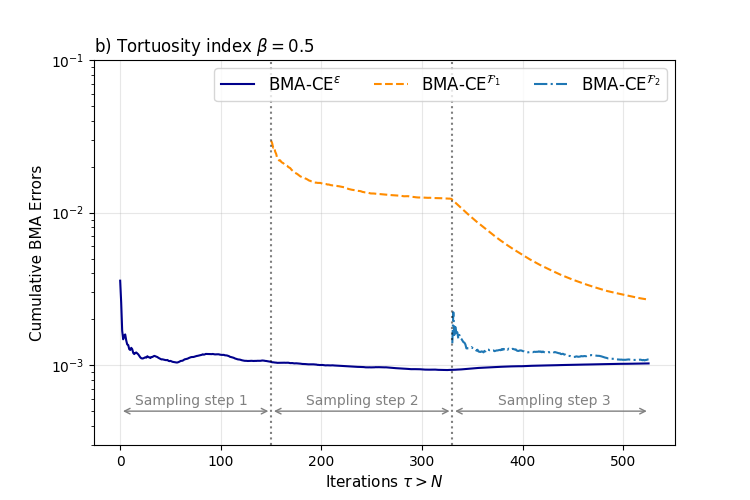} 
    \caption{\textbf{Bayesian Model Average Cumulative Error diagnostics (BMA-CE), as defined in equations \eqref{BMA_CE_1DTime_eps}-\eqref{BMA_CE_1DTime_constraints}, for 1D+Time reactive inverse problem:} BMA-CE on the micro-porosity field prediction $\varepsilon$ throughout the sampling iterations, and BMA-CE on the PDE constraint residuals $\mathcal{F}_1$ and $\mathcal{F}_2$ defined in \eqref{pde_constraints_1DTime} and introduced successively. The dotted vertical lines split the sampling steps in the sequential reinforcement of the multi-potential energy \eqref{multi_pot_1DTime}. Dissolution inverse problem on the initial geometries from a) Fig \ref{fig:Initial_geo_1D}a with tortuosity index $\beta = 1$ and b) Fig \ref{fig:Initial_geo_1D}b with tortuosity index $\beta = 0.5$.}
    \label{fig:BMA_CE_1D}
\end{figure}

Moreover, we rely on the Bayesian Model Average Cumulative Error metric, denoted BMA-CE and introduced in \cite{PEREZ2023112342}, to quantify the sampling efficiency in terms of convergence along the marginalization process. We first compute the BMA-CE diagnostics for the porosity field $\varepsilon$ based on:
\begin{equation}
\label{BMA_CE_1DTime_eps}
        \ds \text{BMA-CE}^\varepsilon(\tau) = \left\|\ds \frac{1}{\tau-N}  \sum_{i=N}^\tau P\left(\varepsilon_\Theta\,|\,(x,t), \Theta^{t_i}\right) -(1-\Im)\right\|^2 \qquad \forall \tau > N.
\end{equation}
 Equation \eqref{BMA_CE_1DTime_eps} hence defines for each sampling step $\tau$, after the adaptive steps, a cumulative error characterizing the convergence of the BMA model toward the groundtruth $\varepsilon$. Such a diagnostic is computed during the overall sampling procedure, covering the three successive steps of sequential reinforcement. In the same manner, we extend this notion of convergence to the PDE constraints by computing the BMA-CE metric on their respective residuals. We, therefore, introduce $\mathcal{F}_1$ and $\mathcal{F}_2$ the PDE constraints residuals arising from the reactive model \eqref{eq:inverse_chem_adim_Vf} and involved in the multi-potential energy \eqref{multi_pot_1DTime}: 
\begin{equation}
    \label{pde_constraints_1DTime}
    \begin{aligned}
        & \mathcal{F}_1(\varepsilon_\Theta, C_\Theta) := \alpha_\Theta \frac{\partial \varepsilon_\Theta}{\partial t} - C_\Theta \\[2mm]
        & \mathcal{F}_2(\varepsilon_\Theta, C_\Theta) := \gamma_\Theta\left( \frac{\partial C_\Theta}{\partial t} + 
    \frac{\partial \varepsilon_\Theta}{\partial t} \right) - \left(\varepsilon_\Theta \frac{\partial^2 C_\Theta}{\partial x^2} - C_\Theta \frac{\partial^2 \varepsilon_\Theta}{\partial x^2} \right) 
    \end{aligned}
\end{equation}
to finally define their corresponding diagnostics $\text{BMA-CE}^{\mathcal{F}_\bullet}$:
\begin{equation}
\label{BMA_CE_1DTime_constraints}   
    \ds \text{BMA-CE}^{\mathcal{F}_\bullet}(\tau) = \left\|\ds \frac{1}{\tau-N}  \sum_{i=N}^\tau P\left(\mathcal{F}_\bullet(\varepsilon_\Theta, C_\Theta) \,|\,(x,t), \Theta^{t_i}\right) \right\|^2 \qquad \forall \tau > N.
\end{equation}
These metrics are respectively computed on the sampling steps 2 and 3 for the residuals $\mathcal{F}_1$ and $\mathcal{F}_2$, and the resulting convergence curves are presented in Fig \ref{fig:BMA_CE_1D} for both initial geometries. Successively introducing the additional PDE constraints results in deviations in the $\text{BMA-CE}^\varepsilon$ curve compared to the purely data-fitting sampling step 1. As we aim for a physics-based data assimilation of imperfect $\mu$CT data and seek to incorporate information on the static initial data through the combined use of observations and dissolution modeling, we can anticipate that introducing physics-based constraints will alter the BMA-CE predictions on the porosity field. The slight increase observed in the $\text{BMA-CE}^\varepsilon$ curves in Fig \ref{fig:BMA_CE_1D} highlights that merely considering the regression step 1 leads to underestimated uncertainties and tends to overfit the data. This means that the PDE model constraints, successively introduced in step 2 and 3, brings information and dynamical insights into the porosity field $\varepsilon$ recovery instead of providing overfitting predictions. One also gets that the PDE constraints are satisfied by the convergence of their residual BMA-CE curves. 

\begin{figure}
    \centering
    \includegraphics[width=0.97\linewidth]{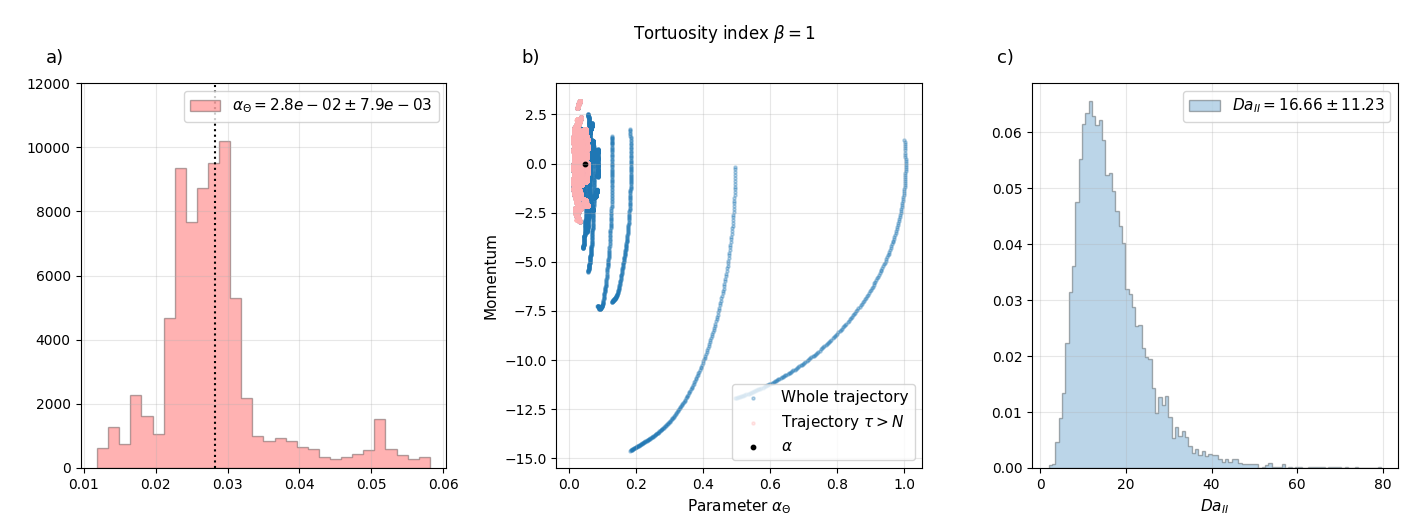}\\[1mm]
    \includegraphics[width=0.97\linewidth]{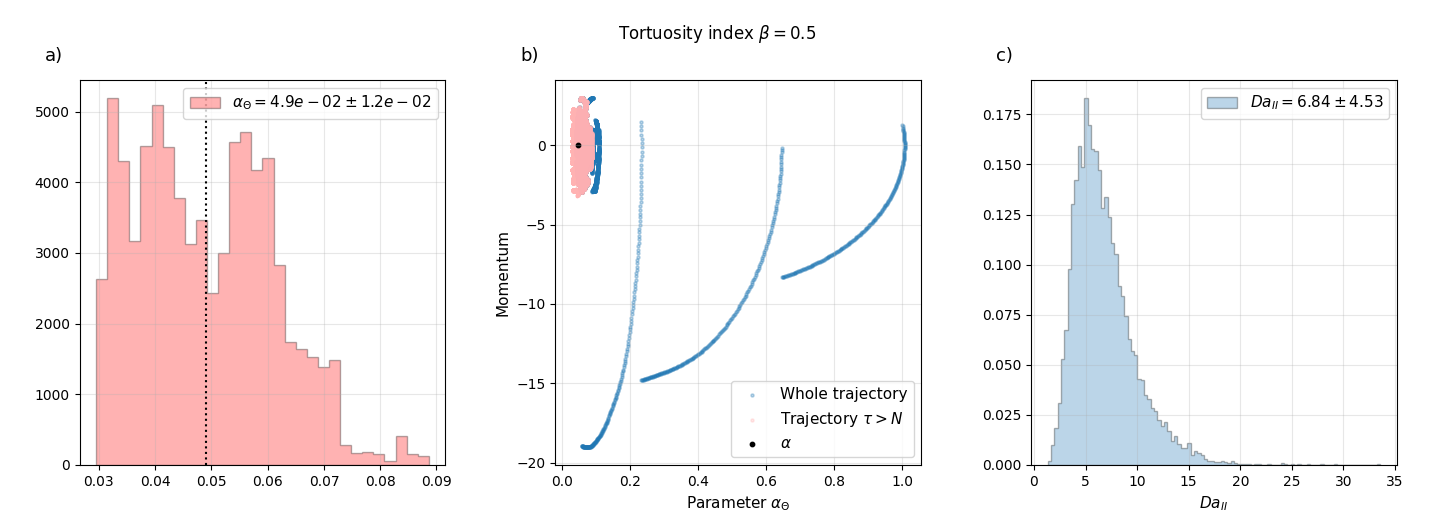}
    \caption{\textbf{Posterior distributions of 1D+Time reactive inverse problem:} a) Histogram of the marginal posterior distributions for the inverse parameter $\alpha_\Theta$. b) Phase diagrams of its trajectory throughout the sampling, with the adaptive steps trajectories (in blue) and effective sampling (in red). The groundtruth values of $\alpha$ are represented by the black dots. c) Resulting posterior distributions of the catalytic Damköhler number $\Da2$, determined through the log-normal distribution from relation \eqref{eq:dist_da2}. The top and bottom row respectively corresponds to the tortuosity indexes $\beta = 1$ and $\beta = 0.5$.}
    \label{fig:Hist_1D}
\end{figure}

Regarding the inverse parameters inference, we represent in Fig \ref{fig:Hist_1D}a the histograms of the marginal posterior distributions of $\alpha_\Theta:=(\Da2^*)^{-1}$ for the two initial geometries from Fig \ref{fig:Initial_geo_1D}. These distributions are obtained throughout the sampling steps 2 and 3, and provide intrinsic uncertainties on the parameter estimations. One also gets from Fig \ref{fig:Hist_1D}b the parameter trajectories when exploring the phase space distribution \eqref{joint_dist}, where the convergence toward the mode during the adaptive steps is represented in blue. The final sampling, corresponding to the phase diagram trajectories for $\tau>N$, thus ensures an efficient exploration of the parameter mode neighborhood. Following the sequential reinforcement strategy detailed in Sect. \ref{sec:Method}, one gets at the end of the sampling step 2 a description of the restricted \RAI$^-$ domain altogether with an initial estimate on the distribution of $D_m^*$. 

\begin{figure}
    \centering
    \includegraphics[width=\linewidth]{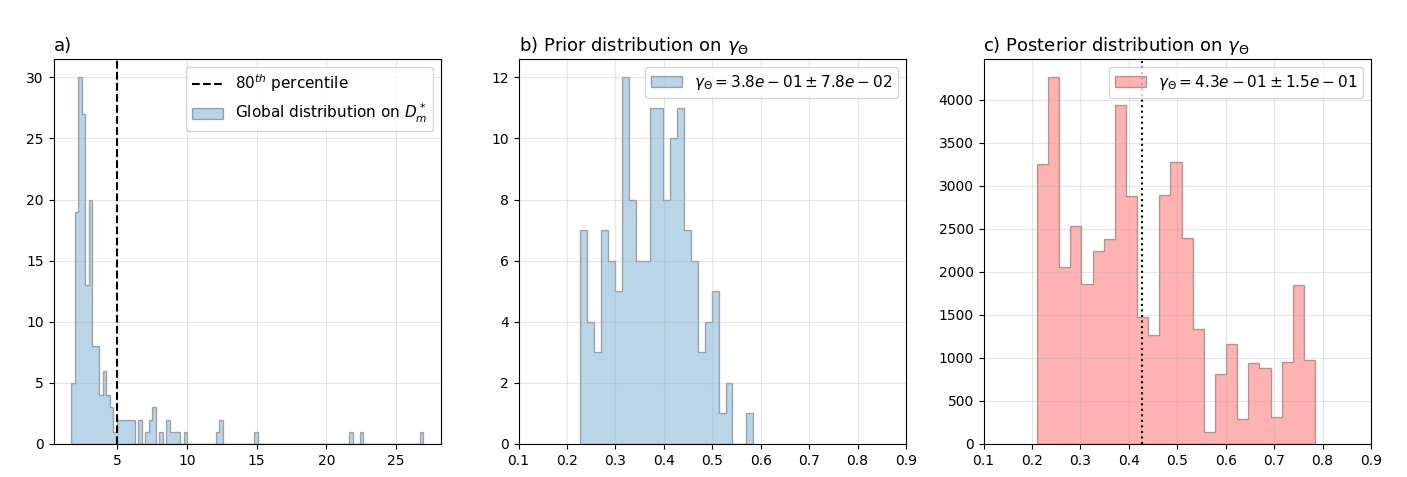}
    \caption{\textbf{Prior and posterior distributions on the inverse parameter $\gamma_\Theta$ for 1D+Time reactive inverse problem with tortuosity index $\beta = 1$:} a) Global distribution on the estimated parameter $D_m^*$ arising from a-posteriori analysis on the sampling step 2. We discard the distribution tail after the vertical dashed line corresponding to the $80^\text{th}$ percentile. b) Resulting prior distribution on the inverse parameter $\gamma_\Theta$, used as an a-priori in step 3. c) Posterior distribution on $\gamma_\Theta$ obtained from the overall data assimilation problem throughout sampling step 3.}
    \label{fig:Hist_1D_gamma}
\end{figure}

The latter is regarded as an initial a-priori on the parameter $\gamma_\Theta$ in step 3, and is obtained as follows: 
\begin{itemize}
    \item for the sampling iteration $\tau=N...N_s$ in step 2, we compute the (cumulative) predictive BMA distributions of the two operators \[\ds\frac{\partial C_\Theta}{\partial t} + \frac{1}{\upsilon C_0}\frac{\partial\varepsilon_\Theta}{\partial t} \quad \text{and}\quad \mathcal{D}_i(\varepsilon_\Theta, C_\Theta), \]
    \item we establish the \RAI$^-_\tau$ as the admissible points of the $\RAI$ where $D_m^*$ is predicted positive for each sampling iteration $\tau>N$,
    \item we compute a spatially averaged estimation of the parameter $ \left(\bar{D_m^*}\right)_\tau$ on this eligible domain, where
    \[\ds \left(\bar{D_m^*}\right)_\tau = \frac{1}{\#\RAI^-_\tau} \sum_{k\in\RAI^-_\tau} (D_m^*)_{\tau}(x_k, t_k) \qquad \forall \tau>N,\]
    \item we can estimate a global distribution on $D_m^*$ throughout the overall samples of step 2 where we discard the distribution tail after the $80^{\text{th}}$ percentile,
    \item we finally evaluate the prior distribution on the inverse parameter $\gamma_\Theta:=(D_m^*)^{-1}$ with its mean value $\bar{\gamma_\Theta}$ used as an initial a-priori in step 3,
\end{itemize}
In the case $\beta = 1$ for instance, one gets the global distribution on $D_m^*$ shown in Fig \ref{fig:Hist_1D_gamma}a, which translates into the prior distribution on $\gamma_\Theta$ given by Fig \ref{fig:Hist_1D_gamma}b with $\bar{\gamma_\Theta}=\num{3.8e-1}$. Finally, we obtain the posterior distribution from the overall data assimilation problem throughout sampling step 3. This results in the distribution on the inverse parameter $\gamma_\Theta$ represented in Fig \ref{fig:Hist_1D_gamma}c, with the uncertainty range $\gamma_\Theta\in[0.274,0.582]$.

Finally, one can estimate the posterior distribution on the catalytic Damköhler number $\Da2$ resulting from the overall data assimilation problem on dynamical $\mu$CT images. This comes from the observation that each inverse parameter, namely $\alpha_\Theta$ and $\gamma_\Theta$, is sought according to a log-normal distribution through the change of variable $\bullet_\Theta = e^{\Tilde{\bullet_\Theta}}$. Hence, we obtain two normal posterior distributions on the random variables $X_1$ and $X_2$ respectively associated with $\ln(\alpha_\Theta)$ and $\ln(\gamma_\Theta)$ such that $X_1 \sim \mathcal{N}\left(\mu_\alpha, \sigma_\alpha^2\right)$ and $X_2\sim \mathcal{N}\left(\mu_\gamma, \sigma_\gamma^2\right)$. This combines into a normal distribution on  $\ln(\gamma_\Theta / \alpha_\Theta)$ given by $(X_1 - X_2)\sim \mathcal{N}\left(\mu_\gamma - \mu_\alpha, \sigma_\gamma^2 +\sigma_\alpha^2\right)$, which is nothing more than a log-normal posterior distribution on the Damköhler number $\Da2$. Indeed, one gets that the random variable $X$ related to the dimensionless $\Da2$ number follows
\begin{equation}
\label{eq:dist_da2}
    X \sim \mathrm{Log-}\mathcal{N}\left(\mu_\gamma - \mu_\alpha, \sigma_\gamma^2 +\sigma_\alpha^2\right) := \mathrm{Log-}\mathcal{N}\left(\mu, \sigma^2 \right),
\end{equation}
whose mean and variance are respectively computed as 
\begin{equation}
\mathbb{E}[X] = e^{\mu+\sigma^2/2} \quad \text{and}\quad \mathrm{Var}(X) = e^{2\mu+\sigma^2}\left(e^{\sigma^2} -1\right).
\end{equation}
The resulting posterior distributions on the Damköhler number $\Da2$ are represented in Fig \ref{fig:Hist_1D}c for both the initial calcite core geometries with their related tortuosity indexes. We finally obtain the uncertainty ranges $\Da2\in[5.43, 27.89]$ and $\Da2\in[2.30,11.37]$ for the tortuosity indexes $\beta = 1$ and $\beta = 0.5$ respectively.

\section{Pore-scale imaging inverse problem of calcite dissolution: 2D+Time application}
\label{sec:Results}

In this section, we apply the data assimilation methodology developed in Sect. \ref{sec:Method} to inverse problems for calcite dissolution with heterogeneous porosity levels. We consider a more realistic application involving the dissolution of a 2D calcite core following the configuration of the benchmark developed in \cite{molins_simulation_2021}. This test case can provide a basis for reactive inverse problems in isotropic porous samples, although the $\mu$CT measurements are still synthetic observations resulting from DNS altered with noise. 

\subsection{Problem set up and dimensionless inverse formulation}

\begin{figure}
    \centering
    \includegraphics[width = 0.5\linewidth]{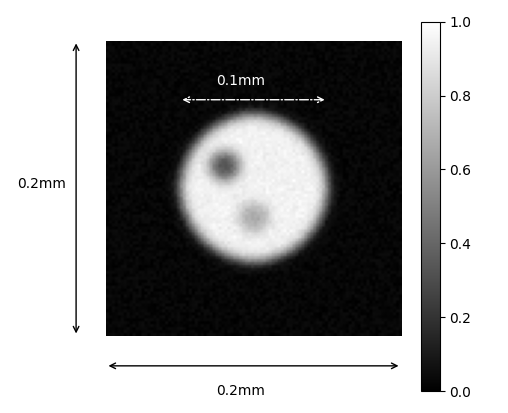}
   \caption{\textbf{Initial $\mu$CT image defining the 2D porous sample geometry:} Synthetic case with tortuosity index $\beta = 0.5$. The $\mu$CT measurements are normalized, corrupted with noise, and provide the observation dataset before the dissolution process. The cylindrical calcite core  has a radius equal to $0.05$ mm.}    
   \label{fig:Geo_2D}
\end{figure}

We consider a 2D calcite crystal with a cylindrical shape, heterogeneous porosity levels, and two apertures, whose initial geometry is defined by the numerical $\mu$CT image from Fig \ref{fig:Geo_2D}, corrupted with Gaussian noise. We define the physical domain $\Omega\subset\mathbb{R}^2$ of width $0.2$ mm corresponding to the two-dimensional flow channel surrounding the calcite core. We first solve the direct formulation of the dissolution process for this initial geometry on a Cartesian spatiotemporal grid of resolution $N_x = N_y = 100$ and $N_t = 350$. We assume a diffusive-dominated regime with continuous acid injection through non-homogeneous Dirichlet boundary conditions on $\partial \Omega$. We also consider, as in the 1D+Time validation test cases, a strong acid solution with $\mathrm{pH} = 0$ such that the normalizing constant is given by $C_0 = 1 \, \mathrm{mol.L^{-1}}$. The characteristic length $L$ of these porous samples is set to $L=0.1$ mm and the reactive parameters are respectively defined by $K_s = 0.8913\, \mathrm{mol.m^{-2}.s^{-1}}$, $D_m = \num{e-9}\, \mathrm{m^{2}.s^{-1}}$,  and $\gamma_{\ce{H+}} = \num{e-3}\, \mathrm{m^{3}.mol^{-1}}$, taken from the benchmark \cite{molins_simulation_2021}. The reactive specific area $A_s$ is set to $A_s = \num{e3}\,\mathrm{m}^{-1}$ --- which is slightly underestimated compared to the computed value around $\num{7.4e3}\,\mathrm{m}^{-1}$. We account for the calcite molar volume $\upsilon = \num{36.93e-3}\, \mathrm{L.mol^{-1}}$, and set a tortuosity index $\beta = 0.5$. The DNS is performed until the overall calcite is dissolved which corresponds to a characteristic final time $T_f = 175$ s. Taken together, one gets a sequence of synthetic $\mu$CT images $\mathfrak{Im}_{i,j}$ characterizing the dissolution process of the calcite core on the spatiotemporal domain $\Omega\times(0, T_f)$.

Given these reactive parameters, we identify the same dissolution regime as in the 1D+Time test cases, with a catalytic Damkölher given by $\Da2 = 8.913$. The inverse formulation, however, results in distinct dimensionless numbers, namely $\Da2^*$ and $D_m^*$, arising from the scaling dimensionless factor $D_\mathrm{ref}$. This scaling factor is here determined by:
\begin{equation}
    D_\mathrm{ref} = \frac{T_f^* L^2}{T_f} = \num{5.714e-11}\, \mathrm{m^2.s^{-1}},
\end{equation}
where the final dimensionless time is $T_f^*=1$. Therefore, one gets from equation \eqref{eq:dim_numb_inv} the following dimensionless numbers characterizing this 2D+Time reactive inverse problem: 
\begin{equation}
    \ds \Da2^* = 155.9775 \quad  \text{ and } \quad D_m^* = 17.5
\end{equation}
which are related to the inverse parameters through $(\mathcal{P}_\mathrm{inv})_\Theta=\left\{\alpha_\Theta, \gamma_\Theta \right\} = \left\{ (\Da2^*)^{-1}, (D_m^*)^{-1} \right\}$. We consider the dimensionless domain $\Omega^*\times(0,T_f^*) = [0,2]\times[-1,1]\times(0,1)$, given the characteristic length $L$, to extract the observation dataset 
\begin{equation}
    \mathcal{D} = \left\{  (x_k,y_k, t_k) \in [0,2]\times[-1,1]\times(0,1) , \quad k=1...N_{\text{obs}} \right\}
\end{equation}
where the number of training points $N_\text{obs} = 15907$ represents less than 1\% of the data required for the full field reconstructions on the spatiotemporal domain $\Omega^*\times(0,T_f^*)$. 
This dataset $\mathcal{D}$ is then divided into $\mathcal{D}^S$, $\mathcal{D}^F$, $\mathcal{D}^\text{RAI}$, $\text{RAI}^+$, $\text{RAI}^-$ and $\mathcal{D}^\partial$ that respectively cover around 15.5\%, 0.5\%, 48\%, 11\%, 20\% and 5\% of the $N_\text{obs}$ training measurements. Finally, the multi-potential energy writes on this dataset decomposition:
\begin{equation}
\label{multi_pot_2DTime}
    \begin{aligned}
    \ds U(\Theta) & =  \frac{\lambda_0}{2\sigma_0^2}\left\|1-\varepsilon_\Theta - \Im\right\|_{\mathcal{D}^S}^2
    +  \frac{\lambda_1}{2\sigma_1^2}\left\|1-\varepsilon_\Theta - \Im\right\|_{\text{RAI}^+}^2
    +\frac{\lambda_2}{2\sigma_2^2}\left\|  (\upsilon C_0)^{-1}\alpha_\Theta \frac{\partial \varepsilon_\Theta}{\partial t} - C_\Theta \right\|_{\text{RAI}}^2\\[2mm]  
    & +  \frac{\lambda_3}{2\sigma_3^2}\left\| \gamma_\Theta \frac{\partial C_\Theta}{\partial t} - \Delta C_\Theta \right\|_{\mathcal{D}^F}^2 
    + \frac{\lambda_4}{2\sigma_4^2} \left( \left\| 1 -C_\Theta \right\|_{\mathcal{D}^\partial}^2+  \left\| \num{e-7} -C_\Theta \right\|_{\mathcal{D}^S}^2 \right) \\[2mm]
    & + \frac{\lambda_5}{2\sigma_5^2}\left\| 
    \gamma_\Theta\left( \frac{\partial C_\Theta}{\partial t} + (\upsilon C_0)^{-1}
    \frac{\partial \varepsilon_\Theta}{\partial t} \right) - \mathcal{D}_i(\varepsilon_\Theta, C_\Theta)
    \right\|_{\text{RAI}^-}^2 + \frac{1}{2\sigma_\Theta^2}\|\Theta\|^2 
    \end{aligned}
\end{equation}
where the heterogeneous diffusion operator $\mathcal{D}_i(\varepsilon_\Theta, C_\Theta)$ is computed in its developed form \eqref{eq:Diff_op_dev} with $\beta = 0.5$. This overall potential energy is sequentially reinforced throughout three successive sampling steps, as detailed in Sect. \ref{subsec:seq_reinforce} and validated in Sect. \ref{sec:Results1D} on 1D+Time data assimilation problems.

\subsection{Deep learning framework and computational efficiency}

Regarding the deep learning strategy, the framework is kept identical to the 1D+Time validation test cases (see Sect. \ref{subsec:deep_learning_1D}). In this sense, we consider two distinct neural network architectures, for the micro-porosity and acid concentration surrogate models, which are respectively composed of 4 and 3 hidden layers with 32 neurons per layer. The number of network parameters is, therefore, $3297$ for the first sampling and $5538$ for the second and third sampling steps with the two additional inverse parameters. The setting of the AW-HMC sampler hyperparameters is also summarized in Table \ref{tab:AW_param_2D} for the successive sampling steps. 

Besides, we investigate the impact of the problem dimensionality by analyzing the computational efficiency of the present data assimilation approach with sequential reinforcement process. In this sense, we compare the computational costs of the three successive sampling steps on the 1D+Time and 2D+Time reactive inverse problems. The results of these computational time measurements are presented in Table \ref{tab:Comp_time_comp} for both configurations. The first columns compare the sampling times, which is the time required to provide the overall samples of the posterior distributions using the AW-HMC sampler. This training phase is performed on the $N_\mathrm{obs}$ observation data which are randomly selected and non-uniformly distributed on the whole Cartesian grids --- respectively $N_x\times N_t = 200\times 240$ in 1D+Time and $N_x\times N_y \times N_t=100\times100\times 350$ in 2D+Time. All the successive sampling steps are performed on GPU devices, and the computational times are expressed in hours, minutes and seconds (hh:mm:ss). In these sampling/training phases, the present methodology does not suffer from the curse of dimensionality. The 1D+Time and 2D+Time data assimilation problems present similar computational times, although the number of training observations is two times larger in 2D+Time. This establishes that most of the computational cost of the problem is correlated to the number of neural network parameters --- as already confirmed in Sect. \ref{subsec:deep_learning_1D} and more specifically in Fig \ref{fig:NN_architecture_1D} --- rather than the number of training points. Since it appears that the same neural network architecture as in 1D+Time is significant to describe the 2D+Time inverse problem, the computational efficiency of this 2D+Time data assimilation is significantly improved. 
\begin{table}[t]
    \centering
    \renewcommand{\arraystretch}{1.25}
    \begin{tabular}{|c|c|c|c|c|}
    \hline
        \multirow{2}{*}{Sampling step} & Number of & Number of  & Number of & Leapfrog time  \\ 
                      & adaptive steps $N$ & samples $N_s$ & leapfrog steps $L$ & step $\delta t$\\ \hline
        1) Preconditioning $\varepsilon_\Theta$ & 50 & 200 & 150 & \num{1e-3}   \\ \hline
        2) Preconditioning $C_\Theta$ + Inference $\alpha_\Theta$ & 40& 200 & 150 & \num{5e-4} \\ \hline
        3) Full data assimilation & 10 & 200 & 150 & \num{3e-4} \\ \hline
    \end{tabular}
\caption{\textbf{AW-HMC hyperparameters on the 2D+Time reactive inverse problem:} Setting of the sampler hyperparameters for the three sequential sampling steps defined in Fig \ref{fig:Algorithm}. The number of adaptive steps $N$ along with the leapfrog parameters $L$ and $\delta t$ are AW-HMC sampler-specific parameters. }
    \label{tab:AW_param_2D}
\end{table}
\begin{table}[t]
    \centering
    \renewcommand{\arraystretch}{1.25}
    \begin{tabular}{|c|c|c|c|c|}
    \multicolumn{3}{l}{a) Computational times for 1D+Time data assimilation}\\[4pt]
    \hline
        & Sampling time $T_{\mathrm{GPU}}$ (in hh:mm:ss) & Prediction time $T_{\mathrm{CPU}}$ (in hh:mm:ss) \\ \hline
        Sequential step 1 & 00:17:13 & 00:00:04   \\ \hline
         Sequential step 2 & 01:38:01 & 00:00:13 \\ \hline
         Sequential step 3 & 02:38:58 & 00:00:14  \\ \hline
        \multicolumn{3}{l}{}\\
        \multicolumn{3}{l}{b) Computational times for 2D+Time data assimilation}\\[4pt]
    \hline
      & Sampling time $T_{\mathrm{GPU}}$ (in hh:mm:ss) & Prediction time $T_{\mathrm{CPU}}$ (in hh:mm:ss) \\ \hline
         Sequential step 1 & 00:16:26 & 00:04:59   \\ \hline
         Sequential step 2 & 01:21:01 & 00:34:34  \\ \hline
         Sequential step 3 & 02:41:10 & 00:41:18  \\ \hline
    \end{tabular}
\caption{\textbf{Computational times of the successive sequential steps on the 1D+Time and 2D+Time inverse problem:} Comparison of the sampling times on GPU devices (first columns), and the prediction times on CPU devices (second columns) between 1D+Time and 2D+Time configurations. All the computational times are expressed under the form hh:mm:ss to ease the readability. }
    \label{tab:Comp_time_comp}
\end{table}

The second columns of Table \ref{tab:Comp_time_comp} then compare the prediction time on CPU devices. This corresponds to the computational time necessary for the potential energy estimation and output field predictions on the whole domain $\Omega^*\times(0,T_f^*)$, along with the computation of the main differential operators required as additional outputs. Among the additional outputs, one finds the porosity time derivative at the end of step 1, and all the first-order derivatives and Laplacian operators for the porosity and concentration field at the end of step 2 --- which are used to evaluate the initial a-priori on $\gamma_\Theta$. The first-order derivatives and Laplacian operators are also considered as output in step 3 to perform the a-posterior analysis based on the BMA-CE diagnostics. In contrast to the sampling phase, the computational time devoted to these predictions is larger when dimensionality increases. This comes from the observation that one needs to evaluate the output fields and additional differential operators on a Cartesian grid which is about 72 times larger in 2D+Time. The predictions of all these differential operators on the overall domain  $\Omega^*\times(0,T_f^*)$ are achieved through automatic differentiation and hence are consistent with their evaluations along the training phase. This can straightforwardly be replaced by other standard differential schemes --- as finite differences or PSE schemes --- to evaluate these operators on the Cartesian grid, using merely the predictive porosity and acid concentration fields. However, the computational improvement of the process may not be that significant, since one needs to evaluate these operators for all the $N_s$ samplings steps --- basically 600 times --- to take into account their intrinsic uncertainties. This means that it only takes a few seconds per sample to evaluate these differential operators through automatic differentiation, which is thus comparable to other usual schemes. On top of that, the prediction phases here occur on CPU devices due to memory usage that is not marginal. Typically the micro-porosity field prediction on its own required the storage of $N_s\times N_x\times N_y\times N_t$ floats in 2D+Time, which is equivalent to about $1.9$GB. In this sense, both memory usage and computational efficiency of this prediction phase could be improved, and efforts must be made in this direction. In prospect, we would like to benefit from the parallel architecture of GPU devices by investigating and using appropriate domain decompositions of $\Omega^*\times(0,T_f^*)$.


\subsection{Results and discussion}
\begin{figure}
    \centering
    \includegraphics[width=0.9\linewidth]{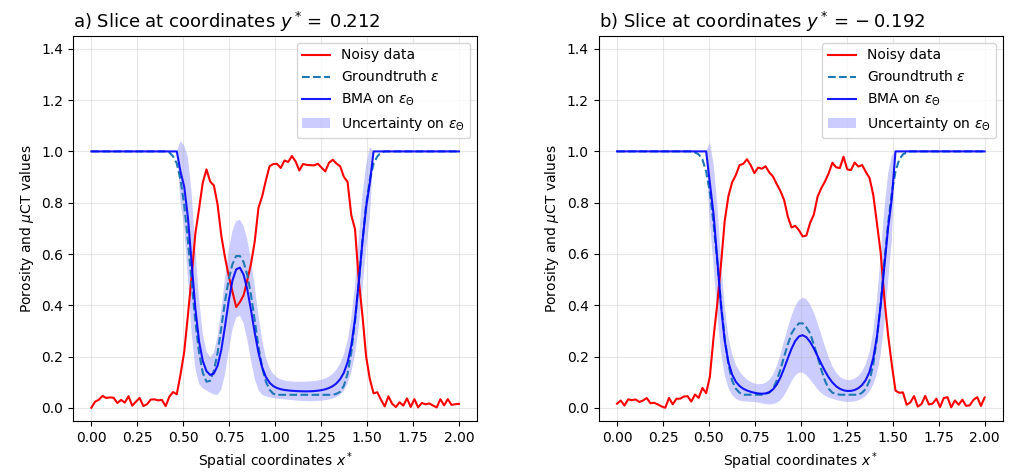}
    \caption{\textbf{Uncertainty Quantification on the micro-porosity field at the initial state ($t^*=0$) for 2D+Time reactive inverse problem:} Corrupted $\mu$CT image before dissolution, groundtruth on $\varepsilon$,  BMA prediction, and uncertainty on $\varepsilon_\Theta$. The results are plotted along the horizontal white dashed lines from Fig \ref{fig:BMA_2D}, at spatial coordinates a) $y^*=-0.192$ and b) $y^*=0.212$.}   \label{fig:Slices_t0_2D}
\end{figure}

We apply our data assimilation approach with sequential reinforcement of the multi-potential energy on this 2D+Time reactive inverse problem of calcite dissolution, based on synthetic dynamical $\mu$CT observations generated by DNS. We guarantee the positivity of the inverse parameter inference by selecting log-normal prior distributions and applying the same change of variable $\bullet_\Theta=e^{\Tilde{\bullet_\Theta}}$ as in the validation Sect. \ref{sec:Results1D}. This is combined with independent normal distribution on the neural network parameters, such that we assume the overall prior distribution $P(\Theta)\sim\mathcal{N}(0, \sigma_\Theta^2 I_{p+d})$. We also impose weakly informed priors on the inverse parameters since we do not impose a-priori information on their respective scaling.

\begin{figure}
    \centering
    \includegraphics[width=\linewidth]{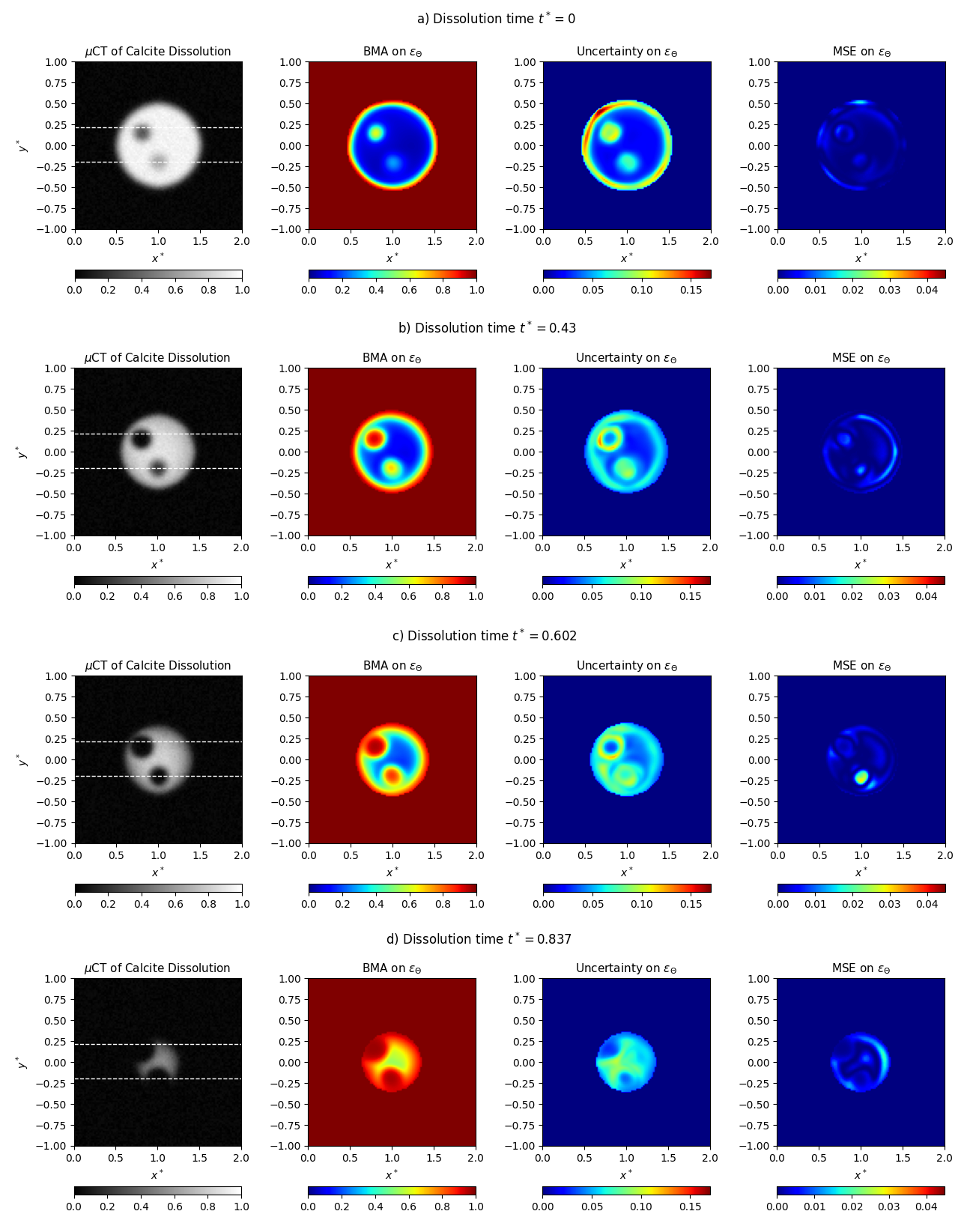}
    \caption{\textbf{Uncertainty Quantification on 2D+Time reactive inverse problem with data assimilation:} Bayesian Model Average (BMA) predictions on the micro-porosity field $\varepsilon_\Theta$ with their local uncertainties and mean squared errors (MSE). Comparison with the $\mu$CT dynamical images at several dissolution times, in the dimensionless formulation: a) Initial condition at $t^* = 0$. Intermediate dissolution times at b) $t^*=0.43$, c) $t^* = 0.602$ and d) $t^*=0.837$. }    
    \label{fig:BMA_2D}
\end{figure}

At the end of the sequential sampling, one gets the Bayesian Model Average prediction on the porosity field $\varepsilon_\Theta$, approximated as in equation \eqref{BMA_eps}, along with its local uncertainties during the whole dissolution process. These uncertainty quantification results are presented in Fig \ref{fig:BMA_2D} for several dissolution times, including the initial condition for $t^*=0$ in Fig \ref{fig:BMA_2D}a. We compare these results with the synthetic $\mu$CT images and mean squared errors computed between the BMA surrogate prediction and the groundtruth $\varepsilon$. Similar to the 1D+Time application, we observe enhanced uncertainties on the calcite core interfaces, including the aperture edges, throughout the dissolution process. These regions of higher uncertainties mainly correspond to the reactive areas of interest, where the dynamic behavior is predominant, highlighting the challenge of capturing reliable core interfaces and asperities from dynamical $\mu$CT. This also establishes the physical effects of these edge uncertainties, as uncertainties in the interface evolution are driven by the dynamical process of dissolution, its regime, and therefore the inverse parameters identification. In particular, the 2D+Time application presents an asymmetry in the spatial distribution of the main asperities, which contributes to explain the slight asymmetry observed in the uncertainty along the edge of the core in Fig \ref{fig:BMA_2D}a at early times.

\begin{figure}
    \centering
    \includegraphics[width=\linewidth]{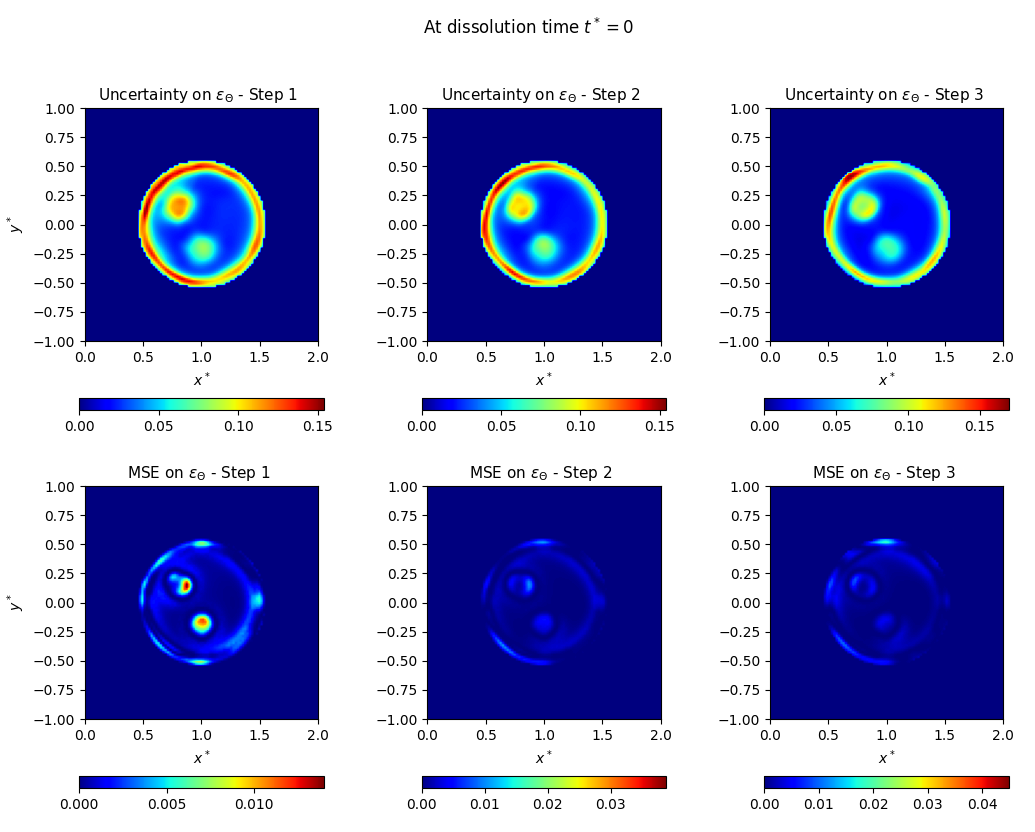}
    \caption{\textbf{Evolution of the Uncertainty Quantification on 2D+Time reactive inverse problem during sequential reinforcement:} Local uncertainty predictions on the micro-porosity field $\varepsilon_\Theta$ with their mean squared errors (MSE) at initial conditions $t^* = 0$. Comparison across the three steps of the sequential reinforcement approach, highlighting the evolution of uncertainties at the core edges and the effects of the spatial distribution of asperities in the physics-based data assimilation. Step 3 corresponds to the entire reactive inverse problem, incorporating all the physical constraints from the coupled PDE system \eqref{eq:inverse_chem_adim_Vf}.}
    \label{fig:BMA_2D_evol_seq}
\end{figure}

Indeed, when comparing the uncertainty at the initial time between the entire data assimilation process - resulting from the three sequential steps of reinforcement learning - and the first regression step, we observe that the uncertainty distribution along the edges differs. At the end of the first sequential step, considering the $\mu$CT data only, we tend to recover a homogeneous distribution of the uncertainty along the edge of the core (see Fig \ref{fig:BMA_2D_evol_seq}), which remains the area of higher uncertainties. The heterogeneous distribution of the uncertainty along the edges is enhanced by the addition of successive physics-based constraints, confirming the physical behavior of such an asymmetry due to geometrical effects in the dissolution process. In particular, the area of higher uncertainty in Figure \ref{fig:BMA_2D}a corresponds to the calcite core interface located in the neighborhood of the largest asperity, involving more complex local interactions throughout the dissolution. Overall, several factors can contribute to this asymmetry, including the influence of the spatial distribution of the asperities within the core and the effect of the dissolution regime driven by the dimensionless numbers. As a prospect, investigating the effects of different spatial configurations on real-life material scans would contribute to validate these observations.

The initial state also exhibits heterogeneous uncertainty distribution on the whole calcite, with lower uncertainties on the pure solid region. As this mineral interface decreases due to the dissolution, the uncertainties tend to become more homogeneously distributed (see Fig \ref{fig:BMA_2D}c and \ref{fig:BMA_2D}d for instance). One can, however, notice that the local mean squared errors are significantly embedded in the micro-porosity uncertainties, ensuring reliable predictions. In Fig \ref{fig:Slices_t0_2D}, we detail these results on the initial state geometry --- for $t^*=0$ --- by plotting along the two dashed lines from  Fig \ref{fig:BMA_2D} the BMA and uncertainty on $\varepsilon_\Theta$, the groundtruth values, and the $\mu$CT observations. Considering the dynamical dissolution process also provide insight into the upper and lower bounds of the residual micro-porosity $\varepsilon_0$ for the initial calcite core geometry. In the porous matrix, we obtain the estimations $3\%\leq\varepsilon_0\leq 10\%$ for the 95\% confidence interval.

\begin{figure}
    \centering
    \includegraphics[width=\linewidth]{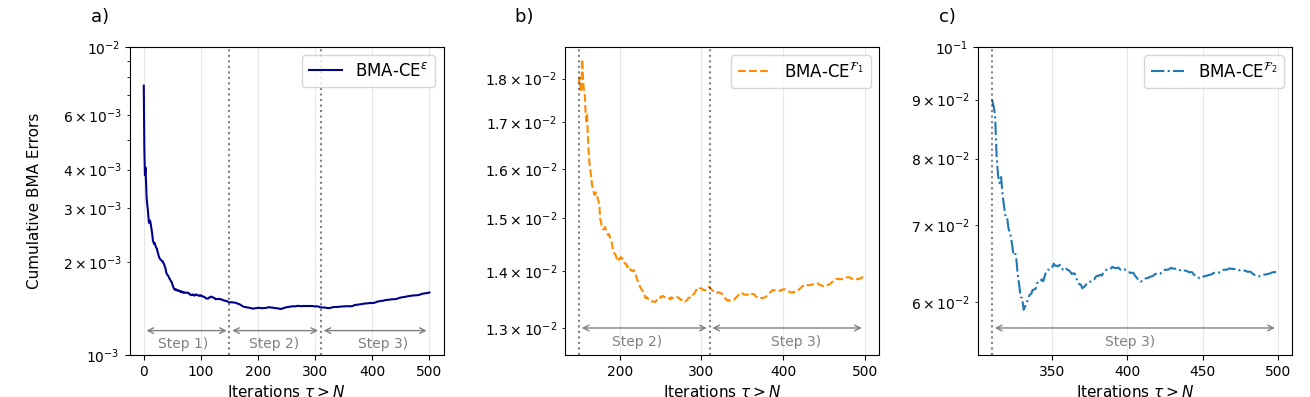}
 \caption{\textbf{Bayesian Model Average Cumulative Error diagnostics (BMA-CE) for 2D+Time reactive inverse problem:} BMA-CE on the micro-porosity field prediction $\varepsilon$ throughout the sampling iterations, and BMA-CE on the PDE constraint residuals $\mathcal{F}_1$ and $\mathcal{F}_2$ defined by equations \eqref{pde_constraints_2DTime} and introduced successively. The dotted vertical lines mark the sequential reinforcement of the multi-potential energy \eqref{multi_pot_2DTime}. }    \label{fig:BMA_CE_2D}
\end{figure}

The validation of the inference is first performed using the Bayesian Model Average Cumulative Error (BMA-CE) on the micro-porosity field, which is computed along the three successive sampling steps of sequential reinforcement. We then introduce $\mathcal{F}_1$ and $\mathcal{F}_2$ the PDE constraints residuals arising from the reactive model \eqref{eq:inverse_chem_adim_Vf} and involved in the multi-potential energy \eqref{multi_pot_2DTime}: 
\begin{equation}
    \label{pde_constraints_2DTime}
    \begin{aligned}
        & \mathcal{F}_1(\varepsilon_\Theta, C_\Theta) :=  (\upsilon C_0)^{-1}\alpha_\Theta \frac{\partial \varepsilon_\Theta}{\partial t} - C_\Theta \\[2mm]
        & \mathcal{F}_2(\varepsilon_\Theta, C_\Theta) := \gamma_\Theta\left( \frac{\partial C_\Theta}{\partial t} + (\upsilon C_0)^{-1}
    \frac{\partial \varepsilon_\Theta}{\partial t} \right) - \mathcal{D}_i(\varepsilon_\Theta, C_\Theta)
    \end{aligned}
\end{equation}
to estimate their $\text{BMA-CE}^{\mathcal{F}_\bullet}$ diagnostics on sampling steps 2 and 3. For the 2D+time data assimilation problem, the BMA-CE metrics on the micro-porosity field $\varepsilon$ and PDE residuals are straightforwardly extended from the formulae \eqref{BMA_CE_1DTime_eps} and \eqref{BMA_CE_1DTime_constraints}. The results are provided in Fig \ref{fig:BMA_CE_2D} and highlight the convergence of each term toward final BMA errors, at the sampling iteration $\tau=500$, scaling respectively about $\text{BMA-CE}^\varepsilon(\tau)=\num{1.6e-3}$, $\text{BMA-CE}^{\mathcal{F}_1}(\tau)=\num{1.4e-2}$ and $\text{BMA-CE}^{\mathcal{F}_2}(\tau)=\num{6.4e-2}$. 
From these convergence diagnostics, we observe a saturation of the PDE constraints that highlights the intrinsic uncertainties of their corresponding tasks in the multi-potential energy \eqref{multi_pot_2DTime}. In this sense, the PDE constraint $\mathcal{F}_2$ involving the heterogeneous diffusion operator \eqref{eq:Diff_op_dev} is the most uncertain term due to its high sensitivity to porosity variations. Nonetheless, we notice, as in the validation test cases from Sect. \ref{sec:Results1D}, that successive introduction of the PDE constraints leads to a slight increase in the $\text{BMA-CE}^\varepsilon$ curve, bringing information on the porosity field recovery by preventing overfitting issues. In particular, we observe in Fig \ref{fig:BMA_2D_evol_seq} that successively incorporating the physics-based constraints affects the uncertainty distribution of the porosity, especially at the core edges, rendering them physically reliable. In this sense, such a bounce in the BMA-CE predictions provides more dynamical insights into the porosity field $\varepsilon$. The increase observed on $\text{BMA-CE}^{\mathcal{F}_1}$ in Fig \ref{fig:BMA_CE_2D} arises from both a deviation in the surrogate micro-porosity field $\varepsilon_\Theta$ and a more physical approximation of the surrogate concentration field $C_\Theta$ in step 3 compared to step 2. Indeed, step 2 considers a simplified coupled model for the reactive system based on the quasi-stationary assumption to provide a first approximation of $C_\Theta$. In step 3, the physics-based assimilation is fully constrained, with the complex coupled dynamical behavior given by the PDE system \eqref{eq:inverse_chem_adim_Vf}, and we therefore expect such deviations both in the BMA-CE curves and uncertainty estimates.

\begin{figure}
    \centering
    \includegraphics[width=\linewidth]{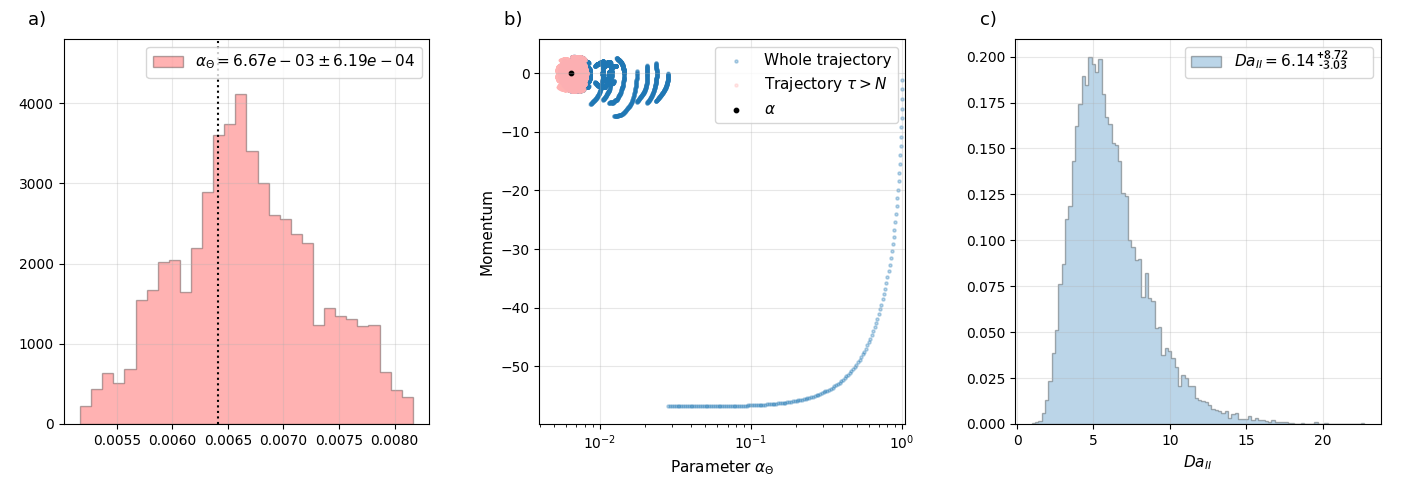}
    \caption{\textbf{Posterior distributions of 2D+Time reactive inverse problem:} a) Histogram of the marginal posterior distribution for the inverse parameter $\alpha_\Theta$. b) Phase diagram of its trajectory throughout the sampling, with the adaptive steps trajectories (in blue) and effective sampling (in red). The groundtruth value of $\alpha$ is represented by the black dot. c) Resulting posterior distribution of the catalytic Damköhler number $\Da2$, determined through the log-normal distribution from relation \eqref{eq:dist_da2}.}    
    \label{fig:Hist_2D}
\end{figure}

\begin{figure}
    \centering
    \includegraphics[width=\linewidth]{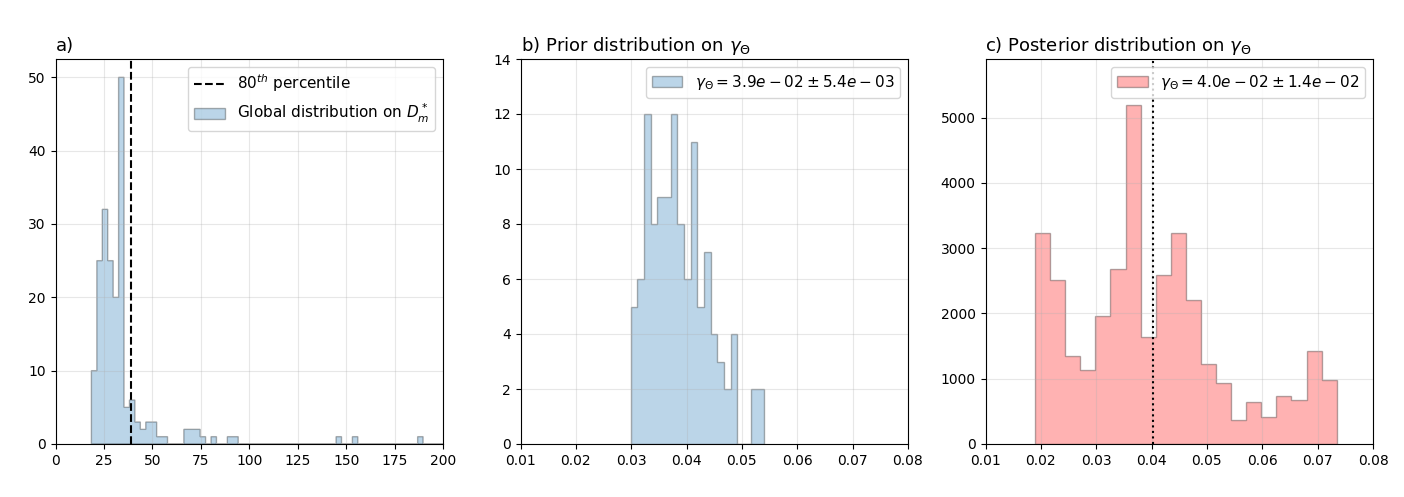}
  \caption{\textbf{Prior and posterior distributions on the inverse parameter $\gamma_\Theta$ for 2D+Time reactive inverse problem:} a) Global distribution on the estimated parameter $D_m^*$ arising from a-posteriori analysis on the sampling step 2. We discard the distribution tail after the vertical dashed line corresponding to the $80^\text{th}$ percentile. b) Resulting prior distribution on the inverse parameter $\gamma_\Theta$, used as an a-priori in step 3. c) Posterior distribution on $\gamma_\Theta$ obtained from the overall data assimilation problem throughout sampling step 3.}    
  \label{fig:Hist_2D_gamma}
\end{figure}

For the inverse parameters inference, we represent in Fig \ref{fig:Hist_2D} the histograms of the marginal posterior distributions of $\alpha_\Theta:=(\Da2^*)^{-1}$ and its trajectory in the phase space illustrating the convergence toward its mode during the adaptive steps --- represented in blue in Fig \ref{fig:Hist_2D}b. The latter shows that our data assimilation approach combined with the AW-HMC sampler from \cite{PEREZ2023112342} makes it possible to capture the correct parameter range without prior knowledge of its scaling. Once the adaptive process ends, we effectively start sampling the inverse parameter mode neighborhood, represented by the phase space trajectories for $\tau>N$ in red. We then follow the same process as in Sect. \ref{subsec:num_results_1D} to estimate the prior distribution on the inverse parameter $\gamma_\Theta$, and we obtain $\bar{\gamma_\Theta} = \num{3.9e-2}$ which is used as an initial a-priori in the sampling step 3 (see Fig \ref{fig:Hist_2D_gamma}a and Fig \ref{fig:Hist_2D_gamma}b). The posterior distribution on the inverse parameter $\gamma_\Theta$, estimated throughout step 3, is represented in Fig \ref{fig:Hist_2D_gamma}c and provides the following predictive interval $\gamma_\Theta\in[\num{2.6e-2},\num{5.4e-2}]$. One can, finally, estimate the posterior distribution on the catalytic Damköhler number $\Da2$ according to the log-normal distribution obtained by the relation \eqref{eq:dist_da2}. This results in the log-normal posterior distribution represented in Fig \ref{fig:Hist_2D} whose mean and standard deviations are respectively given by $\mathbb{E}[X] = 6.14$ and $\sqrt{\mathrm{Var}(X)} = 4.02$. Finally, we obtain the $95\%$ asymmetric confidence interval on the Damköhler number given by $\Da2 = 6.14^{+8.72}_{-3.03}$ for this data assimilation problem of calcite dissolution, which is consistent with the theoretical value.

\newpage
\section{Concluding remarks}
\label{sec:Conclusion}

This work intended to address two major challenges related to uncertainty quantification in pore-scale modeling of reactive flows, which plays a crucial role in the long-term management of \ce{CO2} capture and storage. Providing reliable macro-properties changes due to geochemical processes, such as \ce{CO2} mineral trapping and dissolution within the porous environment, is essential to query reservoir safety. In this sense, we aim to ensure that the evolving petrophysical properties provide meaningful characterizations of these chemical processes instead of intrinsic deviations arising from imaging limitations. Some intrinsic limiting factors remain when modeling pore-scale dynamical systems based on $\mu$CT scans and lead to several trade-offs that can bias the predictions. In particular, this results in unresolved micro-porosity, especially when the scan resolution does not fully capture morphological features of the pore space under the constraint of having a representative elementary volume (REV) of the sample. Quantifying sub-resolution porosity, which is a prevalent imaging artifact, was the first challenge we identified, and therefore we focused on quantifying morphological uncertainties in the micro-porosity field. Our second concern was to investigate the reliability of kinetic parameters, such as mineral reactivity, in the context of reactive processes. Indeed, these are critical parameters to account for in pore-scale modeling, though their experimental estimations can suffer from wide discrepancies. Estimating proper order of magnitude and uncertainty ranges appears essential to ensure reliable calibration of pore-scale models, and afterward trustable management of \ce{CO2} mineral storage. The present article investigated both these issues by integrating uncertainty quantification concerns in the workflow of pore-scale modeling.

Current methodologies investigating these problems regard them independently and fall into pure image treatment analysis related to experimental static or dynamical $\mu$CT images. Deep learning methodologies, for instance, are used to address sub-resolution porosity quantification. Multi-scale image reconstruction, which extrapolates the latent information of the porous structure, is obtained with Generative Adversarial Networks \cite{YANG2022104411}. One also gets super-resolved segmented images from static $\mu$CT scans through Convolutional Neural Networks \cite{alqahtani_super-resolved_2022}, which compensate for the unresolved morphological features. Regarding the mineral reactivity assessment, experimental works have been conducted on dynamical 4D $\mu$CT of carbonate dissolution. This provides, through differential imaging techniques, insight into local reaction rates at the mineral interfaces \cite{Noiriel4D, menke_dynamic_2015}. However, all these approaches do not consider incorporating uncertainty quantification in the estimates. In contrast, alternatives accounting for deviations in the petrophysical properties due to imaging limitations are mainly purely model-related approaches \cite{Soulaine_16, perez_deviation_2022}.

The main novelty of our work, therefore, lay in its ability to address both morphological uncertainty and reaction rate quantification from the perspective of coupling physics-based models with data-driven techniques. In this sense, we have developed a data assimilation approach for pore-scale imaging problems that combine dynamical microtomography and physical regularization induced by PDE models of reactive processes. We integrated this novel data assimilation strategy into the Bayesian inference context through our efficient AW-HMC framework for BPINNs \cite{PEREZ2023112342}. This also confirmed the great potential of this adaptive and self-balancing methodology and rendered BPINNs a promising approach to address complex data assimilation. In the pore-scale imaging context, in particular, we have focused on multitask inverse problems of calcite dissolution based on dynamical $\mu$CT images, along with two dimensionless inverse parameters and a latent concentration field. We have also assumed unknown informative priors on the different tasks scaling and relied on automatic adjustment of the uncertainties including noise-related estimations and model adequacy. In this sense, we provided reliable uncertainty quantification on the micro-porosity field description and reactive parameters. We also built our data assimilation upon a sequential reinforcement strategy of the multi-potential energy and thus the target posterior distribution. This involved successively integrating additional PDE constraints into the overall data assimilation process through dedicated sampling steps. Finally, we have also addressed computational concerns and have shown that suitable formulation of complex non-linear differential operators, especially the heterogeneous diffusion arising from Archie's law, can significantly reduce the computational costs of these operators. Taken together, we presented an intrinsic data assimilation strategy for pore-scale imaging inverse problem and demonstrated its efficiency on several 1D+Time and 2D+time calcite dissolution problems.

Overall, our results confirmed enhanced morphological uncertainties localized on the calcite core edges throughout the dissolution process. This characterized the challenge of capturing reliable mineral interfaces from the dynamical $\mu$CT images, and therefore query the confidence of mineral reactivity assessment using merely differential imaging techniques on the $\mu$CT scans. Combining data-driven and physics-based approaches thus offers a promising alternative to overcome the limitations of each approach individually, and alleviate biased predictions. We also obtained reliable insight into the upper and lower bounds for the residual, potentially unresolved, micro-porosity $\varepsilon_0$ in the porous matrix. These estimations can then be incorporated into direct numerical simulation solvers to measure the impact of these micro-porosity variations on the other petrophysical properties, such as permeability. This can also ensure that the macro-scale porosity evolutions due to the reactive processes are significant compared to these intrinsic morphological uncertainties at the pore scale. Finally, we have obtained posterior distribution on the dimensionless reactive parameters characterizing the dissolution inverse problems. We have shown that our data assimilation approach combined with the AW-HMC sampler made it possible to capture the correct parameter ranges without prior knowledge of their scaling, which confirmed the robustness and reliability of the inferences. Last but not least, we have identified uncertainty ranges on the usual catalytic Damköhler number $\Da2$ resulting from the prescribed PDE model and dynamical observations of the dissolution process. This is of great interest to aggregate experimental investigations and direct numerical simulations, and therefore guarantee the reliability of pore-scale modeling and simulation of reactive flows. We now have the potential to effectively address robust and reliable uncertainty quantification in pore-scale imaging and to manage the impact of $\mu$CT limitations on the petrophysical properties and reactive parameters. 

As future prospects, it would be interesting to apply the current data assimilation framework on real 3D samples obtained from experimental $\mu$CT dissolution scans. This could provide deeper insights into the relationship between experiments and mathematical modeling theory, thereby enhancing the reliability of predictive model calibration and the confidence in computational approaches for real-life reactive materials. However, the shape complexity of porous samples, along with the non-linear operators involved in the advection-diffusion-reaction PDE model, are additional concerns rendering these applications challenging. Preforming 4D data assimilation of real-life reactive material will especially demand the development of efficient computational strategies to speed up the performance. The framework proposed in this article could benefit from enhanced parallelization on GPU devices, although managing the increased computational demand on CPU and GPU resources will be critical. For example, task parallelization across different processors during the sampling phase of a multi-objective data assimilation problem should be investigated, along with potential coupling with the symmetric splitting approach of the standard Hamiltonian Monte Carlo suggested by Cobb \textit{et al.} \cite{cobb_scaling_2021}, allowing parallelization on data subsets during the training phase. Regarding the prediction step, one can also investigate standard domain decomposition distributed across multiple devices to enhance the computational efficiency. These strategies aim to optimize the method efficiency while ensuring sustainable computational costs, which becomes crucial when addressing complex real-world inverse problems and data assimilation. Another prospect is to extend the inference to different kinds of dissolution regime, including advection dominant phenomena. This would require insights into the velocity field to integrate the overall reactive hydrodynamics PDE system within the data assimilation framework. Nonetheless, this implies the recovery of a additional latent fields and inverse parameter, namely the velocity and a modified Peclet number $\Pe^*$. A potential approach is to leverage priors for hydrodynamics through fast surrogate modeling and fluid flow model proxy \cite{fluids4030126, SANTOS2020103539} based on the morphological a-priori on the porous medium evolution. Subsequently, these priors could be incorporated and adjusted into the data assimilation of a full reactive flow problem. In this context, the estimation of the bulk permeability in the Kozeny-Carman correlation remains challenging since it drives the amount of concentration heading inside the porous matrix. On the experimental side, 4D data assimilation of real porous samples could eventually suffer from temporal deformations and collapse due to variations in the mechanical stress during the dissolution, that will require data preprocessing to minimize the noise arising from these mechanical deformations. Overall, considering these additional challenges, the present methodology has already demonstrated sufficient robustness to handle heavily noised data and will offer the opportunity to quantify uncertainties in pore-scale imaging obtained from real $\mu$CT data.

\appendix
\normalsize

\section{Review on the AW-HMC algorithm and stopping criterion for the weight adaptation}
\label{sec:Algo_Stop_Criterion}

We briefly review the AW-HMC methodology previously developed in \cite{PEREZ2023112342}, along with the stopping criterion used for the weight adaptation. In Algorithm \ref{Adaptively Weighted Hamiltonian Monte Carlo}, $\Theta^{t_0}$ refers to the initial state for both the neural network and inverse parameters and characterises the particle positions in the physical analogy. $N_s$ is the total number of samples collected during the training and $N_{\mathrm{burn}}$ the burn-in steps, usually taken as the effective number of adaptive steps $N$. $L$ and $\delta t$ respectively represents the number of iterations and the step size used for in the leapfrog method to solve the Hamiltonian system \eqref{Ham_dyn_sys}, where $\mathbf{M}$ is the covariance or mass matrix for the distribution on the particle momenta $r$. $N_{\max}$ and $S_{\min}$ are the maximum number of adaptive iterations and threshold used as stopping criterion for for the adaptive weighting.

We consider the weighted multi-potential energy  
\begin{equation}
    \ds U(\Theta) = \sum_{k=0}^{K} \lambda_k \mathcal{L}_k(\Theta) + \lambda_{K+1}\|\Theta\|^2 := \sum_{k=0}^{K+1} \lambda_k \mathcal{L}_k(\Theta),
\end{equation}
introduced in Sect. \ref{subsec:BPINNs} and associated to the weighted Hamiltonian 
\begin{equation}
    \label{Weighted_Ham_bis}
    H_{\lambda_\tau}(\Theta, r) = \sum_{k=0}^{K+1} \lambda_k(\tau) \mathcal{L}_k(\Theta) + K(r) = \sum_{k=0}^{K+1} \lambda_k(\tau) \mathcal{L}_k(\Theta) + \frac{1}{2}r^T \mathbf{M}^{-1}r, 
\end{equation}
to define the $\lambda_k(\tau)$ along the sampling steps $\tau$ of automatic and adaptive weighting. One should notice that the weight on the prior term, corresponding to $\lambda_{K+1}$, is not adjusted since it rather acts as regularization term than a specific task in the multi-objective inverse problem (see \cite{PEREZ2023112342} for detailed development). The algorithm then alternates between deterministic steps by solving the Hamiltonian dynamical system for the fictive particle of position $\Theta$ and momentum $r$ during the leapfrog iterations (lines 11 to 16 in Algorithm \ref{Adaptively Weighted Hamiltonian Monte Carlo}), and stochastic steps resulting from the momentum sampling (lines 3 and 4) and allowing the exploration of various energy levels. The Metropolis-Hastings step, finally, plays the role of an acceptance criterion through a transition probability based on the weighted Hamiltonian and ensures energy preservation by rejecting samples that lead to divergent trajectories during the leapfrog numerical integration. Once all the samples are collected, predictive distributions on the main functional fields are built based on the Bayesian Model Average approximation \eqref{BMA}.

\begin{figure}
    \centering
    \includegraphics[width=0.89\linewidth]{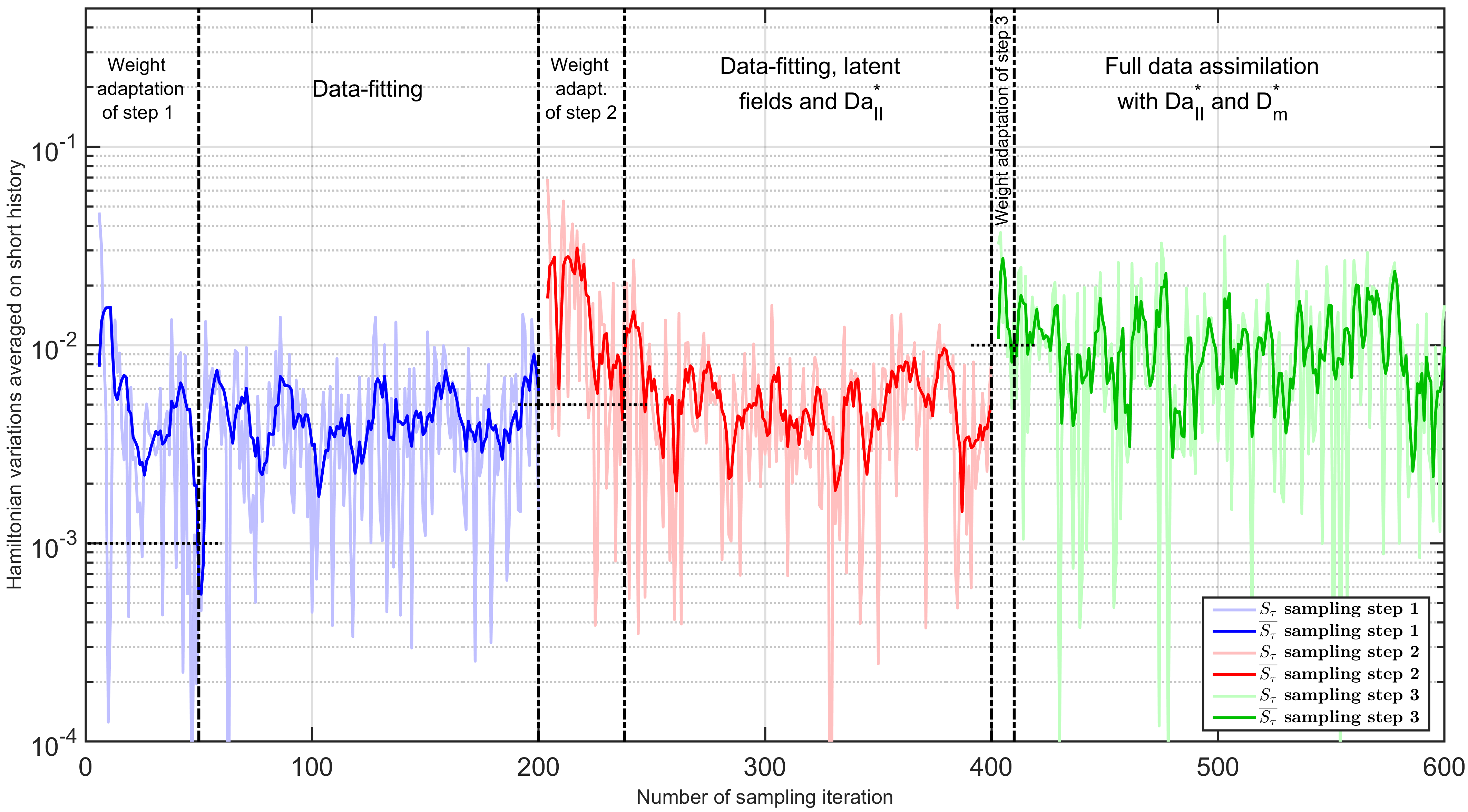}
  \caption{\textbf{Hamiltonian variations averaged on short history as stopping criterion for the weight adaptation:} The vertical dashed lines depict the start and the end of the weight adaptation processes for each step of the sequential reinforcement, developed in Sect. \ref{subsec:seq_reinforce}. The light-colored curves represent the best slopes $S_\tau$ of the Hamiltonian, computed by linear regression over a short history through equation \eqref{eq:SC}. The dark curves correspond to the variations $\bar S_\tau$ along the sampling iterations, with the threshold values $S_{\min}$ for each sequential step represented by the horizontal dashed lines.}
  \label{fig:Exit_2D}
\end{figure}

The stopping criterion for the weighting adaptation (line 6 in Algorithm  \ref{Adaptively Weighted Hamiltonian Monte Carlo}) is based on the maximum number of iterations $N_{\max}$ combined with the convergence threshold $S_{\min}$ on the local variation of the Hamiltonian $\bar{S}_\tau$. Indeed, if there is no significant variation of the Hamiltonian, we have reached a well-fitted weighted posterior distribution targeting the high probability-density region given by the Pareto front neighborhood, and the adaptation can be stopped. Given the stochastic nature of the Hamiltonian evolution, the variation $\bar{S}_\tau$ is computed as the locally average, over the last $p_{\mathrm{mean}}$ values, of the linear regression $S_\tau$ computing the Hamiltonian slopes over a short history ---  the last $p_{\mathrm{slope}}$ values of the Hamiltonian function. Denoting by $H_i$ the statistical series of the Hamiltonian evolution for the sampling iterations $X_i=i$, we thereby define $\bar{S}_\tau$ as:
\begin{equation}
\label{eq:SC}
\ds\bar{S}_\tau=\frac{1}{p_{\mathrm{mean}}}\sum_{i=\tau-p_{\mathrm{mean}}+1}^{\tau} S_i \quad \text{where} \quad
S_\tau=\frac{\mathrm{Cov}\left(\strut\{H_i\}_{i=\tau-p_{\mathrm{slope}}+1}^{\tau},\{X_i\}_{i=\tau-p_{\mathrm{slope}}+1}^{\tau}\right)}{\mathrm{Var}\left(\strut\{X_i\}_{i=\tau-p_{\mathrm{slope}}+1}^{\tau}\right)}    
\end{equation}
with $p_{\mathrm{slope}}=p_{\mathrm{mean}}$ for each step of the sequential reinforcement process. In the 2D+Time application from Sect. \ref{sec:Results}, the previous stopping criterion leads to effective numbers of adaptive steps of $N = 50,\,40$, and $10$ for each sequential sampling step, respectively given the threshold $S_{\min} = \num{1e-3}, \, \num{5e-3}$ and $\num{1e-2}$ (see Figure \ref{fig:Exit_2D}). Such an increase in the threshold value $S_{\min}$ is expected as the potential energy, and thereby the Hamiltonian, is successively reinforced by additional physical constraints.

\RestyleAlgo{boxruled}
\begin{algorithm}[!t]
\label{Adaptively Weighted Hamiltonian Monte Carlo}
\caption{Adaptively Weighted Hamiltonian Monte Carlo (AW-HMC)}

\BlankLine
\emph{Sampling procedure:} \\
\For {$\tau = 1...N_s$}{

Sample $r^{t_{\tau-1}} \sim \mathcal{N}(0, \mathbf{M})$\;
Set $(\Theta_0, r_0) \gets (\Theta^{t_{\tau-1}}, r^{t_{\tau-1}})$\;

\emph{Weights adaptation:} \\
    \eIf{$(\tau \le N_{\max})$ \textbf{\em and} $(\bar S_\tau\ge S_{\min})$}
    {\text{Compute }$\ds \lambda_k(\tau) =\left(\frac{\ds\min_{j=0..K}(\strut\mathrm{Var}\{\nabla_\Theta \mathcal{L}_j(\Theta_0) \})}{\mathrm{Var}\{\nabla_\Theta \mathcal{L}_k(\Theta_0) \} }\right)^{1/2} \quad \forall k = 0..K$ \text{and}  $\lambda_{K+1}(\tau) = 1$\;}
    {$\lambda_k(\tau) = \lambda_k(\tau-1) \quad \forall k = 0..K$ \text{and} $\lambda_{K+1}(\tau) = 1$}

\emph{Leapfrog:}\\
    \For{$i = 0...L-1$}
    {
    $\ds r_i \gets r_i - \frac{\delta t}{2} \sum_{k=0}^{K+1} \lambda_k(\tau) \nabla_\Theta \mathcal{L}_k(\Theta_i) $\;
    $\Theta_{i+1} \gets \Theta_i + \delta t \mathbf{M}^{-1}r_i$\;
    $\ds r_{i+1} \gets r_i -\frac{\delta t}{2} \sum_{k=0}^{K+1} \lambda_k(\tau) \nabla_\Theta \mathcal{L}_k(\Theta_{i+1}) $\; 
    }
\emph{Metropolis-Hastings:} \\
    Sample $p\sim \mathcal{U}(0,1)$\; 
    Compute $\alpha = \mathrm{min}(1, \mathrm{exp}(H_{\lambda_\tau}(\Theta_0 , r_0) - H_{\lambda_\tau}(\Theta_L, r_L) )$ using \eqref{Weighted_Ham_bis}\;

    \eIf{$p\leq \alpha$}
    {$\Theta^{t_\tau} = \Theta_L$\;}
    {$\Theta^{t_\tau} = \Theta_0$\;}
    Collect the samples after burn-in : $\left\{\Theta^{t_i}\right\}_{i=N_{\mathrm{burn}}}^{N_s}$
}
\end{algorithm}

\newpage


\end{document}